\documentclass{article} 
\usepackage[final]{colm2026_conference}

\usepackage{microtype}
\usepackage{hyperref}
\usepackage{url}
\usepackage{graphicx}
\usepackage{subcaption}
\usepackage{booktabs} 
\usepackage{enumitem}
\usepackage{algorithm}
\usepackage{algorithmic}

\usepackage{amsmath}
\usepackage{amssymb}
\usepackage{mathtools}
\usepackage{amsthm}
\usepackage{placeins}

\theoremstyle{plain}

\theoremstyle{definition}

\theoremstyle{remark}

\usepackage{lineno}

\definecolor{darkblue}{rgb}{0, 0, 0.5}
\definecolor{findingbg}{RGB}{250, 245, 224}
\definecolor{findingborder}{RGB}{219, 214, 196}
\hypersetup{colorlinks=true, citecolor=darkblue, linkcolor=darkblue, urlcolor=darkblue}

\newcounter{finding}
\newcommand{\finding}[2][]{%
  \refstepcounter{finding}%
  \par\medskip
  \noindent
  \begingroup
  \setlength{\fboxsep}{8pt}%
  \fcolorbox{findingborder}{findingbg}{%
    \parbox{\dimexpr\linewidth-2\fboxsep-2\fboxrule\relax}{%
      \textbf{Finding \thefinding.}%
      \if\relax\detokenize{#1}\relax\else\label{#1}\fi\ %
      \emph{#2}%
    }%
  }%
  \endgroup
  \par\medskip
}

\newcounter{frameworkextension}
\newcommand{\frameworkextension}[2][]{%
  \refstepcounter{frameworkextension}%
  \par\medskip
  \noindent
  \begingroup
  \setlength{\fboxsep}{8pt}%
  \fcolorbox{findingborder}{findingbg}{%
    \parbox{\dimexpr\linewidth-2\fboxsep-2\fboxrule\relax}{%
      \textbf{Framework Extension \theframeworkextension.}%
      \if\relax\detokenize{#1}\relax\else\label{#1}\fi\ %
      \emph{#2}%
    }%
  }%
  \endgroup
  \par\medskip
}

\title{Curriculum Learning as Transport:\\ Understanding Curricula with Wasserstein Geodesics}

\author{%
  Changho Shin%
  \thanks{Work done during a research internship at Microsoft Research.} \\
  Princeton University \\
  Princeton, NJ, USA \\
  \texttt{cs1095@princeton.edu} \\
  \And
  David Alvarez-Melis \\
  Microsoft Research \& Harvard University \\
  Cambridge, MA, USA \\
  \texttt{daalvare@microsoft.com} \\
}

\begin{document}

\ifcolmsubmission
\linenumbers
\fi

\maketitle

\newcommand{\SYSNAME}{\textsc{warp}}

\begin{abstract}
Curriculum learning is governed by several coupled design choices---how difficulty is defined, how examples are ordered, how much exposure each level receives, and how quickly training moves across levels---making it hard to isolate what actually helps. We present \textit{Wasserstein curriculum paths}, a simple transport-based framework that decouples these factors by representing curricula as trajectories of training distributions over discrete difficulty levels. Across a calibrated synthetic suite with 12 tasks and 33 difficulty axes, we use this framework to isolate the effects of ordering, matched exposure, endpoint smoothness, and pacing under fixed training budgets. We find that curriculum effects are strongly context-dependent: no single strategy dominates across tasks, difficulty axes, and budgets, and curricula mainly change where a fixed budget is spent most effectively. Within this framework, easy-to-hard ordering improves hard-level performance relative to exposure-matched static sampling, showing that the benefit is not explained by cumulative exposure alone. We further show that endpoint smoothness and pacing substantially affect where along the difficulty spectrum a curriculum is effective. Finally, we show that the same transport view naturally supports extensions to learned pacing through geometry and to structured difficulty spaces beyond one-dimensional orderings.\looseness-1
\end{abstract}

\section{Introduction}


Curriculum learning aims to improve training by presenting examples in a structured order, often from easier to harder instances \citep{bengio2009curriculum}. But in practice, curriculum effects depend on several coupled design choices: how difficulty is defined, whether training progresses easy-to-hard or hard-to-easy, how much total exposure each level receives, and how quickly the curriculum moves across levels \citep{hacohen2019power,wucurricula2021,jia2025makes}. As a result, even when a curriculum helps, it is often unclear why: are the gains due to ordering itself, greater exposure to easier examples, smoother transitions between levels, or a better pacing of a fixed training budget?

Prior curriculum methods—including self-paced selection, teacher-student schedulers, bandit approaches, and competence-based pacing—often improve performance \citep{kumar2010selfpaced,matiisen2017teacher,graves2017automated,platanios2019competence}, but they usually change several design choices at once, making it hard to isolate their contributions. Recent studies also ask more directly when ordering helps \citep{wucurricula2021,jia2025makes}, but they do not ultimately provide a common framework for separating ordering from matched exposure, endpoint choice, pacing, and geometry within one controlled setup.

We propose Wasserstein interpolation as a unified controlled framework for studying curriculum design. By viewing a curriculum as a path of training distributions over ordered difficulty levels,  Wasserstein paths provide a smooth way to move mass through nearby levels between easy-heavy and hard-heavy training distributions \citep{mccann1997convexity,peyre2019computational}. This lets us vary one design choice at a time while holding the others fixed, enabling controlled comparisons of ordering, matched exposure, endpoint smoothness, pacing, and geometry within a single parameterization.

Using a calibrated synthetic suite spanning 12 tasks and 33 difficulty axes, we use this framework to isolate which components of curriculum design matter and under what conditions. Three main findings emerge:

\textbf{Curriculum effects are context dependent.} Across the synthetic suite, no single curriculum dominates across tasks, difficulty axes, and budgets. Instead, curricula mainly shift where a fixed training budget is used most efficiently across difficulty levels, with smooth easy-to-hard progression particularly efficient on the hardest levels.

\textbf{Ordering matters beyond cumulative exposure.} Easy-to-hard progression improves hard-level performance relative to an exposure-matched static baseline, whereas hard-to-easy ordering often hurts, indicating that timing matters, not just cumulative exposure.

\textbf{Pacing and endpoint design matter.} Even within an easy-to-hard curriculum, the same path can help or hurt depending on how sharply the endpoint distributions are concentrated and how quickly the path is traversed. Very sharp endpoints are brittle under either overly aggressive or overly conservative pacing.\looseness-1

Beyond these main findings, we showcase how the framework can adapt pacing through the underlying geometry and extend curriculum paths from one-dimensional orderings to structured difficulty spaces.

Taken together, these results establish Wasserstein curricula as a principled framework for studying curriculum components, understanding when they matter, and extending curriculum design to adaptive pacing and structured difficulty spaces.

\section{Related Work}
We review the most relevant prior work here and provide further discussion in Appendix~\ref{app:extended_related_work}.

\paragraph{Curriculum learning and difficulty-aware training.} Curriculum learning studies whether organizing training examples by difficulty can improve optimization and generalization \citep{bengio2009curriculum}. Automated variants choose what to show and when through self-paced, teacher--student, bandit, and competence-based schedules \citep{kumar2010selfpaced,matiisen2017teacher,graves2017automated,platanios2019competence}, while training dynamics and related signals have been used to estimate example difficulty \citep{hacohen2019power,swayamdipta2020dataset,shrivastava2016ohem,toneva2019forgetting}. \citet{wang2021survey} organize this literature around difficulty measurement and training scheduling, and \citet{meng2025psychology} combine these components in a unified dynamic framework. Recent work also asks more directly when ordering helps: \citet{wucurricula2021} find clearer gains under limited budgets or label noise, while \citet{jia2025makes} show in LLM math post-training that the preferred direction depends on model capability, task complexity, and the difficulty metric. Our work studies a different question: given a difficulty structure, which aspects of the resulting curriculum matter? We represent curricula as paths over distributions on discrete difficulty levels and separate the difficulty geometry, endpoints, direction, cumulative exposure, and pacing. This enables controlled comparisons of factors that are often varied together, while also supporting extensions to adaptive geometry and structured difficulty spaces.

\paragraph{Optimal Transport}
Optimal transport (OT) provides a geometry for comparing probability distributions based on the cost of transporting mass between them \citep{villani2008optimal}. In this geometry, Wasserstein interpolation defines geodesic paths that smoothly transform one distribution into another \citep{mccann1997convexity,peyre2019computational,santambrogio2015otam}. Prior OT-based curriculum work, especially in reinforcement learning, uses these ideas to construct or sequence tasks by generating intermediate training stages between task or environment distributions \citep{huang2022crlot,klink2022cot}. In contrast, we use Wasserstein interpolation as an analytical framework over distributions on a fixed difficulty axis, allowing us to cleanly separate ordering, exposure, pacing, and endpoint effects within a single parameterization. This perspective focuses less on designing new curricula and more on isolating which components of curriculum design drive observed gains.


\begin{figure}[t]
  \centering
  \begin{subfigure}[t]{0.56\textwidth}
    \centering
    \includegraphics[width=\linewidth]{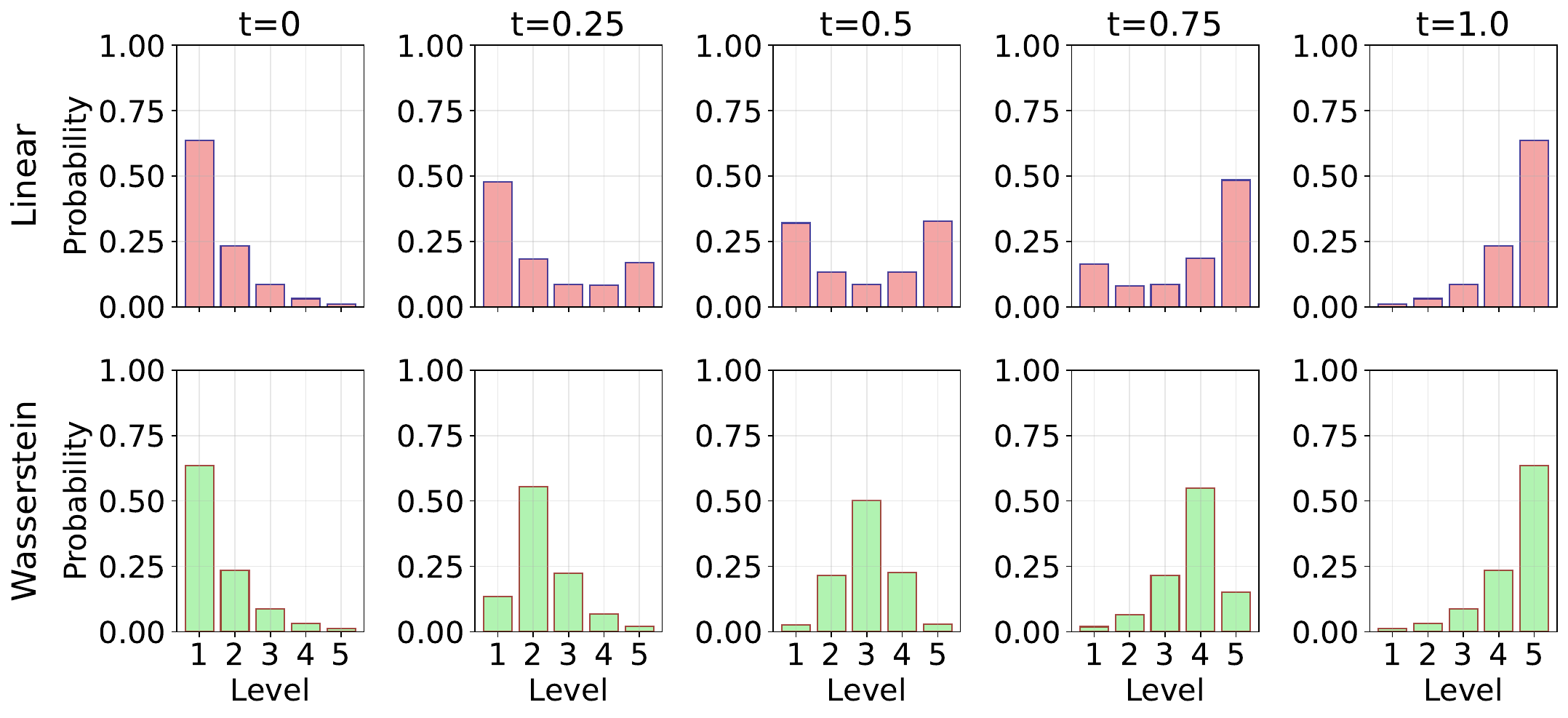}
    \caption{Linear vs.\ Wasserstein interpolation.}
    \label{fig:curriculum_visualization}
  \end{subfigure}%
  \hfill
  \begin{subfigure}[t]{0.43\textwidth}
    \centering
    \includegraphics[width=\linewidth]{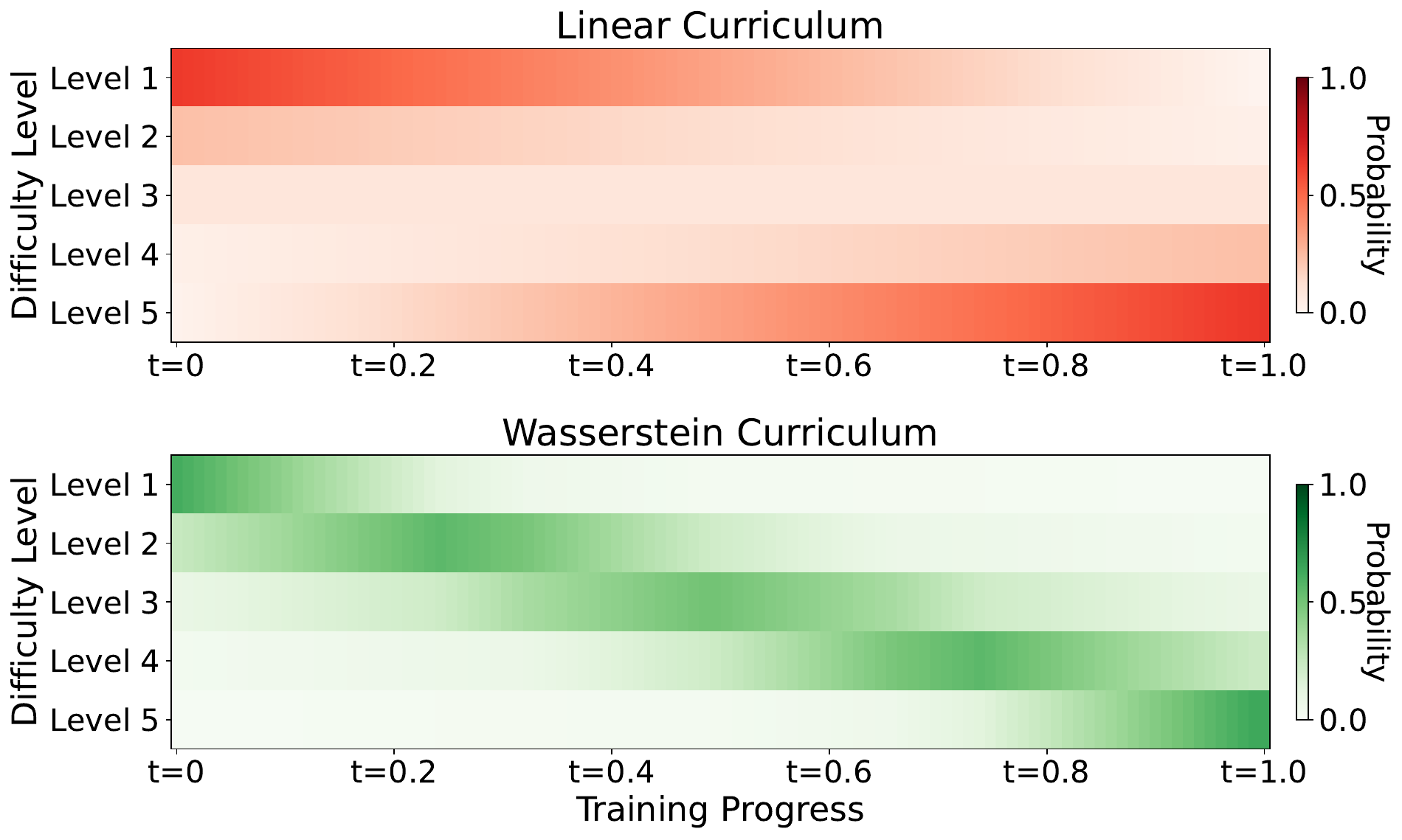}
    \caption{Curriculum heatmap.}
    \label{fig:curriculum_heatmap}
  \end{subfigure}
  \caption{ Comparison of linear and Wasserstein curricula. (a) Wasserstein interpolation moves probability mass progressively through the level geometry, whereas linear interpolation directly mixes the endpoint distributions. (b) Viewing \(P_t\) as the batch-level sampling distribution over time shows that Wasserstein places substantially more mass on intermediate levels during the transition. }
  \label{fig:curriculum_subplots}
  \vspace{-3mm}
\end{figure}

\section{Preliminaries and Experimental Setup}\label{sec:method}
We first introduce Wasserstein-geodesic curricula, then summarize the common protocol that turns them into a controlled study of curriculum design.

\subsection{Wasserstein–Geodesic Curriculum}
For each task, we partition the training data into \(L\) ordered difficulty levels. At training step \(k\), each batch is sampled from a probability vector \(P^k \in \Delta^{L-1}\), where \(P^k_\ell\) is the probability of drawing from level \(\ell\). A curriculum is therefore a sequence of sampling distributions over these levels. We represent it as a continuous path \(\{P_t\}_{t \in [0,1]}\) between endpoint distributions \(P_0, P_1 \in \Delta^{L-1}\), together with a schedule \(t_k \in [0,1]\) that maps training step \(k\) to a position on the path. The path specifies which mixtures of difficulty levels are visited, and the schedule specifies how quickly training moves through them, with \(P^k = P_{t_k}\).

\textbf{Wasserstein interpolation.}
We construct the path on the default one-dimensional ordering of levels: for
\(t \in [0,1]\),
\[
P_t \;=\; \arg\min_{Q \in \Delta^{L-1}}
\Big\{ (1-t)\,W(Q,P_0) + t\,W(Q,P_1) \Big\},
\]
where \(W\) is the Wasserstein distance on that fixed level geometry. This
moves probability mass smoothly through adjacent levels between the two
endpoint distributions.

For comparison, the linear interpolation baseline directly mixes the same
endpoints,
\[
P_t^{\mathrm{lin}} = (1-t)P_0 + tP_1.
\]
As illustrated in Figure~\ref{fig:curriculum_subplots}, linear interpolation
mixes the endpoint distributions directly, whereas the Wasserstein path
progresses through intermediate levels, producing a more natural curriculum.

\textbf{Curriculum with Wasserstein interpolation.}
By setting \(P^k = P_{t_k}\), we obtain curricula that move mass through nearby levels.  
The choice of endpoints \((P_0,P_1)\) determines where the curriculum starts and ends, 
the geometry determines which levels are treated as nearby, and the schedule \(\{t_k\}\) controls how quickly training moves along the path.  
This makes Wasserstein curricula useful as a study lens: once the task axis is fixed, they expose several curriculum factors that can be varied cleanly, including endpoint choice, pacing, and geometry, 
while preserving a natural progression over the same levels. Later sections use this structure to study endpoint families and traversal rates, learned geometry in \SYSNAME{}, and structured difficulty spaces beyond one-dimensional orderings.

\subsection{Common Experimental Setup}
We next introduce the common setup for our experimental suite. Unless varied explicitly in later sections, we keep the task construction, models, sampling protocol, and budget calibration fixed so that the suite can be used to study ordering, matched exposure, pacing, and geometry under shared conditions. We use this calibrated synthetic suite for the main study because it supports controlled comparisons; in Appendix~\ref{app:real_world_sft_analyses}, we show that in real-data SFT settings with pretrained models and noisier difficulty labels, curriculum effects are much weaker and harder to disentangle.

\paragraph{Tasks and difficulty control.} Unless otherwise stated, we study a synthetic suite of 12 tasks, adapted in part from SynLogic \citep{liu2025synlogic}. For each task, we construct ordered difficulty levels by varying one difficulty axis at a time, that is, one task attribute used to order examples from easier to harder while keeping the task semantics fixed. Calibrating and validating these within-task axes is part of the contribution, since it lets us compare curricula on controlled progressions rather than heterogeneous mixtures of examples. In the main study, this yields 33 task-by-difficulty-axis conditions, because several tasks contribute multiple difficulty axes. Appendix~\ref{app:task_details} gives the full task list, examples, and exact level constructions.

\paragraph{Models.} We use decoder-only Transformers \citep{vaswani2017attention} throughout, with model size matched to task complexity. Most tasks use small models trained from scratch, while language-like tasks fine-tune SmolLM-135M \citep{allal2024SmolLM}. Model details are reported in Appendix~\ref{app:model_training_details}.

\paragraph{Training protocol.} At training step \(k\), mini-batches are sampled from a distribution \(P^k \in \Delta^{L-1}\) over difficulty levels. We use \textit{static i.i.d.} for the non-curriculum baseline, where batches are drawn independently from the same fixed distribution throughout training; by default, that distribution is uniform over difficulty levels. Unless varied explicitly, our default easy-to-hard runs use an easy-heavy start distribution \(P_0\) and a hard-heavy end distribution \(P_1\), where \((P_1)_\ell \propto \exp(\ell/\tau)\) and \(P_0\) is the symmetric reversal of \(P_1\), i.e., \((P_0)_\ell = (P_1)_{L-\ell+1}\). We use default temperature \(\tau=1\) and default schedule \(t_k = k/T\).

\paragraph{Training budgets.} Training budget is a key variable in curriculum learning \citep{wucurricula2021}. Rather than fixing one global step count, we calibrate small, medium, and large regimes separately for each benchmark from pilot static i.i.d.\ learning curves, so that the static i.i.d.\ baseline spans early, intermediate, and later stages of learning on that task. This keeps budget comparisons aligned across tasks instead of comparing one task at an early stage of learning and another at a much later stage. Exact budget values are listed in Appendix~\ref{app:task_details}, with the associated model and training details in Appendix~\ref{app:model_training_details}.

\setlength{\abovedisplayskip}{4pt}
\setlength{\belowdisplayskip}{4pt}
\setlength{\abovedisplayshortskip}{2pt}
\setlength{\belowdisplayshortskip}{2pt}

\section{When and How Does Curriculum Learning Help?}\label{sec:when_and_how}
We first study whether curriculum learning helps reliably or only in specific regimes. For this, we compare \textit{static i.i.d.}, linear, and Wasserstein curricula across the 33 synthetic-suite task-by-difficulty-axis conditions in the main study, and report both overall and hardest-level accuracy.

\subsection{When Curricula Help Is Highly Context Dependent}\label{subsec:when_do_curricula_help}

\begin{table}[!t]
\centering
\small
\setlength{\tabcolsep}{6pt}
\begin{tabular}{lcccc}
\toprule
Curriculum & Mean overall (\%) & Mean hardest (\%) & Overall wins & Hardest wins \\
\midrule
\multicolumn{5}{l}{\textit{Small budget}} \\
Static i.i.d. & \(\mathbf{56.6}\) & \(39.9\) & \textbf{13} & 6 \\
Linear & \(55.4\) & \(\mathbf{46.4}\) & 7 & 12 \\
Wasserstein & \(\mathbf{56.6}\) & \(46.3\) & \textbf{13} & \textbf{15} \\
\midrule
\multicolumn{5}{l}{\textit{Medium budget}} \\
Static i.i.d. & \(79.2\) & \(66.1\) & 11 & 5 \\
Linear & \(78.8\) & \(\mathbf{72.6}\) & 10 & \textbf{16} \\
Wasserstein & \(\mathbf{79.7}\) & \(72.1\) & \textbf{12} & 12 \\
\midrule
\multicolumn{5}{l}{\textit{Large budget}} \\
Static i.i.d. & \(92.7\) & \(86.0\) & 6 & 1 \\
Linear & \(92.9\) & \(89.6\) & 12 & 13 \\
Wasserstein & \(\mathbf{93.2}\) & \(\mathbf{89.7}\) & \textbf{16} & \textbf{19} \\
\bottomrule
\end{tabular}
\caption{Absolute summary for the three basic curricula, stratified by training budget. Each block contains 33 task-by-difficulty-axis conditions. The win columns denote the number of conditions in which a curriculum performs best under the corresponding accuracy metric. ``Hardest'' denotes the last available level for each benchmark.}
\label{tab:basic_curriculum_summary}
\vspace{-2mm}
\end{table}

\paragraph{Setup.}
For each task-by-difficulty-axis condition at a given budget, we first average accuracy over 10 random-seed repetitions, then aggregate those condition-level means within each budget block. Each block contains 33 conditions.

\paragraph{Results.} Table~\ref{tab:basic_curriculum_summary} reports the resulting summary separately for the small, medium, and large regimes. No single curriculum dominates the suite. All three methods win substantial subsets of conditions, and the ranking shifts with budget rather than settling on a universal best choice. The main lesson is therefore contextual rather than universal: whether curriculum helps depends jointly on task structure, the difficulty axis, and the available budget. Appendix~\ref{appsubsec:taskwise_final_accuracy_barplots} gives the corresponding task-wise final-accuracy breakdowns.

\finding[find:dependent]{Curriculum effectiveness depends strongly on task, difficulty axis, and budget; no single strategy dominates.}

\begin{figure}[t]
  \centering
  \includegraphics[width=\linewidth]{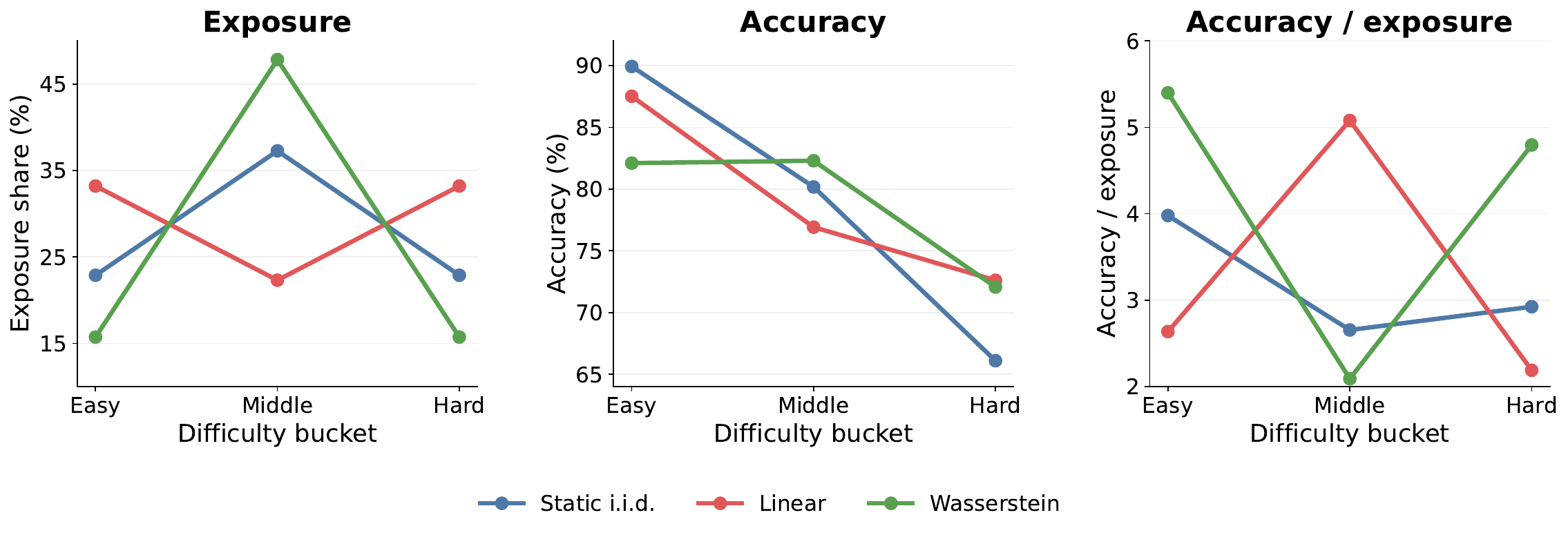}
  \caption{Exposure, accuracy, and exposure-adjusted accuracy under the medium budget. \textbf{Left:} cumulative exposure share over easiest, middle, and hardest probe buckets. \textbf{Middle:} final bucket accuracy on the same buckets. \textbf{Right:} exposure-adjusted accuracy on the same buckets. The middle bucket is the single middle level when the number of levels is odd and the merged middle pair when it is even, so these three probe buckets are not an exhaustive partition for five-level tasks. All panels compare only \textit{static i.i.d.}, \textit{linear}, and \textit{Wasserstein}. The corresponding profiles across all three budgets are shown in Figure~\ref{fig:appendix_b0_curriculum_profiles} in Appendix~\ref{appsubsec:budgetwise_curriculum_profiles}.}
  \label{fig:curriculum_profiles}
  \vspace{-5mm}
\end{figure}

\subsection{Curricula Shift the Data-Efficiency Profile Across Difficulty}\label{subsec:level_profile}

Similar overall accuracy means can hide very different learning profiles. We therefore compare curricula not only by final accuracy, but also by where they are data-efficient across difficulty levels.\looseness-1

\paragraph{Setup.}
Using the same setup as Section~\ref{subsec:when_do_curricula_help}, Figure~\ref{fig:curriculum_profiles} summarizes medium-budget behavior on the easiest, middle, and hardest probe buckets. Exposure is the cumulative share of training mass assigned to each bucket over the full run. As a simple proxy for data efficiency, we use \emph{exposure-adjusted accuracy}: final bucket accuracy divided by cumulative exposure to that bucket.

\paragraph{Results.}
Figure~\ref{fig:curriculum_profiles} shows that with a fixed budget, the main difference between curricula is not a uniform gain in final accuracy, but a different data-efficiency profile across difficulty. \textit{Static i.i.d.} remains strongest in final accuracy on the easiest bucket. Linear is strongest in exposure-adjusted accuracy in the middle bucket, while Wasserstein, which follows a natural smooth progression over the ordered levels, has the highest exposure-adjusted accuracy on the hardest bucket. This pattern is not driven by a few outliers: on the hard bucket, Wasserstein has higher exposure-adjusted accuracy than linear in all 99 task-by-axis-and-budget conditions. Figure~\ref{fig:appendix_b0_curriculum_profiles} in Appendix~\ref{appsubsec:budgetwise_curriculum_profiles} shows that the same qualitative pattern persists across the small, medium, and large budgets. The main effect of curriculum is therefore to reallocate where a fixed budget pays off, with Wasserstein especially efficient on the hardest bucket, which may matter when hard examples are scarce or costly. Appendix~\ref{appsubsec:budgetwise_curriculum_profiles} gives the corresponding budget-wise profile view, Appendix~\ref{appsubsec:taskwise_levelwise_profiles} shows the task-wise final level profiles, and Appendix~\ref{appsubsec:bridge_effect} revisits the hard-bucket advantage through controlled bridge-effect experiments.

\finding[find:sample_efficiency]{With a fixed budget, curricula reshape the data-efficiency profile across difficulty; Wasserstein is especially efficient on the hardest bucket.}

\section{Does Ordering Matter?}\label{sec:ordering}
Section~\ref{subsec:level_profile} showed that curricula with the same budget can induce different exposure and accuracy profiles across difficulty. We now ask whether ordering itself matters beyond cumulative exposure, comparing the default easy-to-hard Wasserstein path, the exposure-matched static baseline, and hard-to-easy Wasserstein on the same suite.

\begin{figure}[t]
  \centering
  \includegraphics[width=\linewidth]{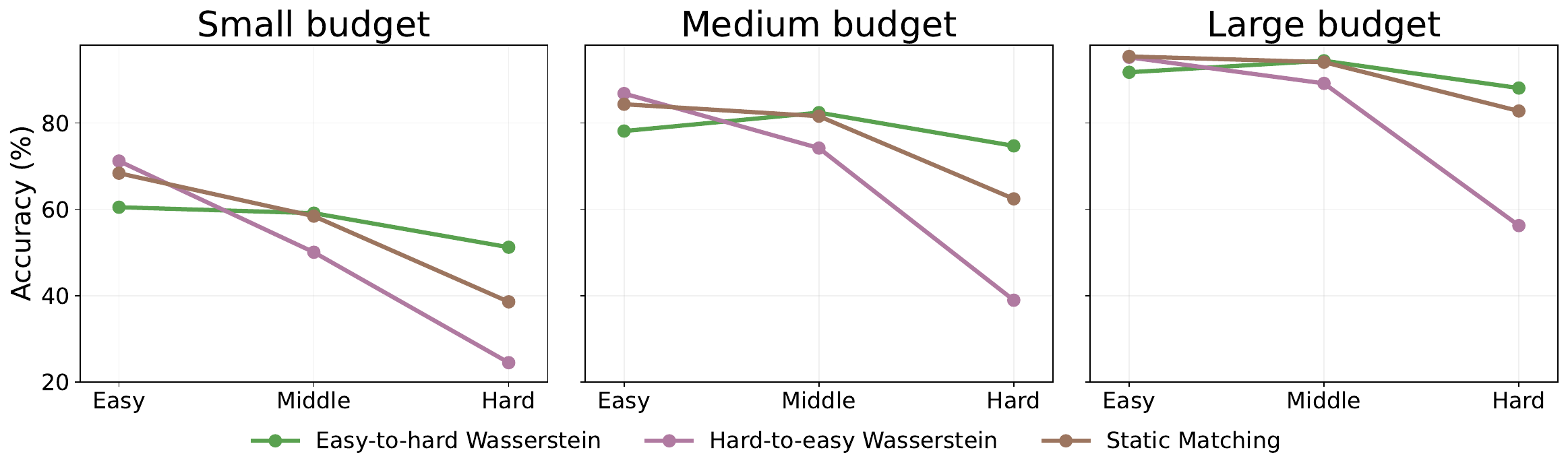}
  \caption{Ordering within the Wasserstein comparison. Each panel shows final accuracy on the easy, middle, and hard buckets for one training budget, corresponding to the first, middle, and hardest levels in the shared summary used here. ``Static Matching'' denotes the exposure-matched static baseline. Easy-to-hard Wasserstein consistently yields a flatter profile and stronger hard-bucket performance than either the static baseline or hard-to-easy Wasserstein.}
  \label{fig:ordering_profiles}
  \vspace{-5mm}
\end{figure}

\paragraph{Setup.}
Using the same default setup as in the previous sections, we compare easy-to-hard Wasserstein, the default ordering used throughout the paper, with two controls that keep cumulative exposure fixed while changing ordering. The exposure-matched static baseline, shown as ``Static Matching'' in Figure~\ref{fig:ordering_profiles}, preserves cumulative exposure to each level but removes the temporal progression, so it uses the same fixed batch-sampling distribution throughout training. Hard-to-easy Wasserstein traverses the same path family in reverse, and therefore preserves the same cumulative exposure while reversing the direction of progression.

\paragraph{Results.}
Figure~\ref{fig:ordering_profiles} shows that neither control recovers the easy-to-hard profile. The exposure-matched static baseline retains more strength on the easy bucket, but it misses the hard-bucket gains of easy-to-hard Wasserstein even when their overall averages remain similar. Hard-to-easy Wasserstein is more damaging still and often harms performance outright: it shifts strength toward the easy end and away from the hard bucket, where curricula matter most. This effect is broad rather than isolated to a few tasks: on the hardest level, forward Wasserstein exceeds reverse Wasserstein in 92 of 99 task-by-axis-and-budget conditions. The benefit of easy-to-hard progression therefore does not come from cumulative exposure alone; it comes from when examples are seen during training. Appendix~\ref{appsubsec:level_transfer} studies these directional transfer
patterns more directly with level-to-level transfer experiments.

\finding[find:matched_ordering]{Ordering matters beyond cumulative exposure: exposure-matched static sampling misses the hard-bucket gains of easy-to-hard progression, and hard-to-easy ordering often harms.}

\setlength{\abovedisplayskip}{4pt}
\setlength{\belowdisplayskip}{4pt}
\setlength{\abovedisplayshortskip}{2pt}
\setlength{\belowdisplayshortskip}{2pt}

\section{Pacing and Smoothness Matter}\label{sec:pacing_smoothness}\label{subsec:gamma_temp}
Even within an easy-to-hard curriculum, do pacing and endpoint smoothness still matter? We vary endpoint smoothness \((\tau)\) and traversal speed \((\gamma)\) along the same Wasserstein path.
\paragraph{Setup.}
We use seven tasks with short enough runtimes to make the full grid sweep feasible, giving 20 task-by-difficulty-axis conditions. Throughout this subsection we keep the same easy-to-hard Wasserstein path
\[
P_t \;=\; \arg\min_{Q \in \Delta^{L-1}}
\Big\{ (1-t)\,W(Q,P_0) + t\,W(Q,P_1) \Big\},
\]
and vary only its endpoint smoothness and pacing. The endpoint family and traversal are
\[
(P_1)_\ell \propto \exp(\ell/\tau), \qquad (P_0)_\ell = (P_1)_{L-\ell+1}, \qquad t_k = (k/T)^\gamma.
\]
Here \(\ell\) indexes difficulty levels, \(L\) is the number of levels, \(k\) is the training step, and \(T\) is the total number of training steps. Smaller \(\tau\) makes the endpoints sharper and larger \(\tau\) smooths them, while \(\gamma\) controls how quickly training moves along the path: smaller values move training toward harder levels sooner, while larger values keep training longer on easier and intermediate levels. We sweep a \(9\times 9\) grid over \(\tau,\gamma \in \{0.1, 0.25, 0.5, 0.75, 1.0, 1.5, 2.0, 5.0, 10.0\}\) in the medium-budget regime only, with three seeds per condition.

\begin{figure}[t]
  \centering
  \begin{subfigure}[t]{0.32\textwidth}
    \centering
    \includegraphics[height=0.155\textheight]{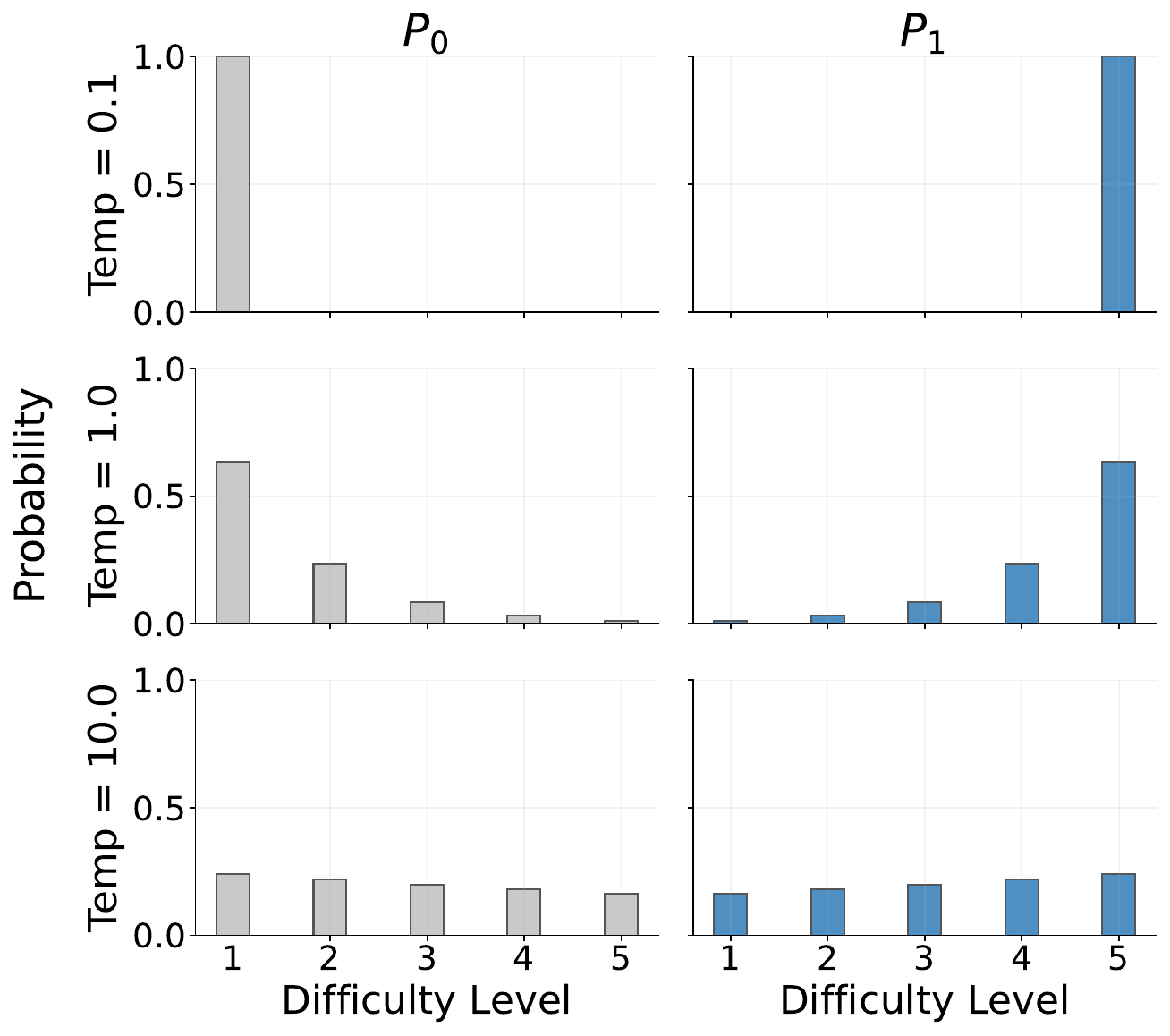}
    \caption{Endpoints vs.\ temperature \(\tau\).}
  \end{subfigure}\hfill
  \begin{subfigure}[t]{0.27\textwidth}
    \centering
    \includegraphics[height=0.155\textheight]{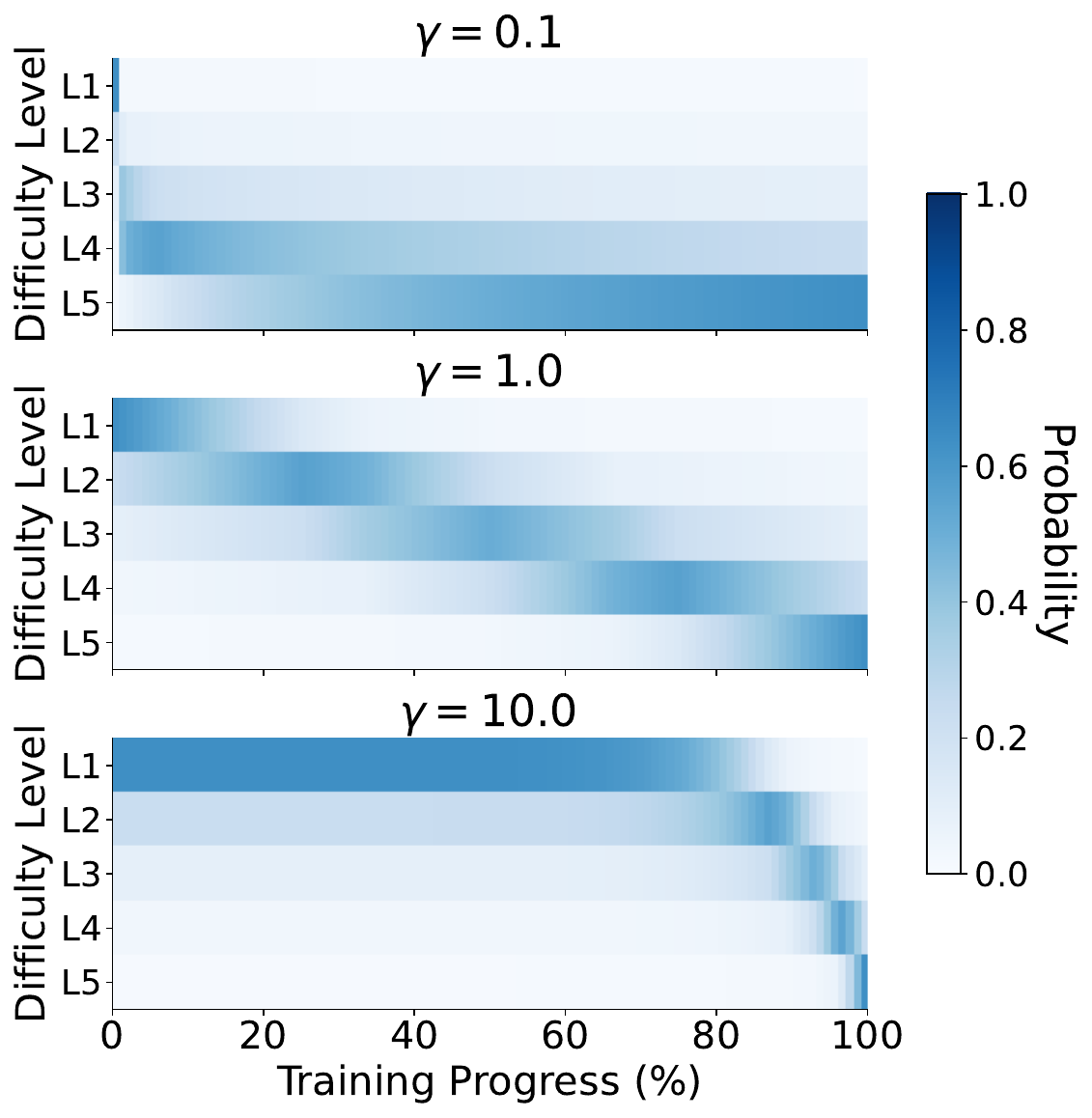}
    \caption{Batch-level path for \(\gamma\).}
  \end{subfigure}\hfill
  \begin{subfigure}[t]{0.33\textwidth}
    \centering
    \includegraphics[height=0.155\textheight]{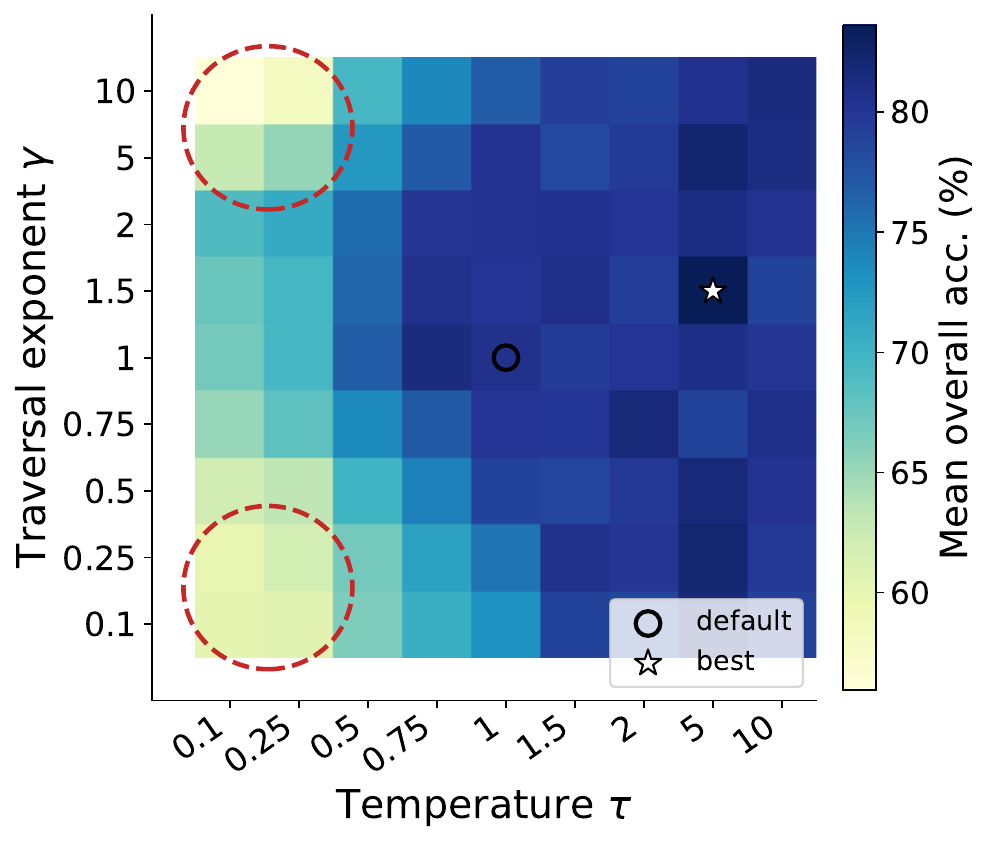}
    \caption{Accuracy across \((\tau,\gamma)\).}
  \end{subfigure}
  \caption{\textbf{(a)} Lower \(\tau\) sharpens the endpoints; larger \(\tau\) smooths them. \textbf{(b)} \(\gamma\) changes how fast training moves along the same easy-to-hard path. \textbf{(c)} Medium-budget sweep over 20 task-by-difficulty-axis conditions and three seeds. A broad region with moderate pacing and smoother endpoints performs best; the red dashed circles highlight collapse under very sharp endpoints with either overly aggressive or overly conservative pacing.}
  \label{fig:pacing_sweep}
  \vspace{-4mm}
\end{figure}

\paragraph{Results.}
Figure~\ref{fig:pacing_sweep}(c) reveals a broad good region rather than a single best setting. Moderate pacing and smoother endpoints work well, while very sharp endpoints are brittle under both overly aggressive and overly conservative pacing. The default setting already lies in a strong region, but it is not uniquely optimal. Easy-to-hard ordering alone is therefore not enough: the same path can help or hurt depending on smoothness and pacing.

\finding[find:pacing_smoothing]{Pacing and endpoint smoothness matter: very sharp endpoints can fail under either overly aggressive or overly conservative pacing.}

\section{Adaptive Geometry as Learnable Pacing}\label{sec:warp_geometry}\label{subsec:warp_geometry}
If pacing matters, the next question is how to adapt it without leaving the Wasserstein framework. We do so indirectly through geometry: because Wasserstein paths depend on the underlying distances between levels, changing that geometry changes pacing while preserving the same easy-to-hard endpoints. As a prototype, we instantiate this idea as \SYSNAME{} (Wasserstein-Adaptive Reshaping of Paths), which learns a geometry once after a short warmup and then follows the same forward Wasserstein path on the learned locations. Appendix~\ref{app:warp_definition} gives the full definition; here we only highlight the learned edge-length rule.

\paragraph{Setup.}
After a short uniform warmup, \SYSNAME{} probes gradient alignment between neighboring levels and converts adjacent cosine couplings into relative edge lengths, so higher similarity shortens an edge and lower similarity lengthens it.

\noindent
\begin{minipage}[t]{0.58\linewidth}
\vspace{0pt}
\textbf{Results.} Table~\ref{tab:warp_budget_summary} shows that even this first prototype beats fixed Wasserstein at small and medium budgets, while matching it at large budget. This proof-of-point result shows that adapting the geometry can improve pacing along the same easy-to-hard path.
\end{minipage}\hfill
\begin{minipage}[t]{0.38\linewidth}
\vspace{0pt}
\centering
\small
\setlength{\tabcolsep}{5pt}
\captionof{table}{Mean overall accuracy (\%) by budget.}
\label{tab:warp_budget_summary}
\vspace{-0.4\baselineskip}
\resizebox{\linewidth}{!}{
\begin{tabular}{lccc}
\toprule
Method & Small & Medium & Large \\
\midrule
Wasserstein & \(56.8\) & \(79.8\) & \(92.9\) \\
\SYSNAME{} & \(\mathbf{57.1}\) & \(\mathbf{80.6}\) & \(92.9\) \\
Gain & \(+0.2\) & \(+0.9\) & \(+0.0\) \\
\bottomrule
\end{tabular}
}
\end{minipage}


\begin{figure}[t]
  \centering
  \includegraphics[width=\linewidth]{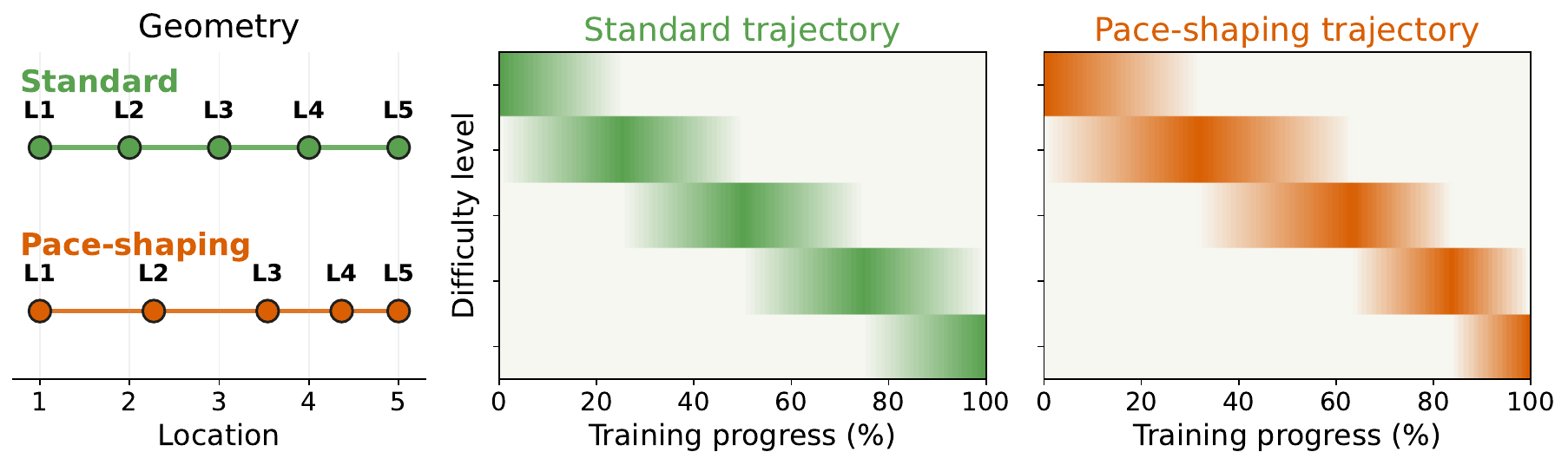}
  \caption{Geometry controls pacing even when endpoints stay fixed. Left: standard and an affinely rescaled learned geometry on the same level axis. Right: the resulting easy-to-hard Wasserstein trajectories. Stretching the early-to-mid region and compressing the hard end makes the path dwell longer before a sharper final handoff.}
  \label{fig:warp_geometry_motivation}
\end{figure}

\section{Structured Difficulty Spaces Beyond One-Dimensional Orderings}\label{sec:nonchain_geometry}
So far, we have treated difficulty as a one-dimensional ordering. Here we ask whether the same framework still works when levels are connected by a richer structure. To illustrate this, we construct two structured difficulty spaces from the arithmetic-operator family, a cube and a tree, and compare \textit{static i.i.d.}, \textit{Wasserstein}, and \SYSNAME{}.

\begin{figure*}[t]
  \centering
  \includegraphics[width=0.98\textwidth]{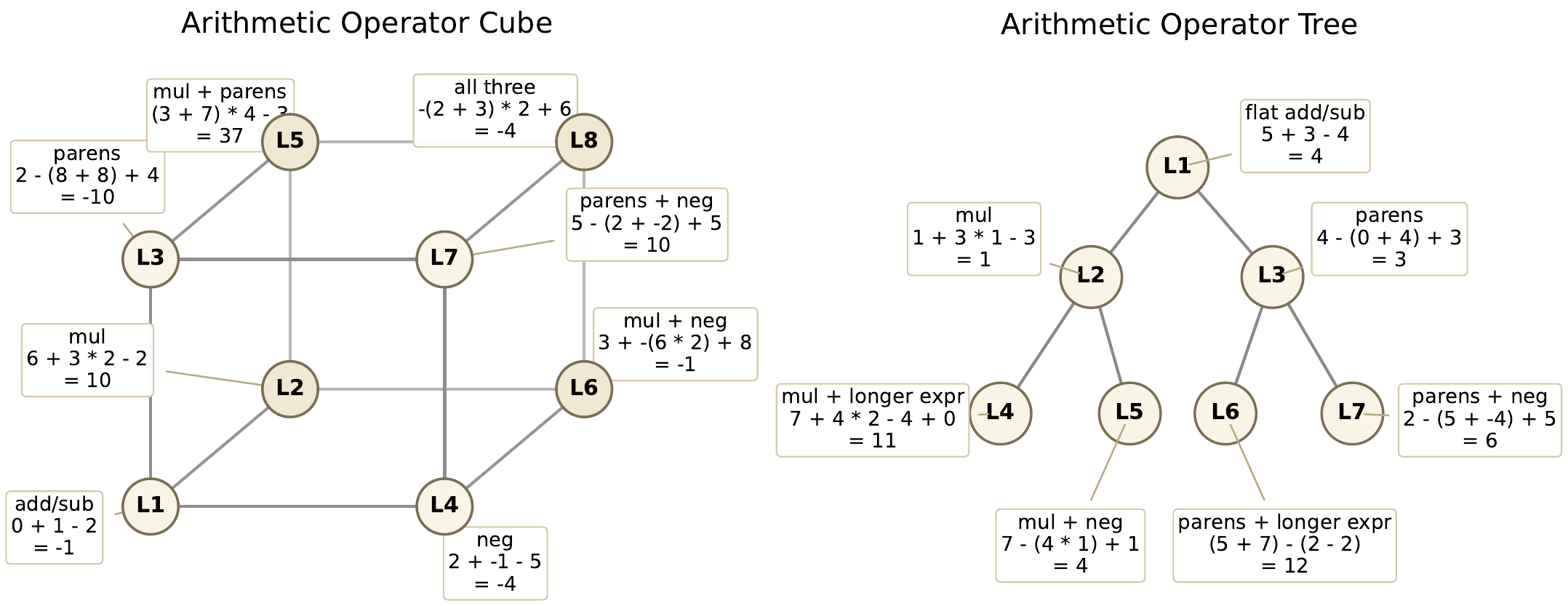}
  \caption{Structured benchmarks, a cube and a tree, built from arithmetic operators. Nodes are labeled by level, and nearby callouts show example expressions.}
  \label{fig:nonchain_geometry_graphs}
\end{figure*}

\begin{table*}[t]
\centering
\scriptsize
\setlength{\tabcolsep}{4pt}
\begin{minipage}[t]{0.48\textwidth}
\centering
\textbf{Cube}\\[2pt]
\begin{tabular}{llccc}
\toprule
Budget & Nodes & Static i.i.d. & Wasserstein & \SYSNAME{} \\
\midrule
Small  & All  & \(61.4\) & \(\mathbf{62.2}\) & \(59.3\) \\
 & Init & \(\mathbf{82.9}\) & \(77.3\) & \(70.7\) \\
 & Mid  & \(60.8\) & \(\mathbf{61.3}\) & \(59.2\) \\
 & End  & \(43.7\) & \(\mathbf{53.0}\) & \(48.4\) \\
Medium & All  & \(\mathbf{77.6}\) & \(75.1\) & \(75.6\) \\
 & Init & \(\mathbf{95.0}\) & \(87.6\) & \(91.4\) \\
 & Mid  & \(\mathbf{77.5}\) & \(74.5\) & \(75.7\) \\
 & End  & \(60.5\) & \(\mathbf{66.3}\) & \(59.1\) \\
Large & All  & \(\mathbf{93.3}\) & \(93.2\) & \(90.5\) \\
 & Init & \(\mathbf{99.4}\) & \(98.4\) & \(96.6\) \\
 & Mid  & \(\mathbf{93.6}\) & \(93.1\) & \(90.5\) \\
 & End  & \(85.9\) & \(\mathbf{89.0}\) & \(84.0\) \\
\bottomrule
\end{tabular}
\end{minipage}\hfill
\begin{minipage}[t]{0.48\textwidth}
\centering
\textbf{Tree}\\[2pt]
\begin{tabular}{llccc}
\toprule
Budget & Nodes & Static i.i.d. & Wasserstein & \SYSNAME{} \\
\midrule
Small  & All  & \(62.9\) & \(67.7\) & \(\mathbf{71.2}\) \\
 & Init & \(72.7\) & \(71.3\) & \(\mathbf{73.0}\) \\
 & Mid  & \(55.3\) & \(63.6\) & \(\mathbf{65.9}\) \\
 & End  & \(64.3\) & \(68.9\) & \(\mathbf{73.4}\) \\
Medium & All  & \(\mathbf{78.9}\) & \(77.2\) & \(76.1\) \\
 & Init & \(85.8\) & \(80.7\) & \(\mathbf{86.0}\) \\
 & Mid  & \(73.3\) & \(\mathbf{75.0}\) & \(72.5\) \\
 & End  & \(\mathbf{80.0}\) & \(77.3\) & \(75.4\) \\
Large & All  & \(86.7\) & \(\mathbf{90.3}\) & \(90.0\) \\
 & Init & \(95.3\) & \(94.5\) & \(\mathbf{95.4}\) \\
 & Mid  & \(83.1\) & \(\mathbf{88.2}\) & \(86.2\) \\
 & End  & \(86.4\) & \(90.4\) & \(\mathbf{90.5}\) \\
\bottomrule
\end{tabular}
\end{minipage}
\caption{Accuracy (\%, ten-seed mean) on the structured arithmetic-operator benchmarks.  \textsc{Init}/\textsc{Mid}/\textsc{End} denote the root, internal nodes, and leaves for the tree, and \(L1\), \(L2\)--\(L7\), and \(L8\) for the cube.}
\label{tab:nonchain_geometry}
\end{table*}

\paragraph{Setup.}
We use the 2-layer transformer and compare \textit{static i.i.d.}, \textit{Wasserstein}, and \SYSNAME{} over 10 seeds on the cube and tree benchmarks shown in Figure~\ref{fig:nonchain_geometry_graphs}. In both benchmarks, each node is a level in the same local problem family, but the connectivity differs. Appendix~\ref{app:nonchain_geometry_details} gives the exact graph constructions and endpoint distributions.

\paragraph{Results.}
Table~\ref{tab:nonchain_geometry} shows that geometry-aware curricula remain useful beyond one-dimensional orderings, with the clearest gains near the hard end. On the cube, \textit{Wasserstein} consistently improves the hard endpoint \(L8\), while \textit{Static i.i.d.} remains strongest on the easier regions and on aggregate at medium and large budgets. On the tree, both geometry-aware curricula are competitive: \SYSNAME{} is strongest in the small-budget regime, especially away from the root, whereas \textit{Wasserstein} becomes strongest overall in the large-budget regime. These experiments show that the same path construction extends to
graph-structured difficulty spaces, with effects that vary across graph, region, and training budget.

\section{Conclusion}\label{sec:conclusion}
We use Wasserstein interpolation as a controlled lens for studying curriculum learning. This lens lets us vary calibrated difficulty axes, ordering, matched cumulative exposure, endpoint smoothness, pacing, and geometry within one framework. Taken together, our results do not support a single universally best curriculum. Instead, they show that different components of curriculum design matter in different ways: easy-to-hard ordering helps beyond cumulative exposure, pacing and endpoint design change where a fixed budget is most effective, adaptive geometry can yield gains, and structured difficulty spaces make those gains more localized and more dependent on structure. At the same time, our real-data SFT results suggest that these effects weaken substantially in settings with pretrained models, natural benchmarks, and noisier difficulty labels.

\paragraph{Limitations and Future Work.}
We intentionally constrain most of the study to from-scratch training and newly constructed tasks, reducing confounding from pretrained knowledge but leaving open how the results transfer to pretrained settings, natural data, and tasks whose difficulty structure interacts with prior knowledge. We also showcase two extensions of the framework, adaptive geometry and structured difficulty spaces, pointing to next directions such as learning geometry or pacing from transfer signals, defining Wasserstein curricula on richer structures, and testing them in pretrained settings.

\section*{Disclosure of LLM Use}
We used LLM-based assistants during this project for parts of task design, code implementation, exploratory analysis, plotting and analysis code, and drafting or revising portions of the manuscript. These tools were used interactively under close author supervision. The authors made the final research decisions, ran and verified the reported experiments and analyses, checked the resulting figures and tables, and reviewed and edited all final paper text.

\section*{Acknowledgements}
The majority of this work was done while CS and DAM were at Microsoft Research. CS thanks Brenden Lake for supporting conference travel. DAM additionally acknowledges support from the Kempner Institute for the Study of Natural and Artificial Intelligence, the Aramont Fellowship Fund and NSF Award No. 2229881, NSF AI Institute for Societal Decision Making (NSF AI-SDM).

\bibliography{ref}

\begin{thebibliography}{35}
\providecommand{\natexlab}[1]{#1}
\providecommand{\url}[1]{\texttt{#1}}
\expandafter\ifx\csname urlstyle\endcsname\relax
  \providecommand{\doi}[1]{doi: #1}\else
  \providecommand{\doi}{doi: \begingroup \urlstyle{rm}\Url}\fi

\bibitem[Amari(2016)]{amari2016information}
Shun-ichi Amari.
\newblock \emph{Information Geometry and Its Applications}, volume 194 of
  \emph{Applied Mathematical Sciences}.
\newblock Springer Japan, 2016.
\newblock \doi{10.1007/978-4-431-55978-8}.
\newblock URL \url{https://doi.org/10.1007/978-4-431-55978-8}.

\bibitem[Belenki et~al.(2025)Belenki, Agarwal, Shi, and Toutanova]{liu2025mde}
Lior Belenki, Alekh Agarwal, Tianze Shi, and Kristina Toutanova.
\newblock Optimizing pre-training data mixtures with mixtures of data expert
  models.
\newblock In \emph{Proceedings of the 63rd Annual Meeting of the Association
  for Computational Linguistics (Volume 1: Long Papers)}, pp.\  32570--32587.
  Association for Computational Linguistics, 2025.
\newblock \doi{10.18653/v1/2025.acl-long.1564}.
\newblock URL \url{https://aclanthology.org/2025.acl-long.1564/}.

\bibitem[Ben~Allal et~al.(2024)Ben~Allal, Lozhkov, Bakouch, von Werra, and
  Wolf]{allal2024SmolLM}
Loubna Ben~Allal, Anton Lozhkov, Elie Bakouch, Leandro von Werra, and Thomas
  Wolf.
\newblock Smollm - blazingly fast and remarkably powerful.
\newblock Hugging Face blog, 2024.
\newblock URL \url{https://huggingface.co/blog/smollm}.

\bibitem[Bengio et~al.(2009)Bengio, Louradour, Collobert, and
  Weston]{bengio2009curriculum}
Yoshua Bengio, J{\'e}r{\^o}me Louradour, Ronan Collobert, and Jason Weston.
\newblock Curriculum learning.
\newblock In \emph{International Conference on Machine Learning}, pp.\  41--48.
  ACM, 2009.
\newblock \doi{10.1145/1553374.1553380}.

\bibitem[Clark et~al.(2018)Clark, Cowhey, Etzioni, Khot, Sabharwal, Schoenick,
  and Tafjord]{clark2018arc}
Peter Clark, Isaac Cowhey, Oren Etzioni, Tushar Khot, Ashish Sabharwal, Carissa
  Schoenick, and Oyvind Tafjord.
\newblock Think you have solved question answering? try {ARC}, the {AI2}
  reasoning challenge.
\newblock \emph{arXiv.org}, 2018.
\newblock \doi{10.48550/arXiv.1803.05457}.
\newblock URL \url{https://arxiv.org/abs/1803.05457}.

\bibitem[Fan et~al.(2024)Fan, Pagliardini, and Jaggi]{fan2024domainmixing}
Simin Fan, Matteo Pagliardini, and Martin Jaggi.
\newblock {DOGE}: Domain reweighting with generalization estimation.
\newblock In \emph{International Conference on Machine Learning}, pp.\
  12895--12915. PMLR, 2024.
\newblock \doi{10.48550/arXiv.2310.15393}.
\newblock URL \url{https://proceedings.mlr.press/v235/fan24e.html}.

\bibitem[Ge et~al.(2025)Ge, Huang, Cooper, Trost, Chu, Gnvv, Cai, Park,
  Roberts, and Sala]{huang2025rb}
Albert Ge, Tzu-Heng Huang, John Cooper, Avi Trost, Ziyi Chu, Satya Sai
  Srinath~Namburi Gnvv, Ziyang Cai, Kendall Park, Nicholas Roberts, and
  Frederic Sala.
\newblock R\&b: Domain regrouping and data mixture balancing for efficient
  foundation model training.
\newblock \emph{arXiv.org}, 2025.
\newblock \doi{10.48550/arXiv.2505.00358}.
\newblock URL \url{https://arxiv.org/abs/2505.00358}.

\bibitem[Geva et~al.(2021)Geva, Khashabi, Segal, Khot, Roth, and
  Berant]{geva2021strategyqa}
Mor Geva, Daniel Khashabi, Elad Segal, Tushar Khot, Dan Roth, and Jonathan
  Berant.
\newblock Did aristotle use a laptop? a question answering benchmark with
  implicit reasoning strategies.
\newblock \emph{Transactions of the Association for Computational Linguistics},
  9:\penalty0 346--361, 2021.
\newblock \doi{10.1162/tacl_a_00370}.
\newblock URL \url{https://aclanthology.org/2021.tacl-1.21/}.

\bibitem[Graves et~al.(2017)Graves, Bellemare, Menick, Munos, and
  Kavukcuoglu]{graves2017automated}
Alex Graves, Marc~G. Bellemare, Jacob Menick, R{\'e}mi Munos, and Koray
  Kavukcuoglu.
\newblock Automated curriculum learning for neural networks.
\newblock In \emph{International Conference on Machine Learning}, pp.\
  1311--1320, 2017.
\newblock URL \url{https://proceedings.mlr.press/v70/graves17a.html}.

\bibitem[Hacohen \& Weinshall(2019)Hacohen and Weinshall]{hacohen2019power}
Guy Hacohen and Daphna Weinshall.
\newblock On the power of curriculum learning in training deep networks.
\newblock In \emph{International Conference on Machine Learning}, pp.\
  2535--2544, 2019.
\newblock URL \url{https://proceedings.mlr.press/v97/hacohen19a.html}.

\bibitem[Hase et~al.(2024)Hase, Bansal, Clark, and
  Wiegreffe]{hase2024unreasonable}
Peter Hase, Mohit Bansal, Peter Clark, and Sarah Wiegreffe.
\newblock The unreasonable effectiveness of easy training data for hard tasks.
\newblock In \emph{Proceedings of the 62nd Annual Meeting of the Association
  for Computational Linguistics (Volume 1: Long Papers)}, pp.\  7002--7024,
  Bangkok, Thailand, 2024. Association for Computational Linguistics.
\newblock \doi{10.18653/v1/2024.acl-long.378}.
\newblock URL \url{https://aclanthology.org/2024.acl-long.378/}.

\bibitem[Hendrycks et~al.(2021)Hendrycks, Burns, Basart, Zou, Mazeika, Song,
  and Steinhardt]{hendrycks2021mmlu}
Dan Hendrycks, Collin Burns, Steven Basart, Andy Zou, Mantas Mazeika, Dawn
  Song, and Jacob Steinhardt.
\newblock Measuring massive multitask language understanding.
\newblock In \emph{International Conference on Learning Representations}, 2021.
\newblock URL \url{https://openreview.net/forum?id=d7KBjmI3GmQ}.

\bibitem[Hu et~al.(2022)Hu, Shen, Wallis, Allen-Zhu, Li, Wang, Wang, and
  Chen]{hu2022lora}
Edward~J. Hu, Yelong Shen, Phillip Wallis, Zeyuan Allen-Zhu, Yuanzhi Li, Shean
  Wang, Lu~Wang, and Weizhu Chen.
\newblock {LoRA}: Low-rank adaptation of large language models.
\newblock In \emph{International Conference on Learning Representations}, 2022.
\newblock URL \url{https://openreview.net/forum?id=nZeVKeeFYf9}.

\bibitem[Huang et~al.(2022)Huang, Xu, Zhu, Shi, Fang, and Zhao]{huang2022crlot}
Peide Huang, Mengdi Xu, Jiacheng Zhu, Laixi Shi, Fei Fang, and Ding Zhao.
\newblock Curriculum reinforcement learning using optimal transport via gradual
  domain adaptation.
\newblock In \emph{Advances in Neural Information Processing Systems 35}, pp.\
  10656--10670. Neural Information Processing Systems Foundation, Inc.
  (NeurIPS), 2022.
\newblock \doi{10.52202/068431-0774}.
\newblock URL \url{https://arxiv.org/abs/2210.10195}.

\bibitem[Jia et~al.(2026)Jia, Zhang, Diao, Yuan, Ouyang, and
  Vosoughi]{jia2025makes}
Yaning Jia, Chunhui Zhang, Xingjian Diao, Xiangchi Yuan, Z.~Ouyang, and
  S.~Vosoughi.
\newblock What makes a good curriculum? disentangling the effects of data
  ordering on llm mathematical reasoning.
\newblock \emph{Proceedings of the 64th Annual Meeting of the Association for
  Computational Linguistics (Volume 1: Long Papers)}, pp.\  34472--34488, 2026.
\newblock \doi{10.18653/v1/2026.acl-long.1591}.
\newblock URL \url{https://arxiv.org/abs/2510.19099}.

\bibitem[Klink et~al.(2022)Klink, Yang, D'Eramo, Peters, and
  Pajarinen]{klink2022cot}
Pascal Klink, Haoyi Yang, Carlo D'Eramo, Jan Peters, and Joni Pajarinen.
\newblock Curriculum reinforcement learning via constrained optimal transport.
\newblock In \emph{International Conference on Machine Learning}, volume 162 of
  \emph{Proceedings of Machine Learning Research}, pp.\  11341--11358. PMLR,
  2022.
\newblock URL \url{https://proceedings.mlr.press/v162/klink22a.html}.

\bibitem[Kumar et~al.(2010)Kumar, Packer, and Koller]{kumar2010selfpaced}
M.~Pawan Kumar, Benjamin Packer, and Daphne Koller.
\newblock Self-paced learning for latent variable models.
\newblock In \emph{Neural Information Processing Systems}, 2010.
\newblock URL
  \url{https://proceedings.neurips.cc/paper/2010/hash/e57c6b956a6521b28495f2886ca0977a-Abstract.html}.

\bibitem[Liu et~al.(2025{\natexlab{a}})Liu, Fan, Jiang, Ding, Hu, Zhang, Shi,
  Weng, Chen, Chen, et~al.]{liu2025synlogic}
Junteng Liu, Yuanxiang Fan, Zhuo Jiang, Han Ding, Yong Hu, Chi Zhang, Yiqi Shi,
  Shitong Weng, Aili Chen, Shiqi Chen, et~al.
\newblock Synlogic: Synthesizing verifiable reasoning data at scale for
  learning logical reasoning and beyond.
\newblock In \emph{Advances in Neural Information Processing Systems 38}, pp.\
  111997--112018. Neural Information Processing Systems Foundation, Inc.
  (NeurIPS), 2025{\natexlab{a}}.
\newblock \doi{10.52202/085713-3380}.
\newblock URL \url{https://openreview.net/forum?id=XtNiw8OQsy}.
\newblock Poster.

\bibitem[Liu et~al.(2025{\natexlab{b}})Liu, Zheng, Muennighoff, Zeng, Dou,
  Pang, Jiang, and Lin]{wu2024regmix}
Qian Liu, Xiaosen Zheng, Niklas Muennighoff, Guangtao Zeng, Longxu Dou, Tianyu
  Pang, Jing Jiang, and Min Lin.
\newblock Regmix: Data mixture as regression for language model pre-training.
\newblock In \emph{International Conference on Learning Representations},
  2025{\natexlab{b}}.
\newblock \doi{10.48550/arXiv.2407.01492}.
\newblock URL
  \url{https://proceedings.iclr.cc/paper_files/paper/2025/hash/5f67d864aae6115374fed7beddd119e0-Abstract-Conference.html}.

\bibitem[Matiisen et~al.(2020)Matiisen, Oliver, Cohen, and
  Schulman]{matiisen2017teacher}
Tambet Matiisen, Avital Oliver, Taco Cohen, and John Schulman.
\newblock Teacher--student curriculum learning.
\newblock \emph{IEEE Transactions on Neural Networks and Learning Systems},
  31\penalty0 (9):\penalty0 3732--3740, 2020.
\newblock \doi{10.1109/TNNLS.2019.2934906}.
\newblock URL \url{https://doi.org/10.1109/TNNLS.2019.2934906}.

\bibitem[McCann(1997)]{mccann1997convexity}
Robert~J. McCann.
\newblock A convexity principle for interacting gases.
\newblock \emph{Advances in Mathematics}, 128\penalty0 (1):\penalty0 153--179,
  1997.
\newblock \doi{10.1006/aima.1997.1634}.

\bibitem[Meng et~al.(2026)Meng, Zeng, Lalor, and Yu]{meng2025psychology}
Guangyu Meng, Qinkai Zeng, John~P. Lalor, and Hong Yu.
\newblock A psychology-based unified dynamic framework for curriculum learning.
\newblock \emph{Computational Linguistics}, pp.\  1--49, 2026.
\newblock \doi{10.1162/COLI.a.584}.
\newblock URL
  \url{https://direct.mit.edu/coli/article/doi/10.1162/COLI.a.584/134522/A-Psychology-based-Unified-Dynamic-Framework-for}.

\bibitem[Peng et~al.(2025)Peng, Zhuang, Qiu, Ma, Yu, Zhu, and
  He]{peng2025topicmixing}
Jiahui Peng, Xinlin Zhuang, Jiantao Qiu, Ren Ma, Jing Yu, He~Zhu, and Conghui
  He.
\newblock Topic over source: The key to effective data mixing for language
  model pre-training.
\newblock \emph{arXiv}, 2025.
\newblock \doi{10.48550/arXiv.2502.16802}.
\newblock URL \url{https://arxiv.org/abs/2502.16802}.

\bibitem[Peyr{\'e} \& Cuturi(2019)Peyr{\'e} and Cuturi]{peyre2019computational}
G.~Peyr{\'e} and Marco Cuturi.
\newblock Computational optimal transport.
\newblock \emph{Found. Trends Mach. Learn.}, 11\penalty0 (5-6):\penalty0
  355--607, 2019.
\newblock \doi{10.1561/2200000073}.

\bibitem[Platanios et~al.(2019)Platanios, Stretcu, Neubig, P{\'o}czos, and
  Mitchell]{platanios2019competence}
Emmanouil~Antonios Platanios, Otilia Stretcu, Graham Neubig, B.~P{\'o}czos, and
  Tom~Michael Mitchell.
\newblock Competence-based curriculum learning for neural machine translation.
\newblock In \emph{North American Chapter of the Association for Computational
  Linguistics}, pp.\  1162--1172. Association for Computational Linguistics,
  2019.
\newblock \doi{10.18653/v1/N19-1119}.
\newblock URL \url{https://aclanthology.org/N19-1119/}.

\bibitem[Santambrogio(2015)]{santambrogio2015otam}
Filippo Santambrogio.
\newblock \emph{Optimal Transport for Applied Mathematicians: Calculus of
  Variations, PDEs, and Modeling}.
\newblock Birkh{\"a}user Cham, 2015.
\newblock \doi{10.1007/978-3-319-20828-2}.

\bibitem[Shrivastava et~al.(2016)Shrivastava, Gupta, and
  Girshick]{shrivastava2016ohem}
Abhinav Shrivastava, Abhinav Gupta, and Ross Girshick.
\newblock Training region-based object detectors with online hard example
  mining.
\newblock In \emph{Computer Vision and Pattern Recognition}, pp.\  761--769.
  IEEE, 2016.
\newblock \doi{10.1109/CVPR.2016.89}.

\bibitem[Swayamdipta et~al.(2020)Swayamdipta, Schwartz, Lourie, Wang,
  Hajishirzi, Smith, and Choi]{swayamdipta2020dataset}
Swabha Swayamdipta, Roy Schwartz, Nicholas Lourie, Yizhong Wang, Hannaneh
  Hajishirzi, Noah~A. Smith, and Yejin Choi.
\newblock Dataset cartography: Mapping and diagnosing datasets with training
  dynamics.
\newblock In \emph{Conference on Empirical Methods in Natural Language
  Processing}, pp.\  9275--9293. Association for Computational Linguistics,
  2020.
\newblock \doi{10.18653/v1/2020.emnlp-main.746}.
\newblock URL \url{https://aclanthology.org/2020.emnlp-main.746/}.

\bibitem[Toneva et~al.(2019)Toneva, Sordoni, Combes, Trischler, Bengio, and
  Gordon]{toneva2019forgetting}
Mariya Toneva, Alessandro Sordoni, R{\'e}mi Tachet~des Combes, Adam Trischler,
  Yoshua Bengio, and Geoffrey~J. Gordon.
\newblock An empirical study of example forgetting during deep neural network
  learning.
\newblock In \emph{International Conference on Learning Representations}, 2019.
\newblock URL \url{https://openreview.net/forum?id=BJlxm30cKm}.

\bibitem[Vaswani et~al.(2017)Vaswani, Shazeer, Parmar, Uszkoreit, Jones, Gomez,
  Kaiser, and Polosukhin]{vaswani2017attention}
Ashish Vaswani, Noam Shazeer, Niki Parmar, Jakob Uszkoreit, Llion Jones,
  Aidan~N. Gomez, Lukasz Kaiser, and Illia Polosukhin.
\newblock Attention is all you need.
\newblock In \emph{Advances in Neural Information Processing Systems},
  volume~30, pp.\  5998--6008. Curran Associates, Inc., 2017.
\newblock URL
  \url{https://proceedings.neurips.cc/paper_files/paper/2017/hash/3f5ee243547dee91fbd053c1c4a845aa-Abstract.html}.

\bibitem[Villani(2008)]{villani2008optimal}
C{\'e}dric Villani.
\newblock \emph{Optimal transport, Old and New}, volume 338.
\newblock Springer Science \& Business Media, 2008.
\newblock ISBN 9783540710493.
\newblock \doi{10.1007/978-3-540-71050-9}.

\bibitem[Wang et~al.(2022)Wang, Chen, and Zhu]{wang2021survey}
Xin Wang, Yudong Chen, and Wenwu Zhu.
\newblock A survey on curriculum learning.
\newblock \emph{IEEE Transactions on Pattern Analysis and Machine
  Intelligence}, 44\penalty0 (9):\penalty0 4555--4576, 2022.
\newblock \doi{10.1109/TPAMI.2021.3069908}.
\newblock URL \url{https://doi.org/10.1109/TPAMI.2021.3069908}.

\bibitem[Wu et~al.(2021)Wu, Dyer, and Neyshabur]{wucurricula2021}
Xiaoxia Wu, Ethan Dyer, and Behnam Neyshabur.
\newblock When do curricula work?
\newblock In \emph{International Conference on Learning Representations}, 2021.
\newblock URL \url{https://openreview.net/forum?id=tW4QEInpni}.

\bibitem[Xie et~al.(2023)Xie, Pham, Dong, Du, Liu, Lu, Liang, Le, Ma, and
  Yu]{xie2023data}
Sang~Michael Xie, Hieu Pham, Xuanyi Dong, Nan Du, Hanxiao Liu, Yifeng Lu, Percy
  Liang, Quoc~V. Le, Tengyu Ma, and Adams~Wei Yu.
\newblock {DoReMi}: Optimizing data mixtures speeds up language model
  pretraining.
\newblock In \emph{Advances in Neural Information Processing Systems 36}, pp.\
  69798--69818. Neural Information Processing Systems Foundation, Inc.
  (NeurIPS), 2023.
\newblock \doi{10.52202/075280-3059}.
\newblock URL
  \url{https://proceedings.neurips.cc/paper_files/paper/2023/hash/dcba6be91359358c2355cd920da3fcbd-Abstract-Conference.html}.

\bibitem[Ye et~al.(2025)Ye, Liu, Sun, Zhan, Zhou, and
  Qiu]{abnar2024datamixinglaws}
Jiasheng Ye, Peiju Liu, Tianxiang Sun, Jun Zhan, Yunhua Zhou, and Xipeng Qiu.
\newblock Data mixing laws: Optimizing data mixtures by predicting language
  modeling performance.
\newblock In \emph{International Conference on Learning Representations}, 2025.
\newblock \doi{10.48550/arXiv.2403.16952}.
\newblock URL
  \url{https://proceedings.iclr.cc/paper_files/paper/2025/hash/cc84bfabe6389d8883fc2071c848f62a-Abstract-Conference.html}.

\end{thebibliography}
\bibliographystyle{colm2026_conference}

\appendix
\section{Extended Related Work}\label{app:extended_related_work}

\paragraph{Curriculum learning and difficulty-aware training.} Classical curriculum learning moves from easy to hard and can stabilize optimization and improve generalization \citep{bengio2009curriculum}. Automated curriculum-learning methods choose what data to present and when: self-paced learning matches example difficulty to learner competence \citep{kumar2010selfpaced}; teacher--student curricula prioritize items with high learning progress \citep{matiisen2017teacher}; bandit-style schedulers adapt task proportions online \citep{graves2017automated}; and competence-based schedules control how quickly harder examples are introduced \citep{platanios2019competence}. Evidence suggests that pacing strongly affects outcomes \citep{hacohen2019power}, while training-dynamics diagnostics provide signals for difficulty and ambiguity \citep{swayamdipta2020dataset}; related heuristics include hard-example mining and forgetting-based selection \citep{shrivastava2016ohem,toneva2019forgetting}. Our framework instead represents a curriculum as a path over distributions on discrete difficulty levels and separates the path itself from its pacing. This makes it possible to move probability mass smoothly through nearby levels while controlling where the curriculum travels and how quickly it advances.

\paragraph{Optimal transport.}
Optimal transport measures distances between probability distributions via the minimal cost of moving mass; in this metric, Wasserstein (displacement) interpolation follows geodesics connecting two endpoint distributions \citep{mccann1997convexity,peyre2019computational,santambrogio2015otam}. 
OT has also been used to \emph{construct curricula} in reinforcement learning by generating geodesic sequences or constrained OT plans between task distributions \citep{huang2022crlot,klink2022cot}. 
Our approach is related in that it also uses Wasserstein geodesics for curriculum learning, but the emphasis here is empirical control rather than a new RL-specific scheduler. We use Wasserstein paths as a common scaffold for comparing endpoint choice, smoothness, traversal speed, and adaptive geometry on calibrated synthetic tasks and language-model reasoning benchmarks. This keeps the distributional formulation fixed while studying which curriculum-design choices matter.

\paragraph{Data Mixing.} 
Designing training distributions from heterogeneous sources is an important part of large-scale model training. 
Static approaches fix mixture weights in advance, e.g., DoReMi \citep{xie2023data} optimizes domain sampling via a proxy model, and DoGE \citep{fan2024domainmixing} estimates domain importance from small runs. 
Recent work also predicts or balances effective mixtures from proxy runs and learned signals: RegMix regresses performance from proxy runs \citep{wu2024regmix}, data mixing laws model weight--loss functions \citep{abnar2024datamixinglaws}, mixtures of data experts approximate source contributions \citep{liu2025mde}, and regroup-and-balance strategies adapt mixtures during training \citep{huang2025rb}. 
Other work goes beyond source-level mixing, such as topic-based mixtures \citep{peng2025topicmixing}. Our work shares the perspective of viewing curricula as data mixing, but differs in two ways: (i) we emphasize \emph{distributional paths} rather than static or adaptive reweighting, and (ii) we study principled design factors such as endpoint choice and pacing, complementing existing online weighting methods.

\section{Detailed Experimental Setup}\label{app:details}
We include detailed experiment setups and data examples.

\subsection{Task and Dataset Details}\label{app:task_details}
Each difficulty axis below is described using the writing terms from our task-to-difficulty-axis mapping. When one difficulty axis varies, the remaining task parameters stay at the dataset defaults stated in the same paragraph.

\subsubsection{k-Parity Dataset}\label{app:kparity_dataset_construction}

\paragraph{Task.}
Each example is a binary string \(x\in\{0,1\}^n\) with label
\[
y=f_k(x)=\bigoplus_{i=1}^k x_i.
\]
Only the leading bits matter; the remaining bits are nuisance features. We report classification accuracy.

\paragraph{Construction and difficulty levels.}
We study two difficulty axes, each with 5 levels:
\begin{itemize}[leftmargin=*,nosep]
  \item \textbf{Informative-bit difficulty axis:} levels use \(\{1,2,3,4,5\}\) informative prefix bits while keeping the total string length fixed at 32.
  \item \textbf{Length difficulty axis:} levels use total string lengths \(\{24,28,32,36,40\}\) while keeping the number of informative prefix bits fixed at 3.
\end{itemize}
For the informative-bit difficulty axis, later levels strictly contain earlier ones: before the final level, the later informative positions are fixed to 0 and are released only when that level is introduced. We therefore also evaluate each level on a unique slice where the newly introduced informative bit is 1 and all later informative positions remain 0, so that each test set isolates the new bit introduced at that level.

\paragraph{Example.}
\begin{quote}\raggedright
\textit{Input.} \texttt{01100110001000100001010101010110}\\
\textit{Target output.} \texttt{0}
\end{quote}
In this string, the first three bits are \(0,1,1\), so the correct label is \(0\oplus 1\oplus 1=0\). A model that still behaves like the previous level, using only the first two bits, would predict \(0\oplus 1=1\) instead. This is exactly the failure that the unique-slice evaluation is designed to expose.

\paragraph{Dataset size.}
We use batch size 1000. For the informative-bit difficulty axis, the small, medium, and large budgets are 100, 200, and 300 steps. For the length difficulty axis, they are 5000, 7500, and 10000 steps.

\subsubsection{Dyck Language Dataset}  \label{app:dyck_dataset_construction} 

\paragraph{Task.}
Each example is a prefix-completion problem for the Dyck language. We sample a valid bracket string, reveal an initial prefix, and ask the model to output the full completed sequence, not just the missing tail. We report exact-match accuracy on the completed string.

\paragraph{Construction and difficulty levels.}
We vary one factor at a time:
\begin{itemize}[leftmargin=*,nosep]
  \item \textbf{Bracket-type difficulty axis:} 4 levels with \(\{1,2,3,4\}\) bracket types, while total sequence length stays fixed at 20 and hidden suffix length stays fixed at 3.
  \item \textbf{Length difficulty axis:} 4 levels with total sequence lengths \(\{14,16,18,20\}\), while the number of bracket types stays fixed at 4 and the hidden suffix length stays fixed at 3.
  \item \textbf{Missing-suffix difficulty axis:} 4 levels with hidden suffix lengths \(\{3,5,7,9\}\), while the number of bracket types stays fixed at 4 and the total sequence length stays fixed at 20.
\end{itemize}

\paragraph{Example.}
\begin{quote}\raggedright
\textit{Input prefix.} \texttt{[[()\{\}]][]()()[<<}\\
\textit{Target output.} \texttt{[[()\{\}]][]()()[<<>>]}
\end{quote}
The missing part is short, but the model still has to track the open brackets that remain unresolved and finish the entire balanced string correctly.

\paragraph{Dataset size.}
We use batch size 100. For the bracket-type difficulty axis, the small, medium, and large budgets are 200, 350, and 1000 steps; for the length difficulty axis they are 200, 300, and 1000; and for the missing-suffix difficulty axis they are 300, 500, and 1000.

\subsubsection{Survo Dataset} \label{app:survo_dataset_construction}

\paragraph{Task.}
Survo asks the model to fill missing entries in a partially observed \(4\times 4\) grid whose last row and last column give row and column sums. The unknown cells always lie in the upper-left \(3\times 3\) block, and we report exact-match accuracy on the completed grid.

\paragraph{Construction and difficulty levels.}
We use two difficulty axes:
\begin{itemize}[leftmargin=*,nosep]
  \item \textbf{Blank-count difficulty axis:} 5 levels with \(\{1,2,3,4,5\}\) blanks, while the grid size stays fixed at \(4\times 4\) and the allowed cell range stays fixed to 1--9.
  \item \textbf{Value-range difficulty axis:} 4 levels with maximum allowed cell values \(\{6,9,12,15\}\), while the grid size stays fixed at \(4\times 4\), the number of blanks stays fixed at 3, and the minimum cell value stays fixed at 1.
\end{itemize}

\paragraph{Example.}
\begin{quote}\raggedright
\textit{Input.} \texttt{[[X, 1, 2, 9], [6, X, 1, 10], [4, 2, X, 8], [16, 6, 5, 27]]}\\
\textit{Target output.} \texttt{[[6, 1, 2, 9], [6, 3, 1, 10], [4, 2, 2, 8], [16, 6, 5, 27]]}
\end{quote}
The first row must begin with 6 because \(6+1+2=9\); the second row must place 3 in the middle because \(6+3+1=10\); and the third row then takes 2.

\paragraph{Dataset size.}
We use batch size 100. For both difficulty axes, the small, medium, and large budgets are 7500, 15000, and 25000 steps.

\subsubsection{Dyck Language Errors Dataset}\label{app:dyck_errors_dataset_construction}

\paragraph{Task.}
This task presents a bracket string and asks for the first position at which it becomes invalid, using 1-indexing. Errors include an unmatched closing bracket, a closing bracket of the wrong type, or a prefix that is locally valid but still unfinished at the end; in the last case the answer is one position past the end of the string. Valid strings receive \(-1\). We report exact-match accuracy.

\paragraph{Construction and difficulty levels.}
We use two difficulty axes, each with 4 levels:
\begin{itemize}[leftmargin=*,nosep]
  \item \textbf{Bracket-type difficulty axis:} levels use \(\{1,2,3,4\}\) bracket types while keeping the total string length fixed at 20.
  \item \textbf{Length difficulty axis:} levels use total string lengths \(\{10,15,20,25\}\) while keeping the number of bracket types fixed at 3.
\end{itemize}

\paragraph{Example.}
\begin{quote}\raggedright
\textit{Input.} \texttt{[](()))([()][[]((()(}\\
\textit{Target output.} \texttt{7}
\end{quote}
The first six symbols can still be parsed consistently, but the seventh symbol is an extra closing parenthesis with nothing left to match.

\paragraph{Dataset size.}
We use batch size 100. For the bracket-type difficulty axis, we use 8{,}000 training examples and 1{,}000 evaluation examples per level, and the small, medium, and large budgets are 300, 600, and 1{,}000 steps. For the length difficulty axis, we keep the original setup with 2{,}400 training examples and 2{,}000 evaluation examples per level, and the small, medium, and large budgets are 750, 1500, and 3000.

\subsubsection{Calendar Scheduling Dataset}\label{app:calendar_scheduling_dataset_construction}

\paragraph{Task.}
Each example specifies task durations, precedence constraints, and a finite time horizon for a single-machine schedule with no overlap. The solver always picks the alphabetically earliest available task and starts it immediately after the previous task ends. The model must output the full schedule as \texttt{Task:start-end} pairs, and we report exact-match accuracy on that string.

\paragraph{Construction and difficulty levels.}
We vary three factors separately:
\begin{itemize}[leftmargin=*,nosep]
  \item \textbf{Task-count difficulty axis:} 4 levels with \(\{4,6,8,10\}\) tasks, while the time horizon stays fixed at 20, the number of precedence constraints stays fixed at 4, and the maximum task duration stays fixed at 3.
  \item \textbf{Horizon difficulty axis:} 4 levels with time horizons \(\{12,16,20,24\}\), while the number of tasks stays fixed at 6, the number of precedence constraints stays fixed at 4, and the maximum task duration stays fixed at 3.
  \item \textbf{Dependency difficulty axis:} 4 levels with \(\{2,4,6,8\}\) precedence constraints, while the number of tasks stays fixed at 6, the time horizon stays fixed at 20, and the maximum task duration stays fixed at 3.
\end{itemize}

\paragraph{Example.}
\begin{quote}\raggedright
\textit{Input.} \texttt{Time horizon 0--19; durations A=1 B=1 C=1 D=2 E=2 F=1; dependencies B->D, B->E, A->D, E->F}\\
\textit{Target output.} \texttt{A:0-1 B:1-2 C:2-3 D:3-5 E:5-7 F:7-8}
\end{quote}
Tasks \(A\), \(B\), and \(C\) are initially available, so the alphabetical rule places them first. Task \(D\) must wait for both \(A\) and \(B\), and task \(F\) must wait for \(E\).

\paragraph{Dataset size.}
We use batch size 100. For the task-count difficulty axis, the small, medium, and large budgets are 2000, 3000, and 4500 steps; for the horizon difficulty axis they are 1400, 1700, and 3000; and for the dependency difficulty axis they are 2000, 3000, and 4200.

\subsubsection{Game of 24 Dataset}\label{app:game_of_24_dataset_construction}

\paragraph{Task.}
Each instance gives four integers and a target value. The model must produce an arithmetic expression that uses each number exactly once and combines them with the standard binary operations to reach the target. Because many expressions can be correct, we report verifier-based success rate rather than exact string match.

\paragraph{Construction and difficulty levels.}
We use two difficulty axes, each with 5 levels:
\begin{itemize}[leftmargin=*,nosep]
  \item \textbf{Target-value difficulty axis:} levels use target values \(\{18,21,24,27,30\}\), while the number of input numbers stays fixed at four, the allowed operations stay fixed to \(+,-,\times,\div\), and the input-number range stays fixed at 1--15.
  \item \textbf{Number-range difficulty axis:} levels use input-number ranges \(\{1\text{--}12,1\text{--}15,1\text{--}18,1\text{--}21,1\text{--}24\}\), while the target value stays fixed at 24 and the number of input numbers stays fixed at four.
\end{itemize}
The second change creates a broader and less repetitive arithmetic search space.

\paragraph{Example.}
\begin{quote}\raggedright
\textit{Input.} \texttt{Numbers: 3, 11, 4, 8; target: 24}\\
\textit{Target output.} \texttt{4*((3+11)-8)}
\end{quote}
Any verifier-equivalent expression is counted as correct; this is one valid target output.

\paragraph{Dataset size.}
We use batch size 100. For the target-value difficulty axis, the small, medium, and large budgets are 4200, 5000, and 9500 steps. For the number-range difficulty axis, they are 2500, 5000, and 9000.

\subsubsection{Goods Exchange Dataset}\label{app:goods_exchange_dataset_construction}

\paragraph{Task.}
Each example starts from a one-to-one ownership map between people and objects, followed by a short sequence of exchange statements. The model must return the final ownership map, sorted alphabetically and formatted one line per person as \texttt{Person: item}. We report exact-match accuracy after canonicalizing outputs to that format.

\paragraph{Construction and difficulty levels.}
We use two difficulty axes, each with 4 levels:
\begin{itemize}[leftmargin=*,nosep]
  \item \textbf{People-count difficulty axis:} levels use \(\{3,4,5,6\}\) people, and therefore the same number of items, while keeping the number of exchange statements fixed at 2.
  \item \textbf{Exchange-length difficulty axis:} levels use \(\{2,3,4,5\}\) exchange statements while keeping the number of people fixed at 3.
\end{itemize}
The first broadens the state being tracked; the second lengthens the chain of ownership updates.

\paragraph{Example.}
\begin{quote}\raggedright
\textit{Input.} \texttt{Initial ownership: Sophia -> coral kettle; Betty -> beige kettle; Donna -> beige watch.}\\
\texttt{Statements: Betty asked to exchange with Donna, but Donna refused; Sophia and Donna exchanged items;}\\
\texttt{Sophia and Betty exchanged items.}\\
\textit{Target output.} \texttt{Betty: beige watch}\\
\texttt{Donna: coral kettle}\\
\texttt{Sophia: beige kettle}
\end{quote}
Here the first statement is a decoy: Donna refuses Betty's request, so ownership does not change. The next two exchanges do the real work. After Sophia and Donna exchange items, Donna holds the coral kettle and Sophia holds the beige watch; after Sophia and Betty exchange items, Betty ends with the beige watch and Sophia with the beige kettle.

\paragraph{Dataset size.}
We use batch size 100. For the people-count difficulty axis, the small, medium, and large budgets are 500, 1000, and 1500 steps. For the exchange-length difficulty axis, they are 1500, 2250, and 3000.

\subsubsection{Index Select Dataset}\label{app:index_select_dataset_construction}

\paragraph{Task.}
The input is a token sequence together with a list of inclusive 0-indexed ranges. The model must concatenate the selected spans in order and output the resulting token list, space-separated. We report exact-match accuracy.

\paragraph{Construction and difficulty levels.}
We vary four factors separately:
\begin{itemize}[leftmargin=*,nosep]
  \item \textbf{Source-length difficulty axis:} 5 levels with source sequence lengths \(\{10,12,14,16,18\}\), while selected-output length stays fixed at 8, the number of ranges stays fixed at 3, and the alphabet size stays fixed at 4.
  \item \textbf{Selected-output-length difficulty axis:} 5 levels with \(\{4,6,8,10,12\}\) selected tokens, while the source sequence length stays fixed at 16, the number of ranges stays fixed at 3, and the alphabet size stays fixed at 4.
  \item \textbf{Range-count difficulty axis:} 5 levels with \(\{1,2,3,4,5\}\) ranges, while the source sequence length stays fixed at 16, the selected-output length stays fixed at 8, and the alphabet size stays fixed at 4.
  \item \textbf{Alphabet-size difficulty axis:} 5 levels with alphabet sizes \(\{2,3,4,5,6\}\), while the source sequence length stays fixed at 16, the selected-output length stays fixed at 8, and the number of ranges stays fixed at 3.
\end{itemize}
These changes make the copy operation longer, more fragmented, or less repetitive.

\paragraph{Example.}
\begin{quote}\raggedright
\textit{Input.} \texttt{Sequence: d a d a a b a b a d d a b c a c; ranges: 0-1, 3-5, 10-12}\\
\textit{Target output.} \texttt{d a a a b d a b}
\end{quote}
The range \texttt{0-1} contributes \texttt{d a}, the range \texttt{3-5} contributes \texttt{a a b}, and the range \texttt{10-12} contributes \texttt{d a b}.

\paragraph{Dataset size.}
We use batch size 100. For source length and selected-output length, the small, medium, and large budgets are 3000, 4000, and 6000 steps. For the number of ranges they are 3600, 4000, and 6400, and for alphabet size they are 2500, 3400, and 6000.

\subsubsection{Sequence Reverse Dataset}\label{app:sequence_reverse_dataset_construction}

\paragraph{Task.}
This is the standard sequence-reversal task: given a token sequence, output the same sequence in reverse order. We report exact-match accuracy on the reversed sequence.

\paragraph{Construction and difficulty levels.}
We use two difficulty axes, each with 5 levels:
\begin{itemize}[leftmargin=*,nosep]
  \item \textbf{Length difficulty axis:} levels use sequence lengths \(\{6,8,10,12,14\}\) while keeping the alphabet size fixed at 4.
  \item \textbf{Alphabet-size difficulty axis:} levels use alphabet sizes \(\{2,3,4,5,6\}\) while keeping the sequence length fixed at 12.
\end{itemize}

\paragraph{Example.}
\begin{quote}\raggedright
\textit{Input.} \texttt{d b c b b c b b b c a a}\\
\textit{Target output.} \texttt{a a c b b b c b b c b d}
\end{quote}

\paragraph{Dataset size.}
We use batch size 100. For both difficulty axes, the small, medium, and large budgets are 600, 700, and 800 steps.

\subsubsection{Run-Length Encoding Dataset}\label{app:run_length_encoding_dataset_construction}

\paragraph{Task.}
Each example contains a token sequence that must be converted to run-length form as space-separated \texttt{<count> <symbol>} pairs. The task is deterministic, and we report exact-match accuracy on the encoded string.

\paragraph{Construction and difficulty levels.}
We vary four properties separately:
\begin{itemize}[leftmargin=*,nosep]
  \item \textbf{Length difficulty axis:} 5 levels with sequence lengths \(\{6,7,8,9,10\}\), while the maximum number of runs stays fixed at 8, the alphabet size stays fixed at 4, and the maximum run length stays fixed at 4.
  \item \textbf{Run-count difficulty axis:} 5 levels with maximum numbers of runs \(\{4,5,6,7,8\}\), while the sequence length stays fixed at 12, the alphabet size stays fixed at 4, and the maximum run length stays fixed at 4.
  \item \textbf{Alphabet-size difficulty axis:} 4 levels with alphabet sizes \(\{2,3,4,5\}\), while the sequence length stays fixed at 12, the maximum number of runs stays fixed at 8, and the maximum run length stays fixed at 4.
  \item \textbf{Run-length difficulty axis:} 4 levels with maximum run lengths \(\{2,3,4,5\}\), while the sequence length stays fixed at 12, the maximum number of runs stays fixed at 8, and the alphabet size stays fixed at 4.
\end{itemize}
These difficulty axes make the compressed description longer, more segmented, or more varied.

\paragraph{Example.}
\begin{quote}\raggedright
\textit{Input.} \texttt{c a a a d d d d b b b b}\\
\textit{Target output.} \texttt{1 c 3 a 4 d 4 b}
\end{quote}
The sequence contains one \texttt{c}, then three \texttt{a}'s, then four \texttt{d}'s, and finally four \texttt{b}'s.

\paragraph{Dataset size.}
We use batch size 100. The small, medium, and large budgets are 1000, 1500, and 2200 steps when varying sequence length; 1000, 1350, and 2100 when varying the number of runs; 1000, 1500, and 2100 when varying alphabet size; and 1300, 1500, and 2200 when varying maximum run length.

\subsubsection{Stack Operations Dataset}\label{app:stack_operations_dataset_construction}

\paragraph{Task.}
This task simulates a LIFO stack over a finite alphabet. Inputs are sequences of \texttt{PUSH} and \texttt{POP} operations, and the model must output the final stack from top to bottom, or \texttt{EMPTY} if no items remain. We report exact-match accuracy.

\paragraph{Construction and difficulty levels.}
We vary four factors separately:
\begin{itemize}[leftmargin=*,nosep]
  \item \textbf{Operation-count difficulty axis:} 4 levels with \(\{6,8,10,12\}\) operations, while the alphabet size stays fixed at 4 and the pop probability stays fixed at 0.4; for this difficulty axis, the stack-depth cap scales with the operation count, with a minimum cap of 4.
  \item \textbf{Alphabet-size difficulty axis:} 4 levels with alphabet sizes \(\{2,3,4,5\}\), while the number of operations stays fixed at 8, the maximum stack depth stays fixed at 6, and the pop probability stays fixed at 0.4.
  \item \textbf{Maximum-depth difficulty axis:} 4 levels with maximum stack depths \(\{2,3,4,5\}\), while the number of operations stays fixed at 8, the alphabet size stays fixed at 4, and the pop probability stays fixed at 0.4.
  \item \textbf{Pop-rate difficulty axis:} 4 levels with pop probabilities \(\{0.65,0.5,0.35,0.2\}\), listed from easy to hard, while the number of operations stays fixed at 8, the alphabet size stays fixed at 4, and the maximum stack depth stays fixed at 6.
\end{itemize}
Later levels therefore produce longer traces and, in the last difficulty axis, more live stack contents to remember.

\paragraph{Example.}
\begin{quote}\raggedright
\textit{Input.} \texttt{PUSH a; PUSH a; POP; POP; PUSH d; POP; PUSH d; PUSH d}\\
\textit{Target output.} \texttt{d d}
\end{quote}
The first two pushes are canceled by the next two pops, the next \texttt{PUSH d} is immediately removed, and the last two pushes remain.

\paragraph{Dataset size.}
We use batch size 100. The small, medium, and large budgets are 300, 380, and 550 steps when varying the number of operations; 280, 420, and 600 when varying alphabet size; 180, 220, and 280 when varying maximum depth; and 280, 360, and 540 when varying the pop rate.

\subsubsection{Queue Operations Dataset}\label{app:queue_operations_dataset_construction}

\paragraph{Task.}
This task simulates a FIFO queue. Inputs are \texttt{ENQUEUE} and \texttt{DEQUEUE} operations, and the model must output the final queue from front to back, or \texttt{EMPTY} if the queue is empty. We report exact-match accuracy.

\paragraph{Construction and difficulty levels.}
We report three difficulty axes for this task:
\begin{itemize}[leftmargin=*,nosep]
  \item \textbf{Operation-count difficulty axis:} 5 levels with \(\{12,16,20,24,28\}\) operations while keeping the other task parameters fixed at their default values.
  \item \textbf{Alphabet-size difficulty axis:} 4 levels with alphabet sizes \(\{4,6,8,10\}\) while keeping the remaining parameters fixed.
  \item \textbf{Queue-capacity difficulty axis:} 4 levels with capacities \(\{2,3,4,5\}\), listed from easy to hard, while keeping the remaining parameters fixed.
\end{itemize}

\paragraph{Example.}
\begin{quote}\raggedright
\textit{Input.} \texttt{ENQUEUE b; ENQUEUE d; DEQUEUE; DEQUEUE; ENQUEUE b; DEQUEUE; ENQUEUE c; ENQUEUE a; ENQUEUE d; ENQUEUE d; DEQUEUE; DEQUEUE}\\
\textit{Target output.} \texttt{d d}
\end{quote}
The first three dequeues remove \texttt{b}, then \texttt{d}, then the later \texttt{b}. The remaining front-to-back queue is \texttt{d d}.

\paragraph{Dataset size.}
We use batch size 100. The small, medium, and large budgets are 450, 570, and 720 steps when varying the number of operations; 300, 400, and 640 when varying alphabet size; and 280, 390, and 550 when varying the queue-cap parameter.
\subsection{Architecture and Training Details}\label{app:model_training_details}
All models were trained on a single NVIDIA A100 GPU.

\paragraph{Scratch transformer tasks}
Most tasks in the current synthetic sweep, including \emph{k}-Parity, Dyck Language, Dyck Language Errors, Sequence Reverse, Index Select, Run-Length Encoding, Stack Operations, Queue Operations, Survo, and Calendar Scheduling, are trained from scratch as decoder-only transformers. Most use hidden size 256, 4 layers, and 8 attention heads with AdamW, learning rate \(5\times10^{-5}\), weight decay \(0.1\), gradient clipping at \(1.0\), and a constant learning-rate schedule with no warmup. The main architectural exceptions are \emph{k}-Parity, which uses a smaller hidden size 128 / 1-layer / 4-head model and learning rate \(5\times10^{-4}\) with batch size 1{,}000; Dyck Language, which uses 3 layers; Survo, which uses 6 layers; and Calendar Scheduling, which uses a larger 6-layer model with hidden size 384 and 12 attention heads. Other scratch tasks typically use train batch size 100, with task-specific evaluation batch sizes and calibrated small/medium/large budgets given in the corresponding dataset subsections and task defaults.

\paragraph{Pretrained language-like tasks.} \textsc{Game of 24} and \textsc{Goods Exchange} use \texttt{HuggingFaceTB/SmolLM-135M} \citep{allal2024SmolLM}. We fine-tune these
models with AdamW using learning rate \(5\times10^{-5}\), weight decay \(0.1\), gradient clipping at \(1.0\), and a constant learning-rate schedule without warmup.

\subsection{Level-Wise Cumulative Exposure}\label{app:default_geometry_distributions}
\begin{figure}[h]
  \centering
  \includegraphics[width=\linewidth]{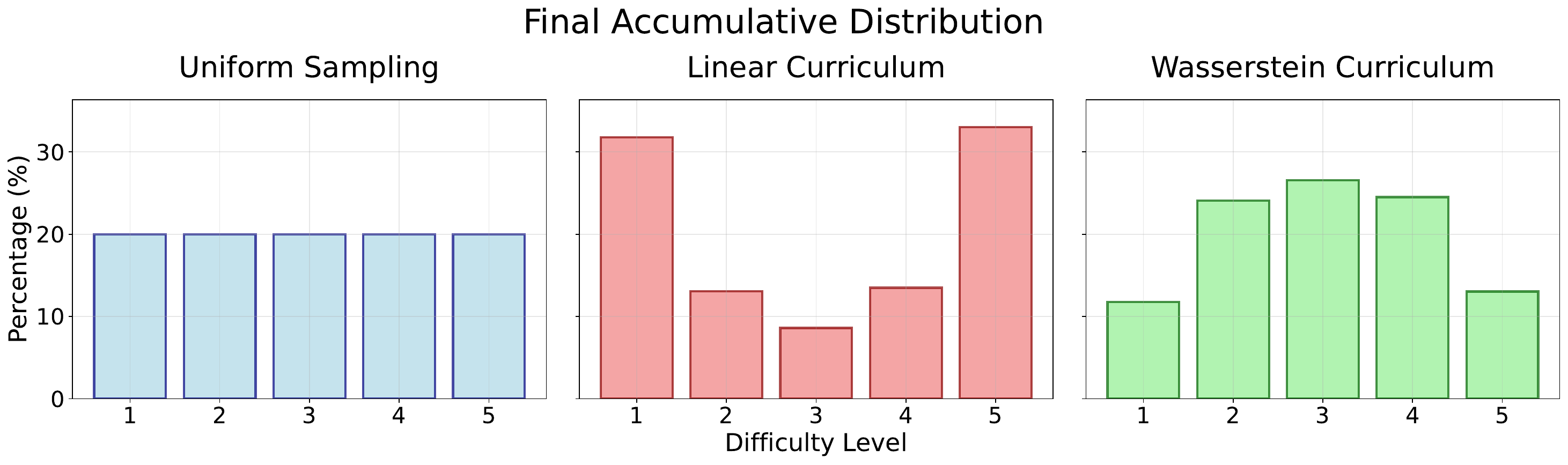}
  \caption{Cumulative level-wise exposure under different curricula.
\textbf{Left:} uniform sampling; \textbf{middle:} linear interpolation;
\textbf{right:} Wasserstein interpolation.}
  \label{fig:curriculum_final_distribution}
\end{figure}

Figure~\ref{fig:curriculum_final_distribution} shows the cumulative level-wise exposure induced over the full training trajectory with \(\gamma=1\) and \(\tau=1\). Uniform sampling gives equal exposure to all levels, linear interpolation allocates more exposure to the boundary levels, and Wasserstein interpolation allocates more exposure to the intermediate levels.
In the default one-dimensional setting, this Wasserstein path uses the path metric \(d(\ell,\ell') = |\ell-\ell'|\), i.e., the standard \(W_1\) geometry on ordered levels.
This illustrates that even when different curricula converge to similar overall performance, the Wasserstein curriculum achieves better sample efficiency at the higher levels.

\subsection{Adaptive Geometry in \SYSNAME{}} \label{app:warp_definition}
Section~\ref{sec:warp_geometry} uses a simple adaptive geometry on an ordered set of difficulty levels. This can be viewed as a discrete, gradient-informed geometry on the level axis. In the default setting, adjacent levels are equally spaced; \SYSNAME{} instead learns nonuniform edge lengths from local gradient alignment, yielding a discrete analogue of a one-dimensional Riemannian metric. This is loosely related to information geometry, where gradient information is used to define local geometry through the Fisher information matrix \citep{amari2016information}. Here, we use cosine similarity between neighboring level gradients as a simpler measure of local similarity. Let \(P_0,P_1 \in \Delta^{m-1}\) denote the easy and hard endpoint distributions, let \(T\) be the total number of training steps, and let \(T_w=\lfloor \rho T \rfloor\) be a short warmup, with \(\rho=0.05\) in our experiments.

During warmup, \SYSNAME{} uses uniform sampling:
\[
P^k_{\text{\SYSNAME}}=\mathrm{Unif}(\{1,\dots,m\}), \qquad k < T_w.
\]
At the end of warmup, it probes each level to obtain gradient vectors \(g_1,\dots,g_m\) and forms the cosine-coupling matrix
\[
C_{ij}=\frac{\langle g_i,g_j\rangle}{\lVert g_i\rVert\,\lVert g_j\rVert}.
\]
It then converts adjacent couplings into edge lengths
\[
\Delta_\ell = 2-\mathrm{clip}(C_{\ell,\ell+1},0,1), \qquad \ell=1,\dots,m-1,
\]
defines locations \(x_1=0\) and \(x_{\ell+1}=x_\ell+\Delta_\ell\), and follows the same forward Wasserstein geodesic on the learned locations \(x=(x_1,\dots,x_m)\):
\[
\tau_k=\left(\frac{k-T_w}{T-T_w-1}\right)^\gamma,\qquad
P^k_{\text{\SYSNAME}}=\mathcal{G}_{x}(\tau_k;P_0,P_1), \qquad k \ge T_w,
\]
where \(\mathcal{G}_{x}\) denotes Wasserstein displacement interpolation on locations \(x\). Thus \SYSNAME{} changes geometry, not endpoints or direction; pacing changes as a consequence. For visualization, Figure~\ref{fig:warp_geometry_motivation} rescales the learned locations affinely to a fixed display span; this does not change the relative geometry or the induced pacing pattern.

\subsection{Details for Structured Difficulty Spaces}\label{app:nonchain_geometry_details}
Section~\ref{sec:nonchain_geometry} uses two structured variants of the same arithmetic-operator family: an 8-node cube and a 7-node tree. Figure~\ref{fig:nonchain_geometry_graphs} visualizes these two benchmark structures. In the cube, each node adds a subset of three operator modifications---multiplication, parentheses, and negation---so the graph follows the natural Boolean-cube structure from the simplest add/sub benchmark to the all-three endpoint. In the tree, the root splits first into multiplication and parentheses branches, and each branch then refines into longer-expression and negation leaves. In both cases, nodes are grouped into \textit{easy}, \textit{mid}, and \textit{hard} regions; for the cube these correspond to \(L1\), \(L2\)--\(L7\), and \(L8\), while for the tree they correspond to the root, internal branch nodes, and leaves.

For the cube, the Wasserstein endpoints over \(L1\)–\(L8\) are \((0.39, 0.14, 0.14, 0.14, 0.05, 0.05, 0.05, 0.02)\) and \((0.02, 0.05, 0.05, 0.05, 0.14, 0.14, 0.14, 0.39)\). For the tree, the endpoints over \(L1\)–\(L7\) are \((0.44, 0.16, 0.16, 0.06, 0.06, 0.06, 0.06)\) and \((0.03, 0.08, 0.08, 0.21, 0.21, 0.21, 0.21)\). The main comparison on structured difficulty spaces reports ten seeds for \textit{static i.i.d.}, \textit{Wasserstein}, and \SYSNAME{}.

\begin{figure*}[t]
  \centering
  \includegraphics[width=0.98\textwidth]{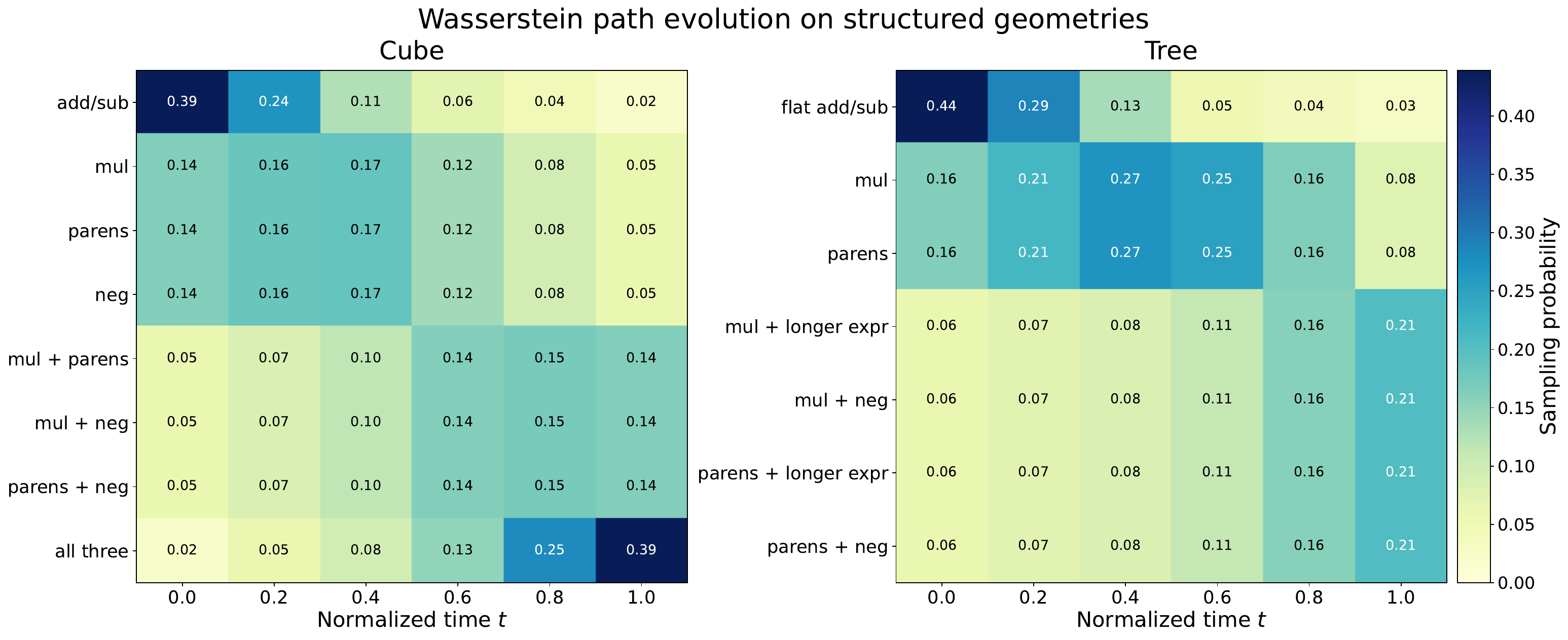}
  \caption{Wasserstein path evolution on the cube and tree structures used in Section~\ref{sec:nonchain_geometry}. Each column corresponds to normalized training progress \(t \in \{0.0, 0.2, \dots, 1.0\}\), and each cell shows the sampling probability assigned to that node along the graph-Wasserstein path.}
  \label{fig:appendix_nonchain_wasserstein_path}
\end{figure*}

\section{Detailed Results and Mechanism Analyses}\label{app:taskwise_curriculum_comparisons}

\subsection{Task-Wise Final-Accuracy Barplots}\label{appsubsec:taskwise_final_accuracy_barplots}

\paragraph{Setup.}
To complement Section~\ref{subsec:when_do_curricula_help}, we report final overall accuracy separately for each task and difficulty axis. Each figure corresponds to one task. Within a figure, panels correspond to the task's difficulty axes, the x-axis groups the small, medium, and large budgets, and colored bars compare the eight curricula: \textit{static i.i.d.}, linear, Wasserstein, \SYSNAME{}, the reverse variants, and the two static-matching baselines. Bars show the mean over 10 random-seed repetitions, and error bars show one standard deviation.

\paragraph{Findings.}
These plots expose the heterogeneity behind Table~\ref{tab:basic_curriculum_summary}: the ranking shifts across tasks, difficulty axes, and budgets, rather than collapsing to a single dominant curriculum.

\newcommand{\taskaccuracyfigure}[3]{%
\begin{center}
  \begin{minipage}{\textwidth}
    \centering
    \includegraphics[width=0.98\textwidth]{figures/#1}
    \captionof{figure}{Task-wise final overall accuracy for #2. Panel titles denote the difficulty axes. Bars show mean final overall accuracy over 10 random-seed repetitions, and error bars show one standard deviation.}
    \label{#3}
  \end{minipage}
\end{center}
\vspace{0.5em}
}

\taskaccuracyfigure{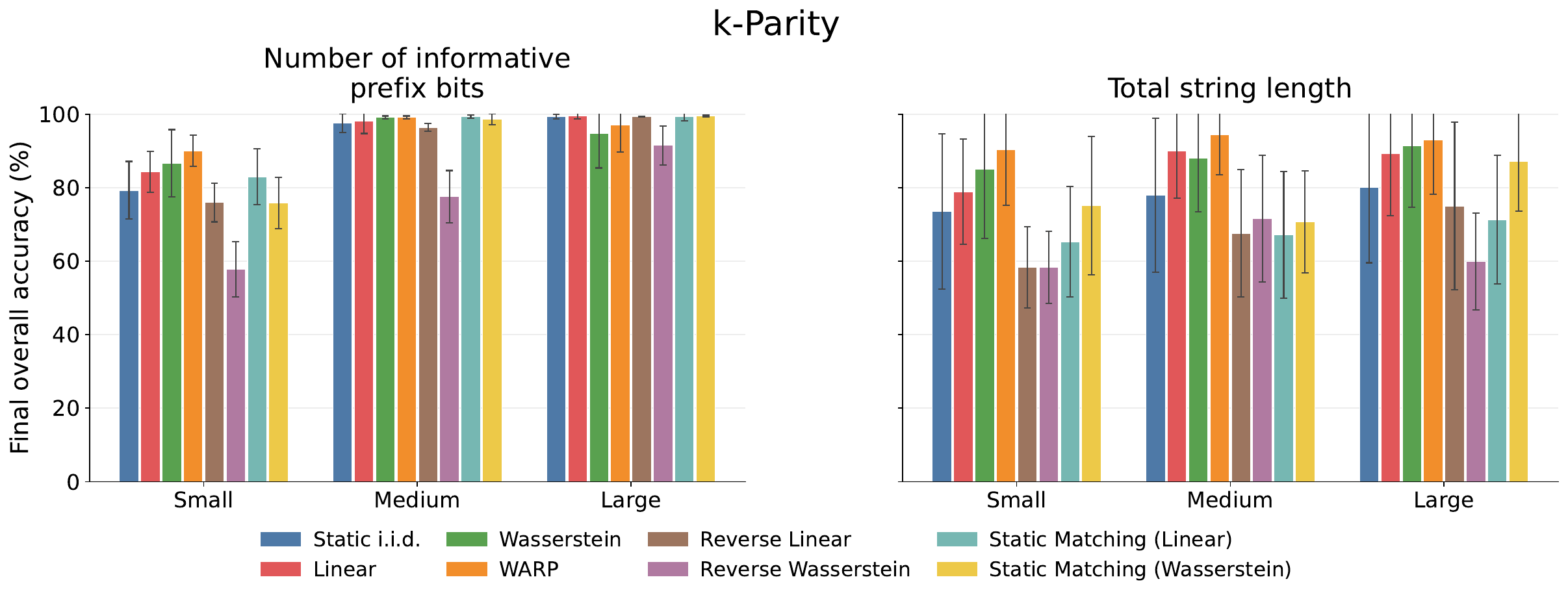}{K-Parity}{fig:appendix_b1_k_parity}
\taskaccuracyfigure{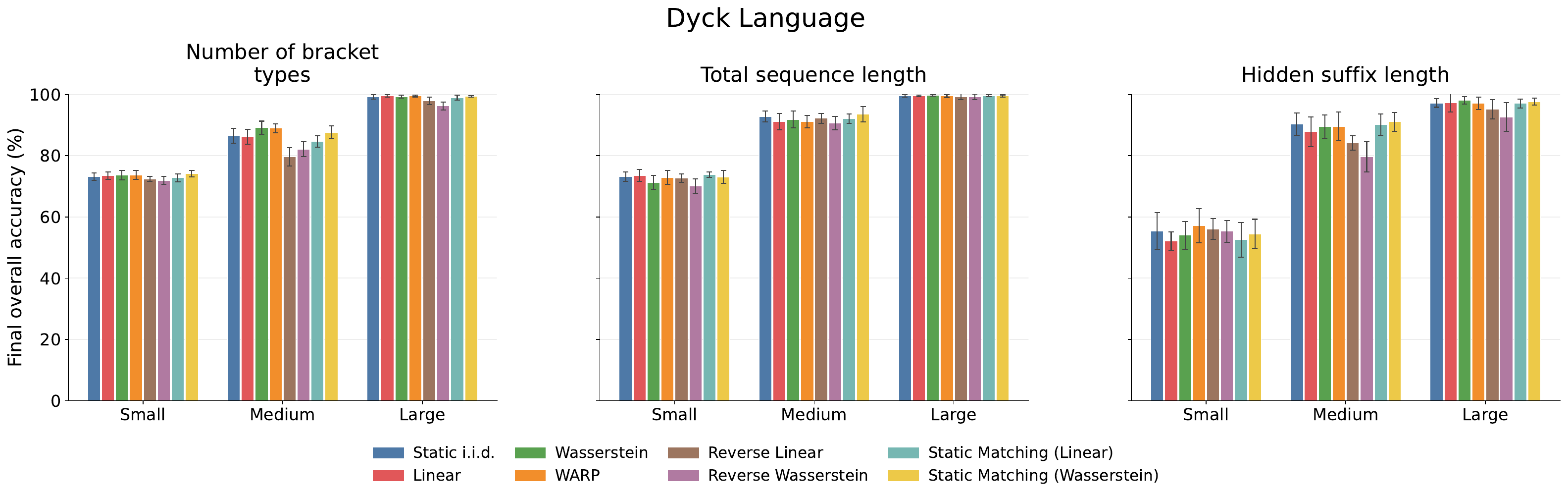}{Dyck Language}{fig:appendix_b1_dyck_language}
\taskaccuracyfigure{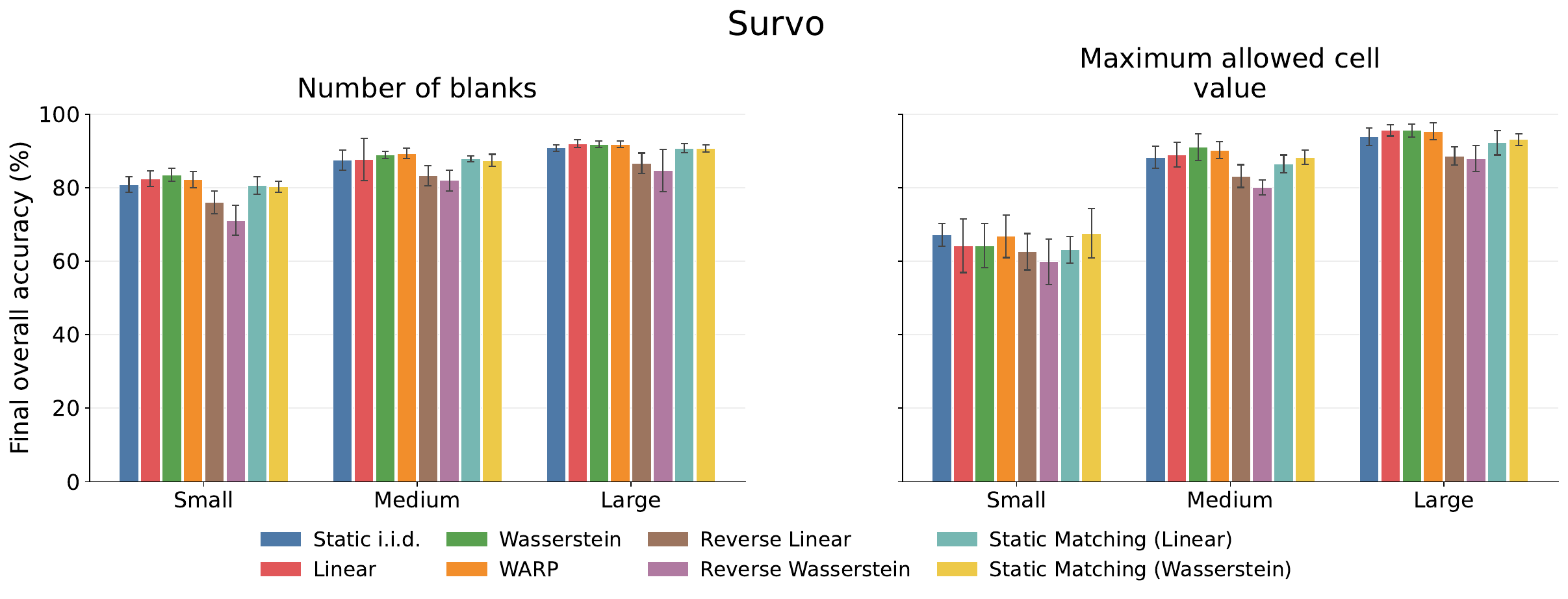}{Survo}{fig:appendix_b1_survo}
\taskaccuracyfigure{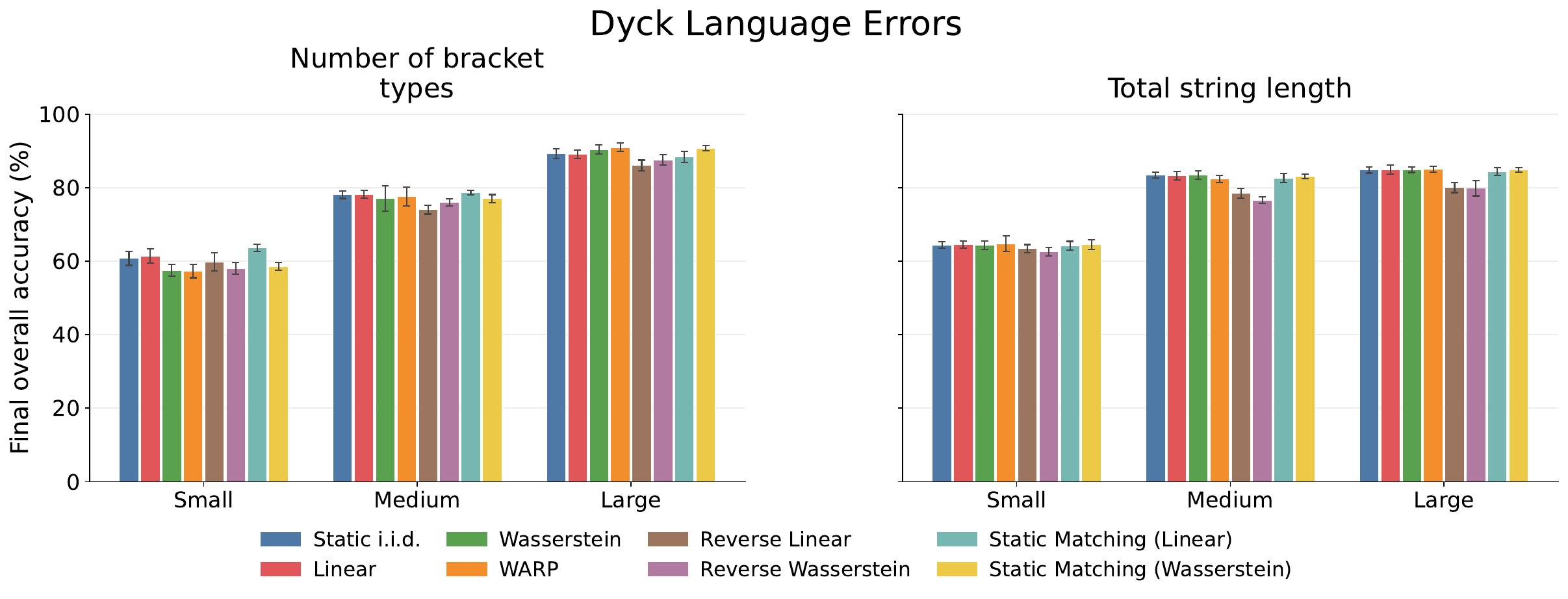}{Dyck Language Errors}{fig:appendix_b1_dyck_language_errors}
\taskaccuracyfigure{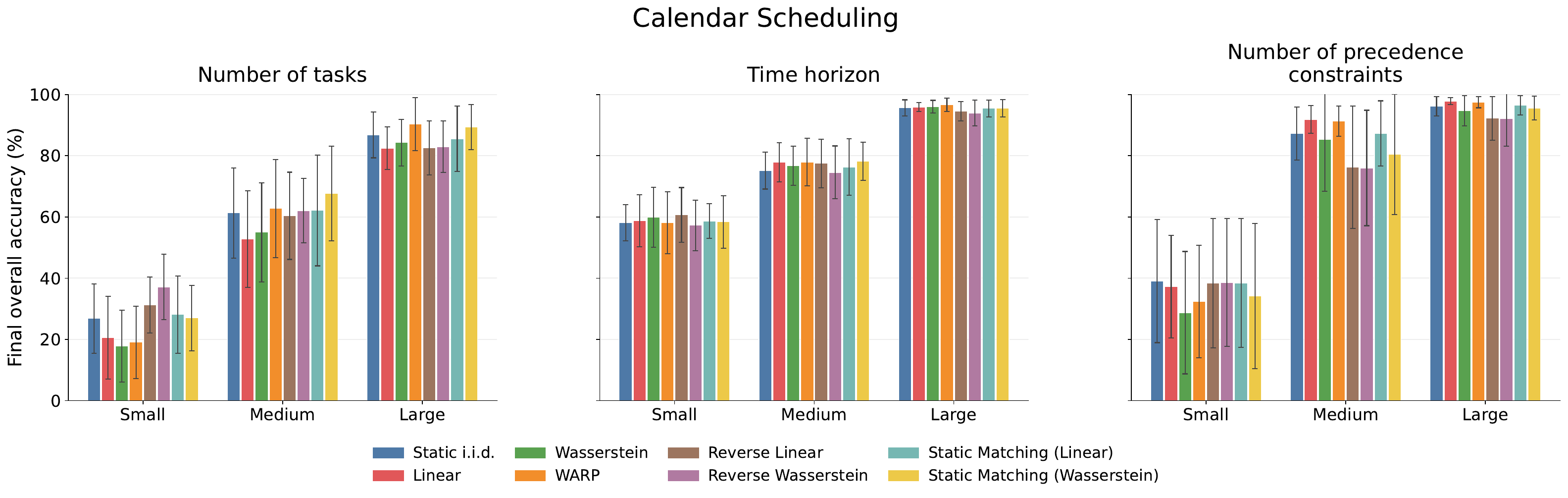}{Calendar Scheduling}{fig:appendix_b1_calendar_scheduling}
\taskaccuracyfigure{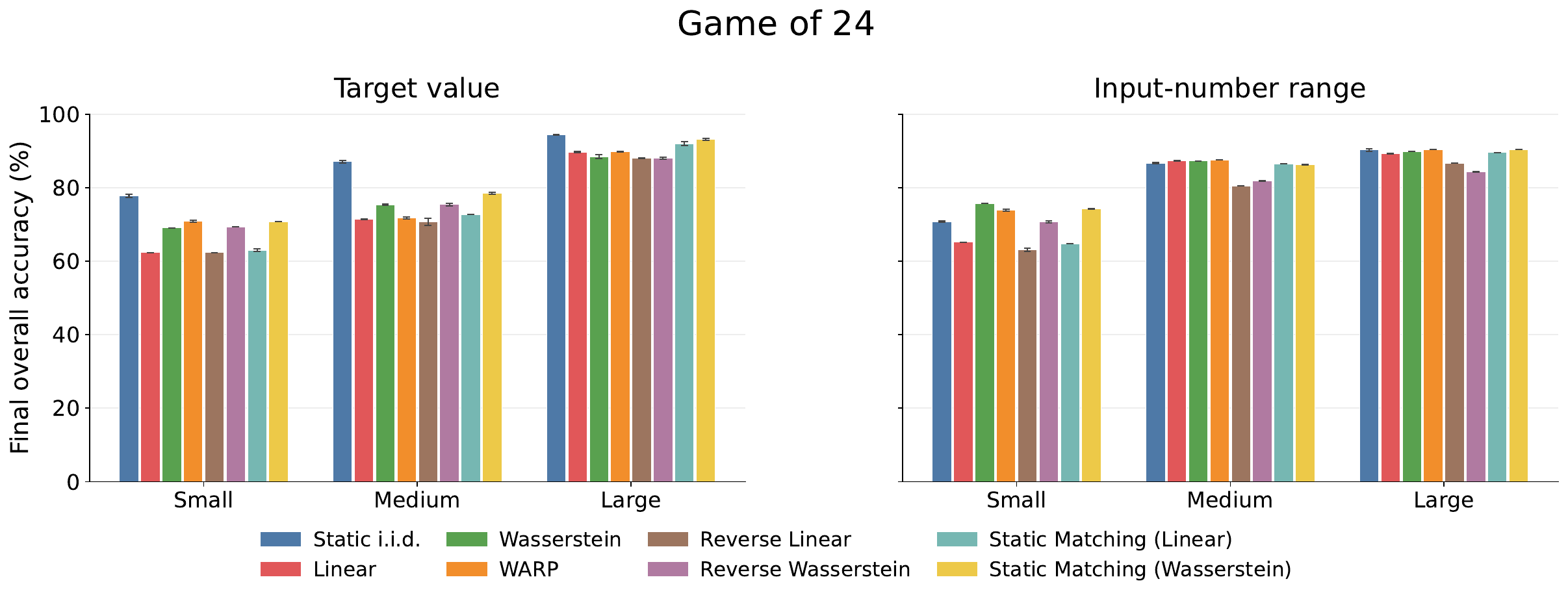}{Game of 24}{fig:appendix_b1_game_of_24}
\taskaccuracyfigure{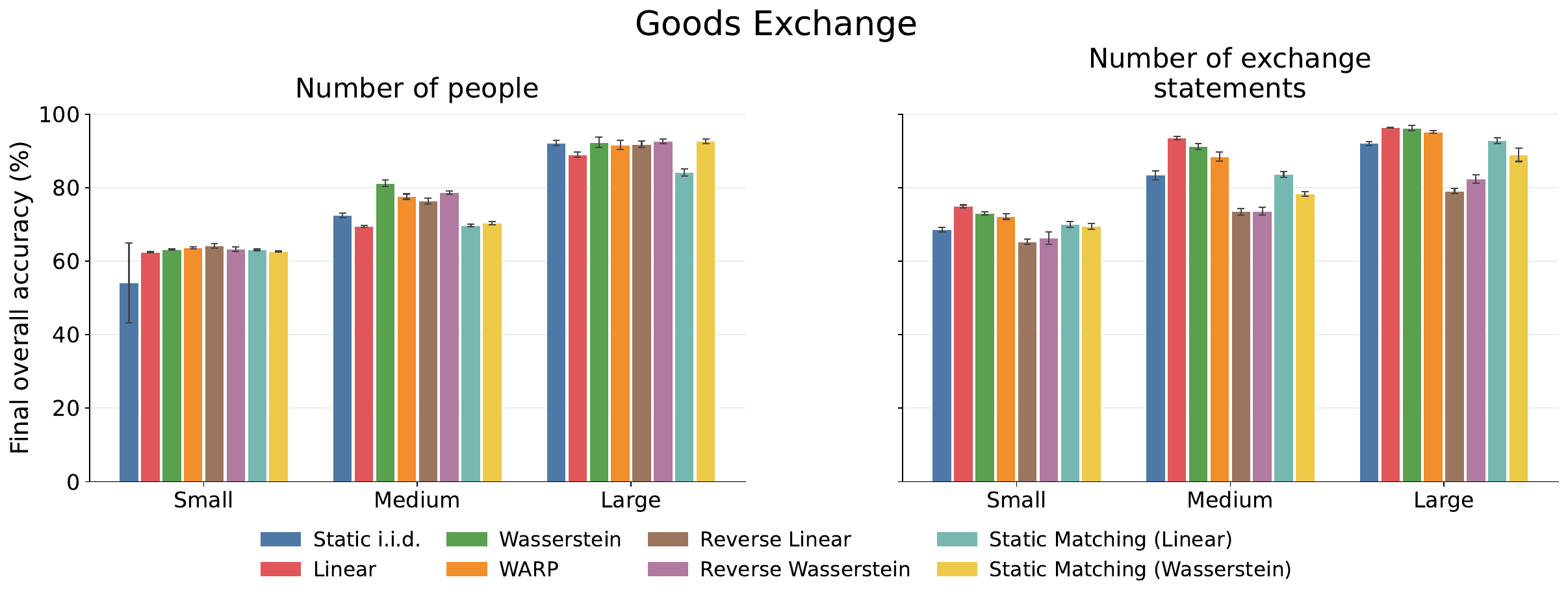}{Goods Exchange}{fig:appendix_b1_goods_exchange}
\taskaccuracyfigure{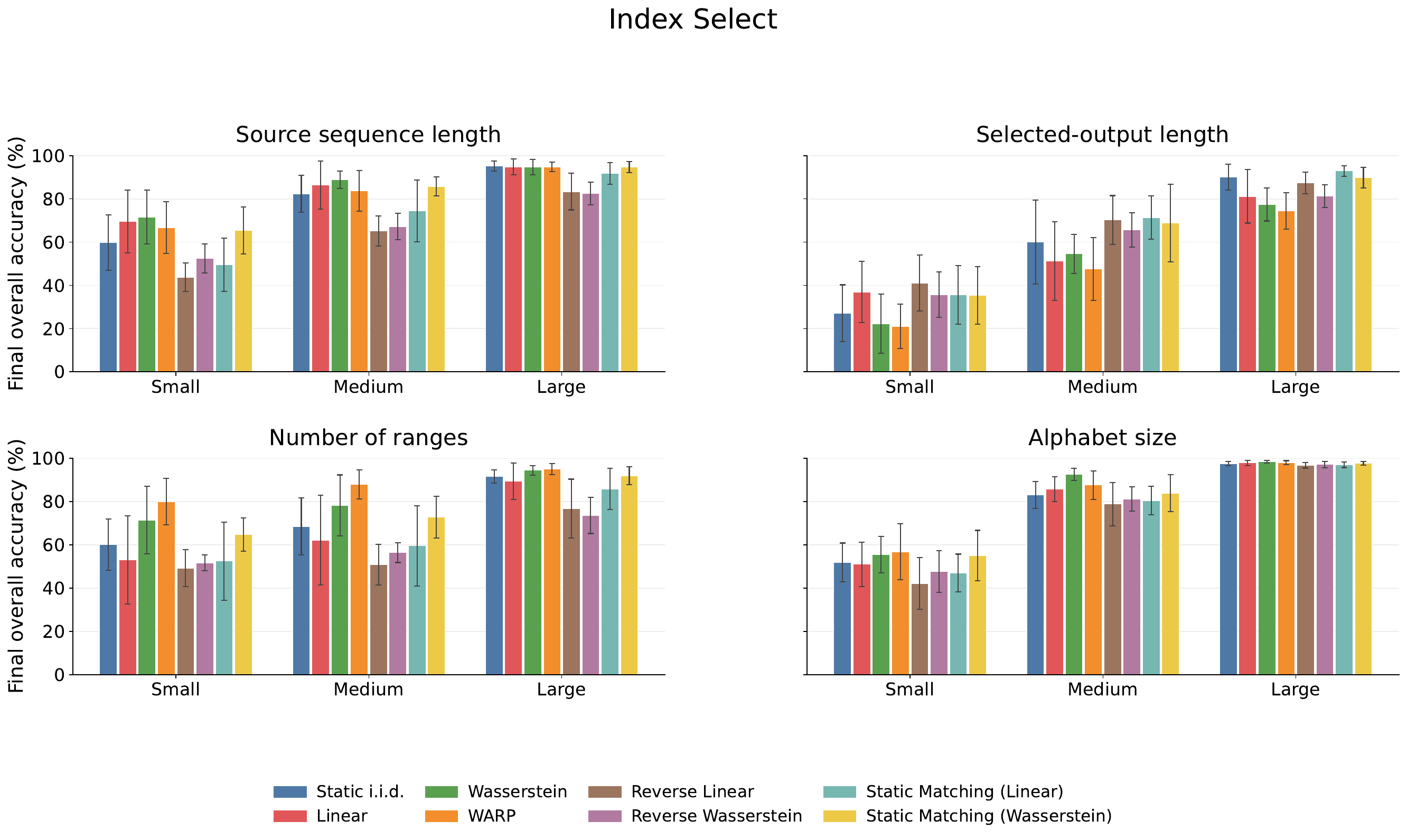}{Index Select}{fig:appendix_b1_index_select}
\taskaccuracyfigure{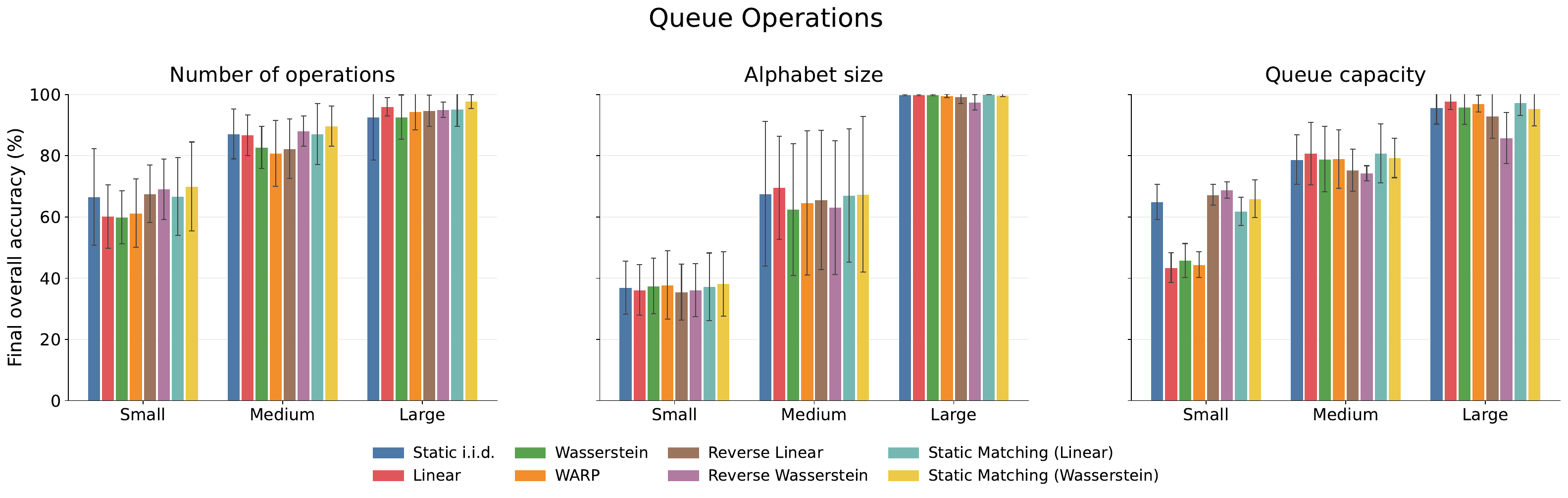}{Queue Operations}{fig:appendix_b1_queue_operations}
\taskaccuracyfigure{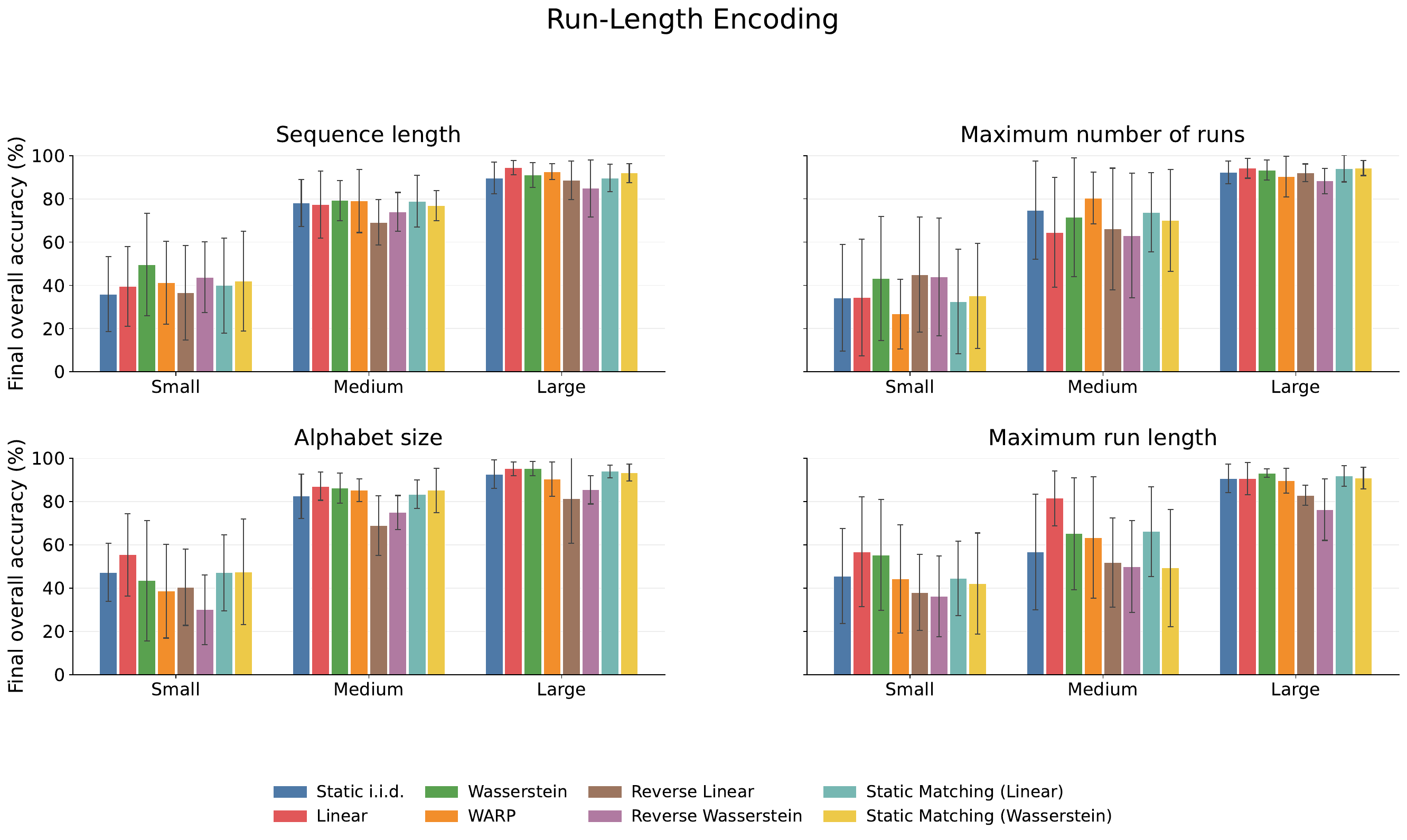}{Run-Length Encoding}{fig:appendix_b1_run_length_encode}
\taskaccuracyfigure{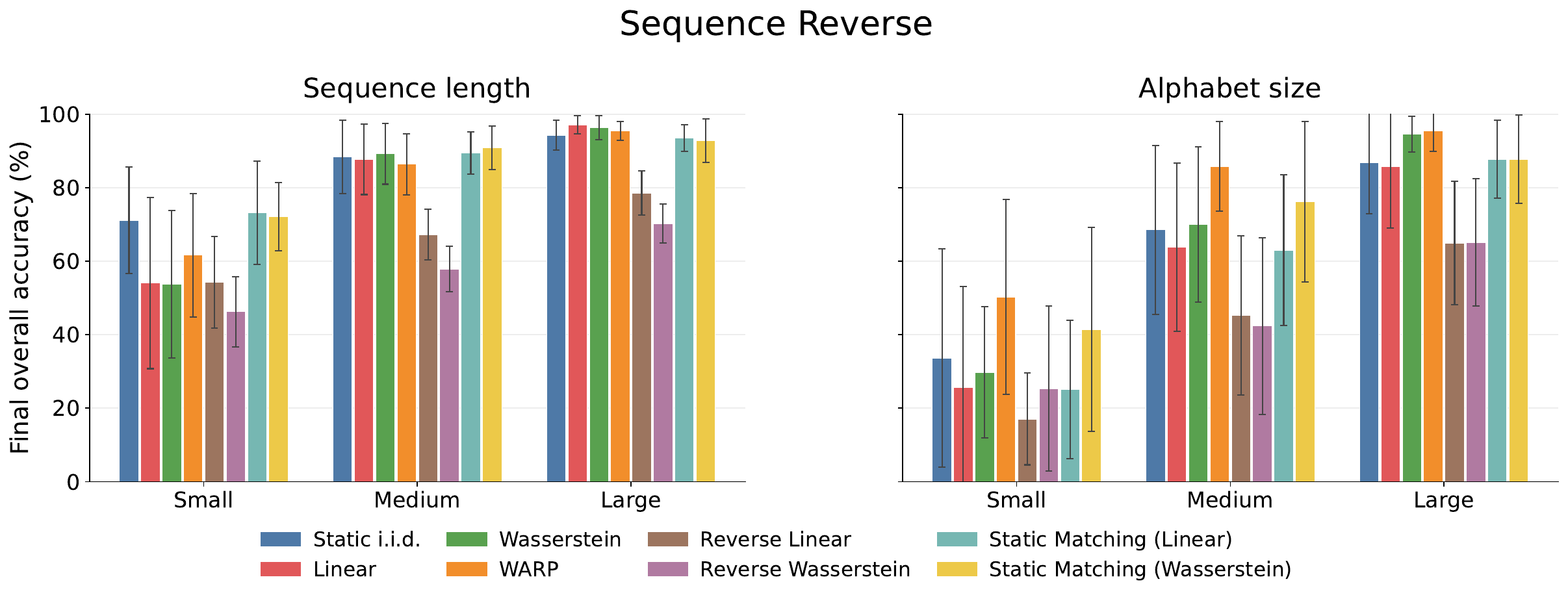}{Sequence Reverse}{fig:appendix_b1_sequence_reverse}
\taskaccuracyfigure{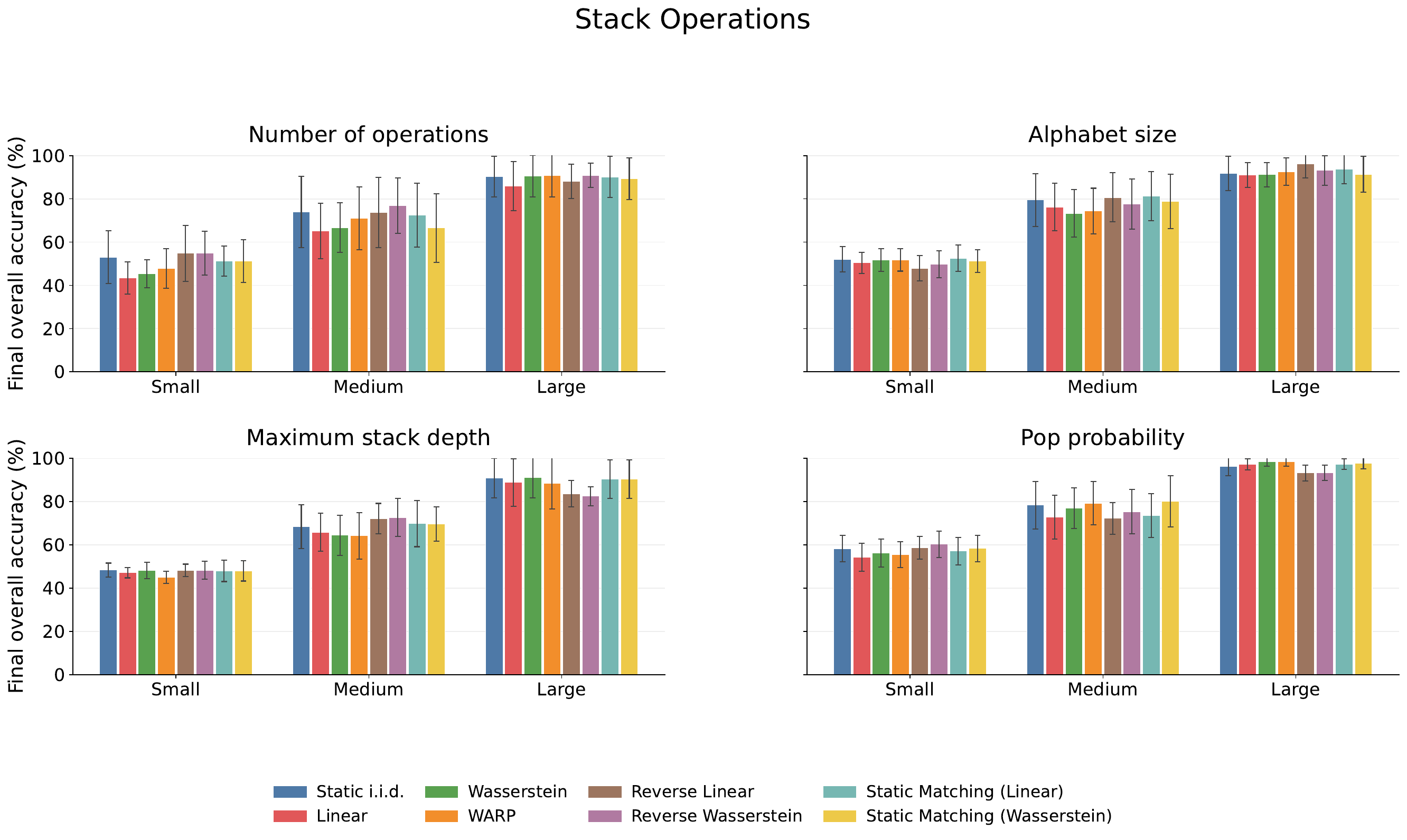}{Stack Operations}{fig:appendix_b1_stack_operations}

\subsection{Budget-Wise Exposure, Accuracy, and Exposure-Adjusted Accuracy}\label{appsubsec:budgetwise_curriculum_profiles}

\paragraph{Setup.}
To complement Section~\ref{subsec:level_profile}, we report the same easiest, middle, and hardest probe-bucket summaries separately for the small, medium, and large budgets. Exposure is the cumulative share of training mass assigned to each bucket over the full run. As a simple proxy for data efficiency, we use exposure-adjusted accuracy: final bucket accuracy divided by cumulative exposure to that bucket.

\paragraph{Findings.}
The medium row reproduces Figure~\ref{fig:curriculum_profiles}, while the small and large rows show how the same tradeoff evolves with budget. The qualitative pattern is stable: \textit{static i.i.d.} remains strongest on easy-end accuracy, linear is strongest in exposure-adjusted accuracy in the middle bucket, and Wasserstein is strongest on the hardest bucket.

\begin{figure*}[p]
  \centering
  \includegraphics[width=0.98\textwidth]{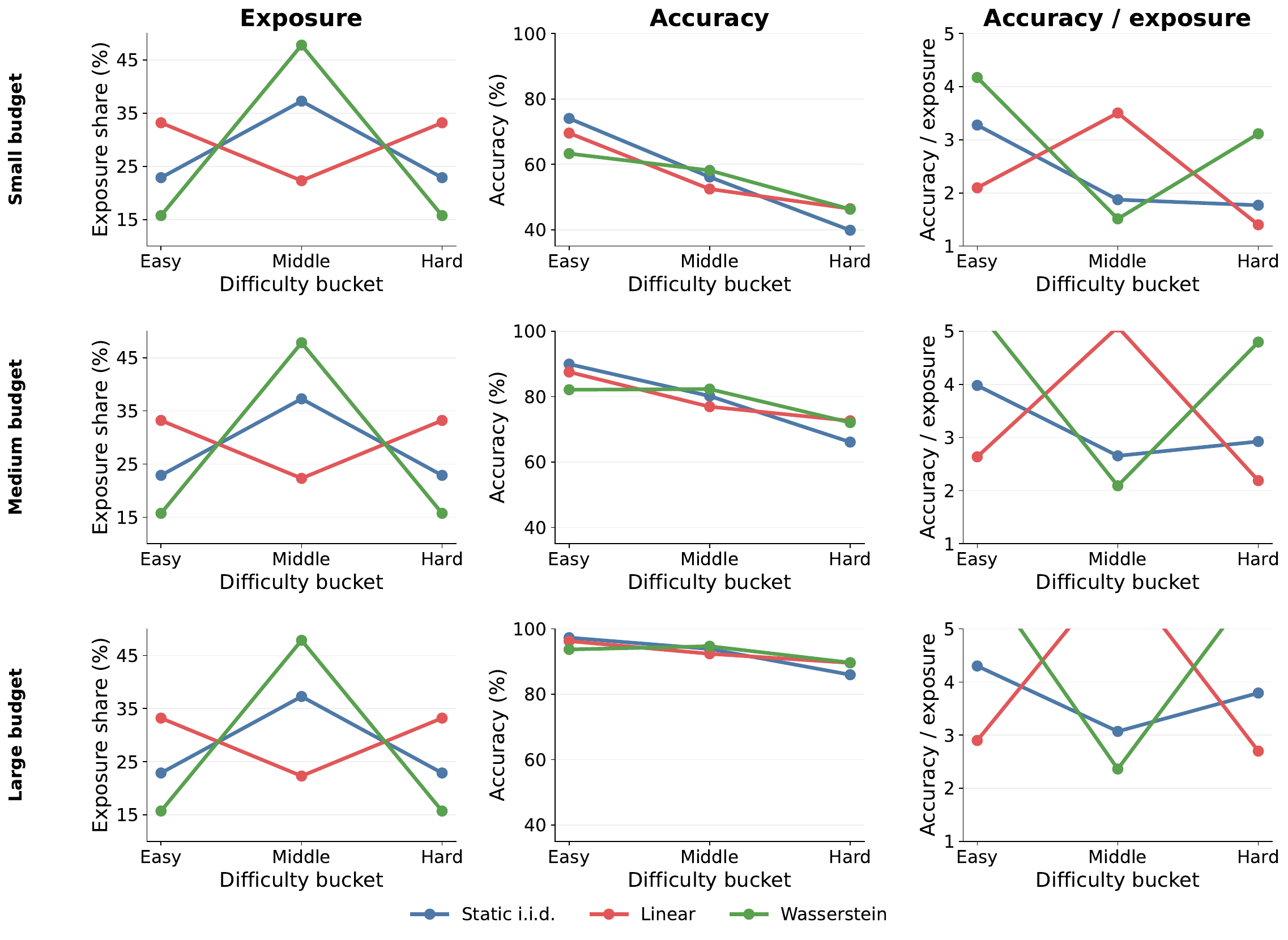}
  \caption{Exposure, accuracy, and exposure-adjusted accuracy across budgets. Rows denote the small, medium, and large budgets. Columns denote cumulative exposure share, final bucket accuracy, and exposure-adjusted accuracy on the easiest, middle, and hardest probe buckets.}
  \label{fig:appendix_b0_curriculum_profiles}
\end{figure*}
\FloatBarrier

\subsection{Task-Wise Final Level Profiles}\label{appsubsec:taskwise_levelwise_profiles}

\paragraph{Setup.}
To complement the task-wise endpoint barplots above, we also visualize where final performance lands across the available difficulty levels. Each figure again corresponds to one task, with rows denoting difficulty axes and columns denoting the small, medium, and large budgets. Within each panel, colored curves trace the mean final accuracy at each level for the eight curricula. Tasks with four levels stop at Level~4, while five-level tasks extend through Level~5.

\paragraph{Findings.}
These profile grids make the location of gains explicit: some curricula preserve more of the easy-end accuracy, while others shift the profile upward at the harder end. They therefore complement the budget-wise summaries and task-wise endpoint barplots by showing not only which curriculum wins overall, but also where along the task's level structure the final gains and tradeoffs occur.

\newcommand{\tasklevelprofilefigure}[3]{%
\begin{center}
  \begin{minipage}{\textwidth}
    \centering
    \includegraphics[width=0.98\textwidth,height=0.92\textheight,keepaspectratio]{figures/#1}
    \captionof{figure}{Task-wise final level profiles for #2. Rows denote difficulty axes, columns denote budgets, and curves show mean final accuracy at each available difficulty level over 10 random-seed repetitions.}
    \label{#3}
  \end{minipage}
\end{center}
\vspace{0.5em}
}

\tasklevelprofilefigure{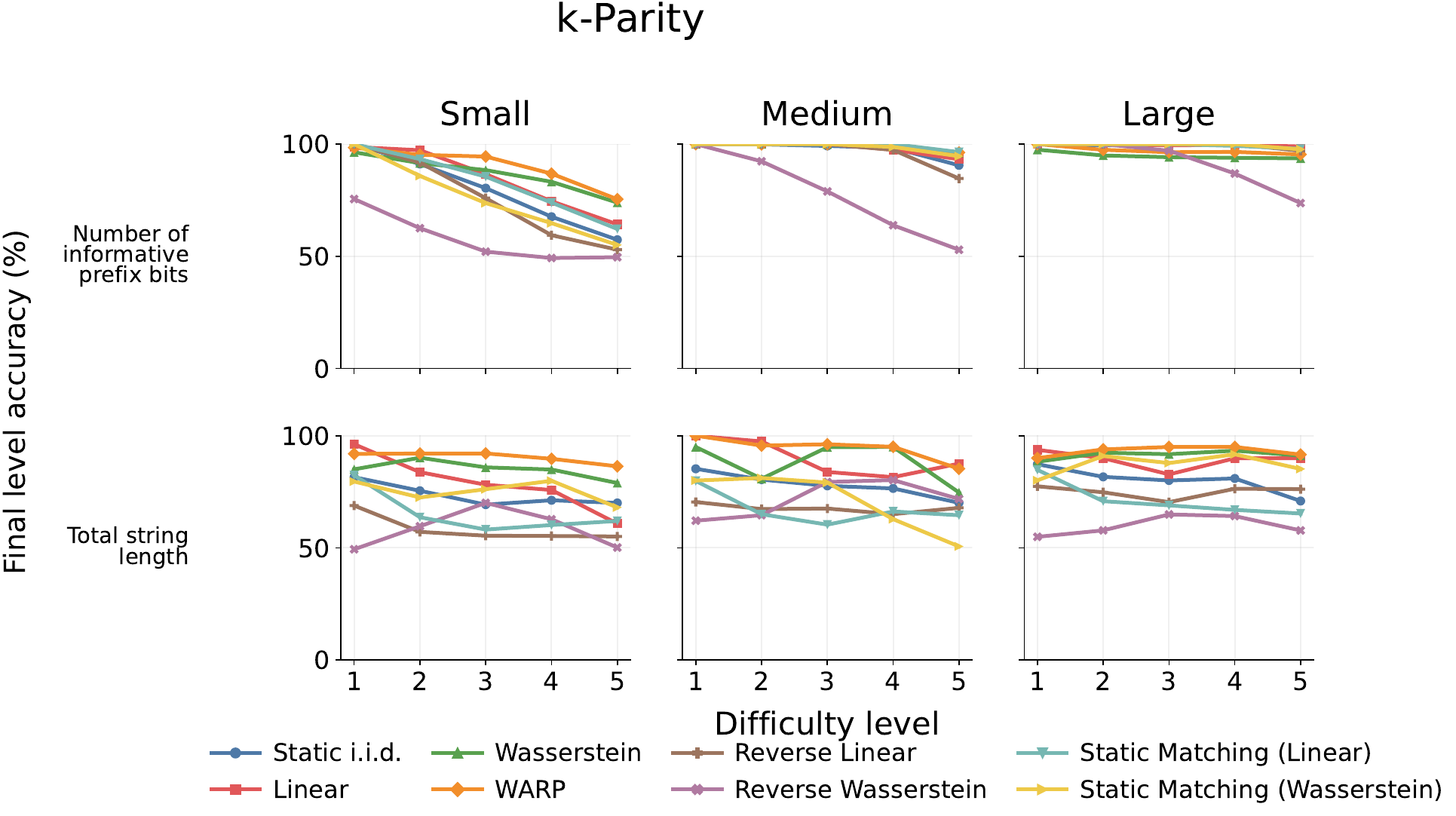}{K-Parity}{fig:appendix_b3_k_parity}
\tasklevelprofilefigure{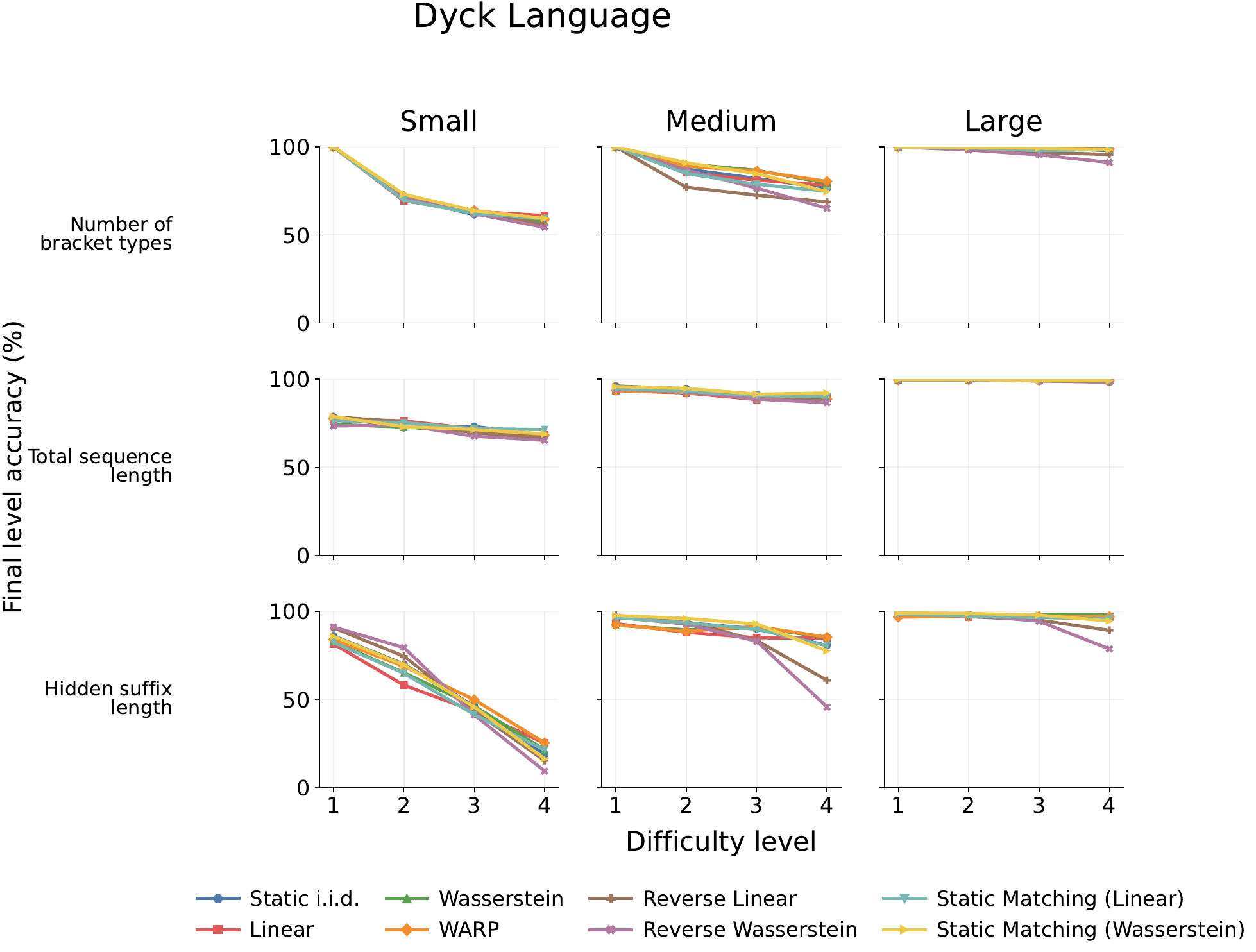}{Dyck Language}{fig:appendix_b3_dyck_language}
\tasklevelprofilefigure{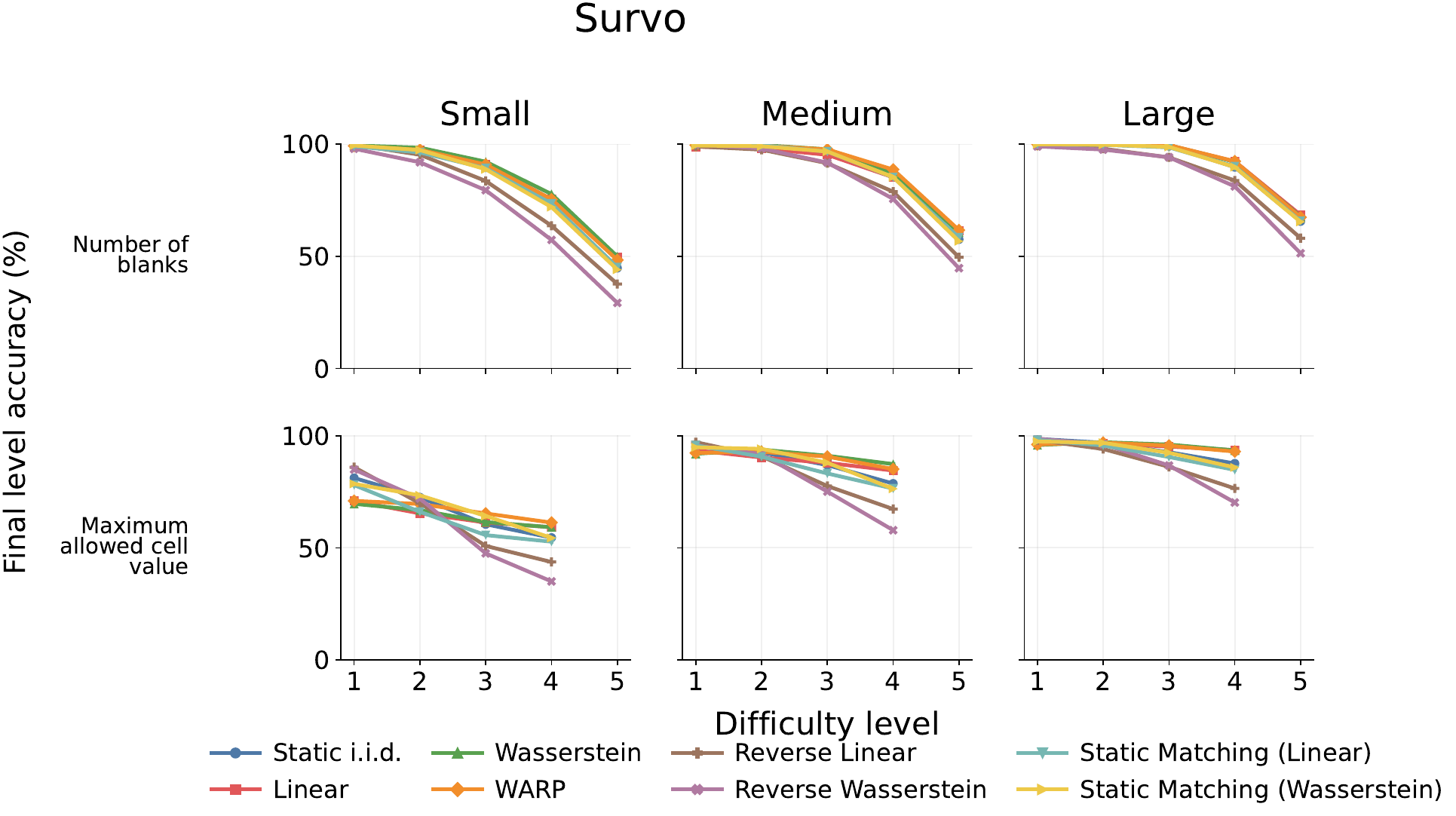}{Survo}{fig:appendix_b3_survo}
\tasklevelprofilefigure{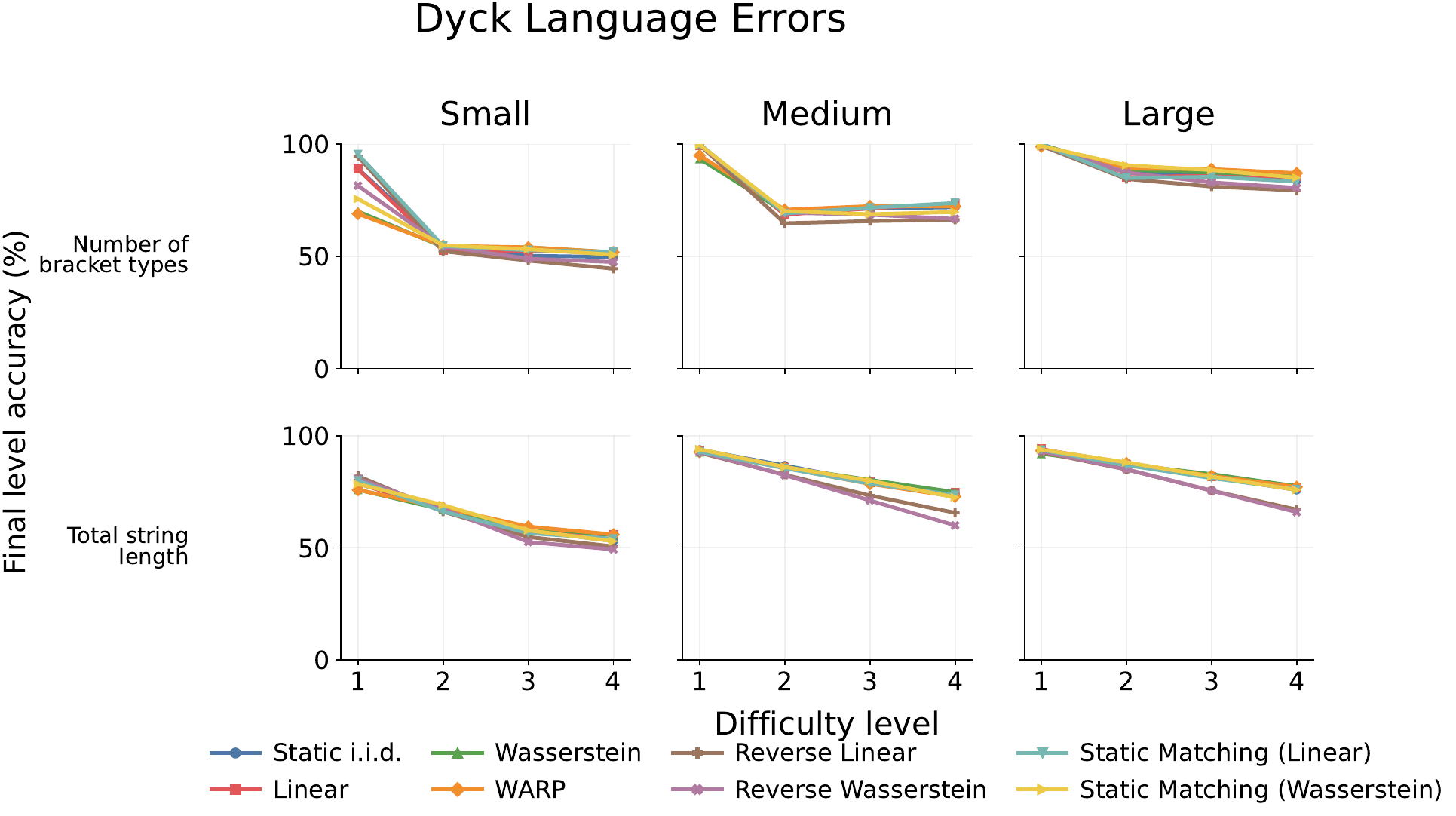}{Dyck Language Errors}{fig:appendix_b3_dyck_language_errors}
\tasklevelprofilefigure{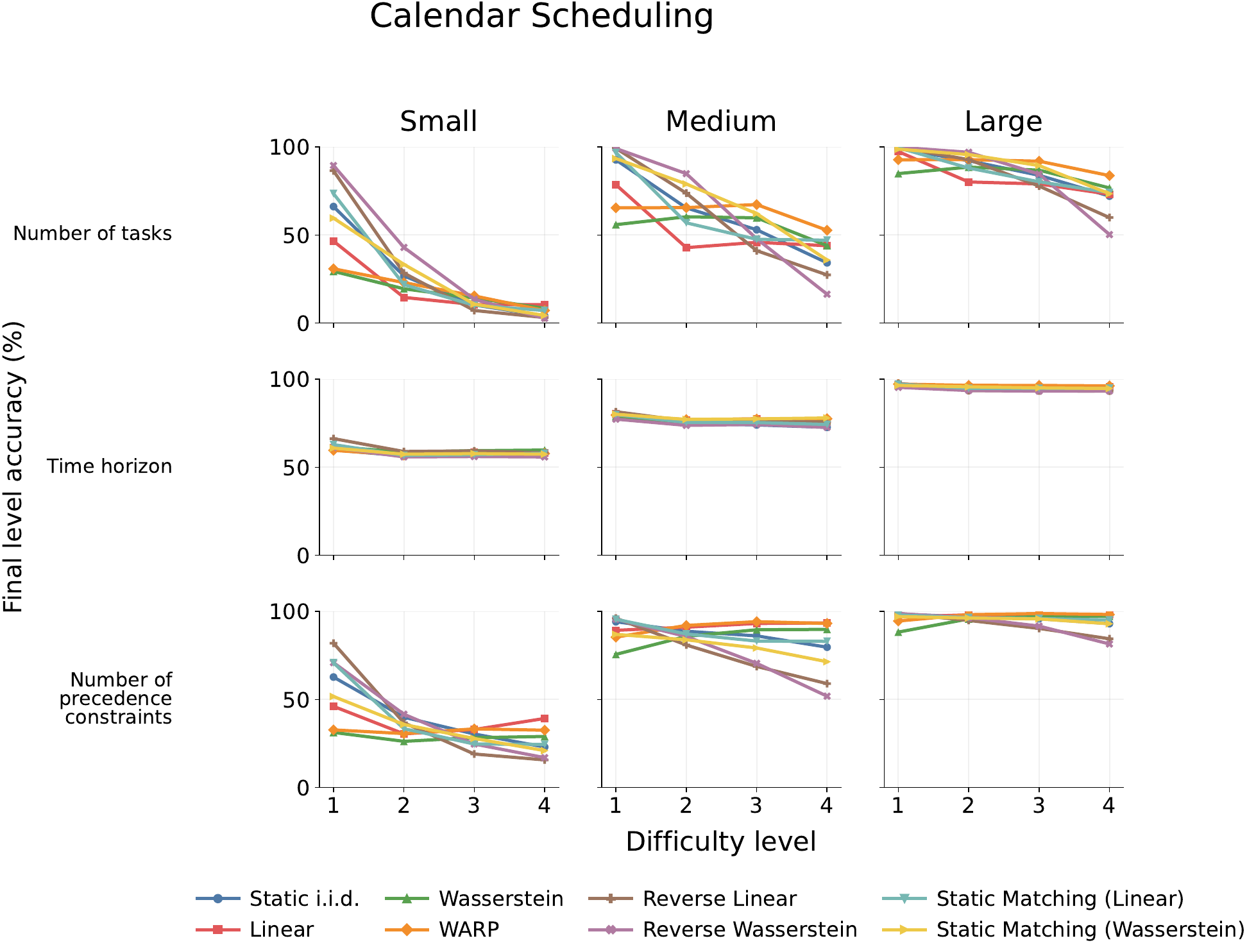}{Calendar Scheduling}{fig:appendix_b3_calendar_scheduling}
\tasklevelprofilefigure{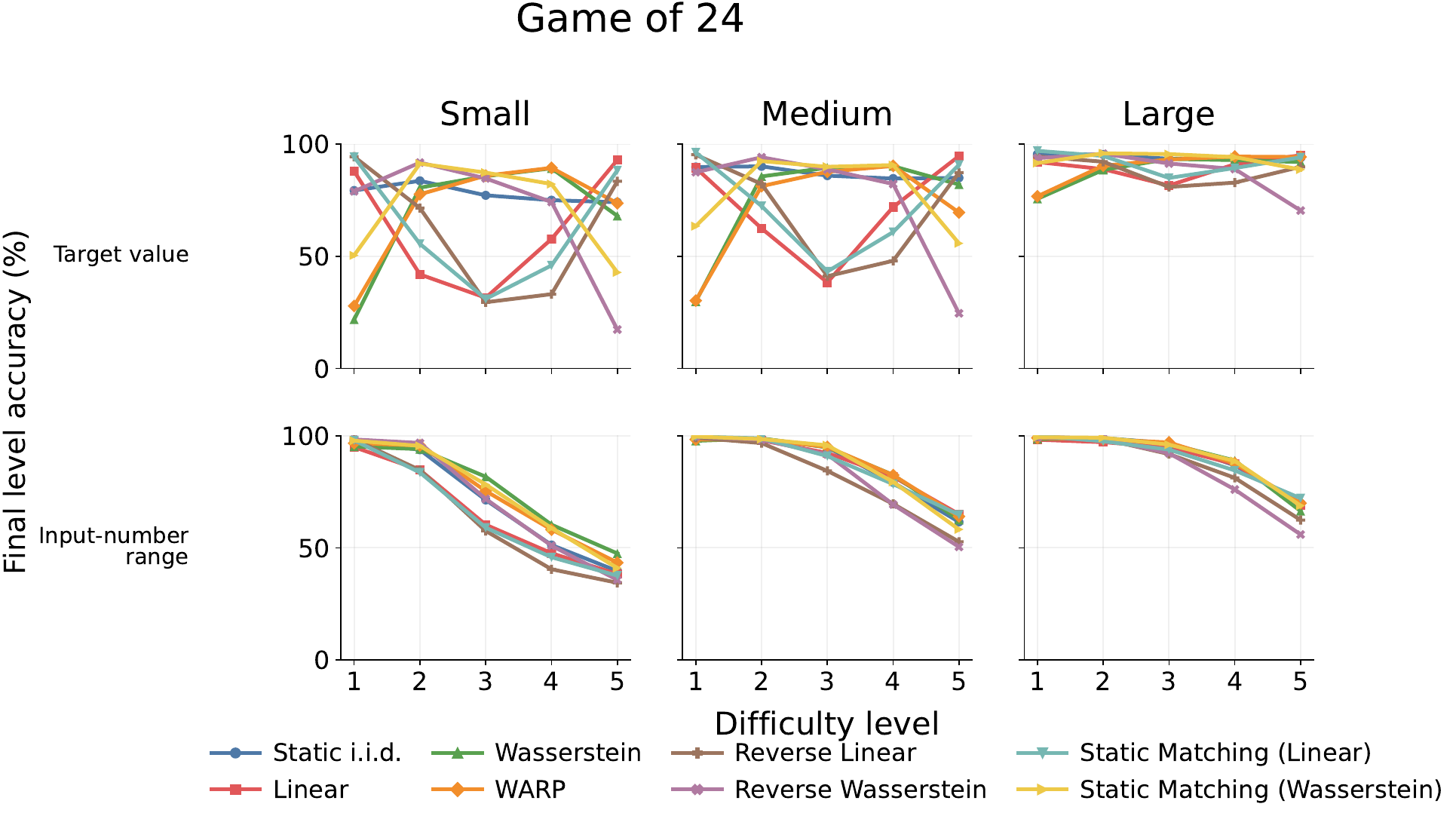}{Game of 24}{fig:appendix_b3_game_of_24}
\tasklevelprofilefigure{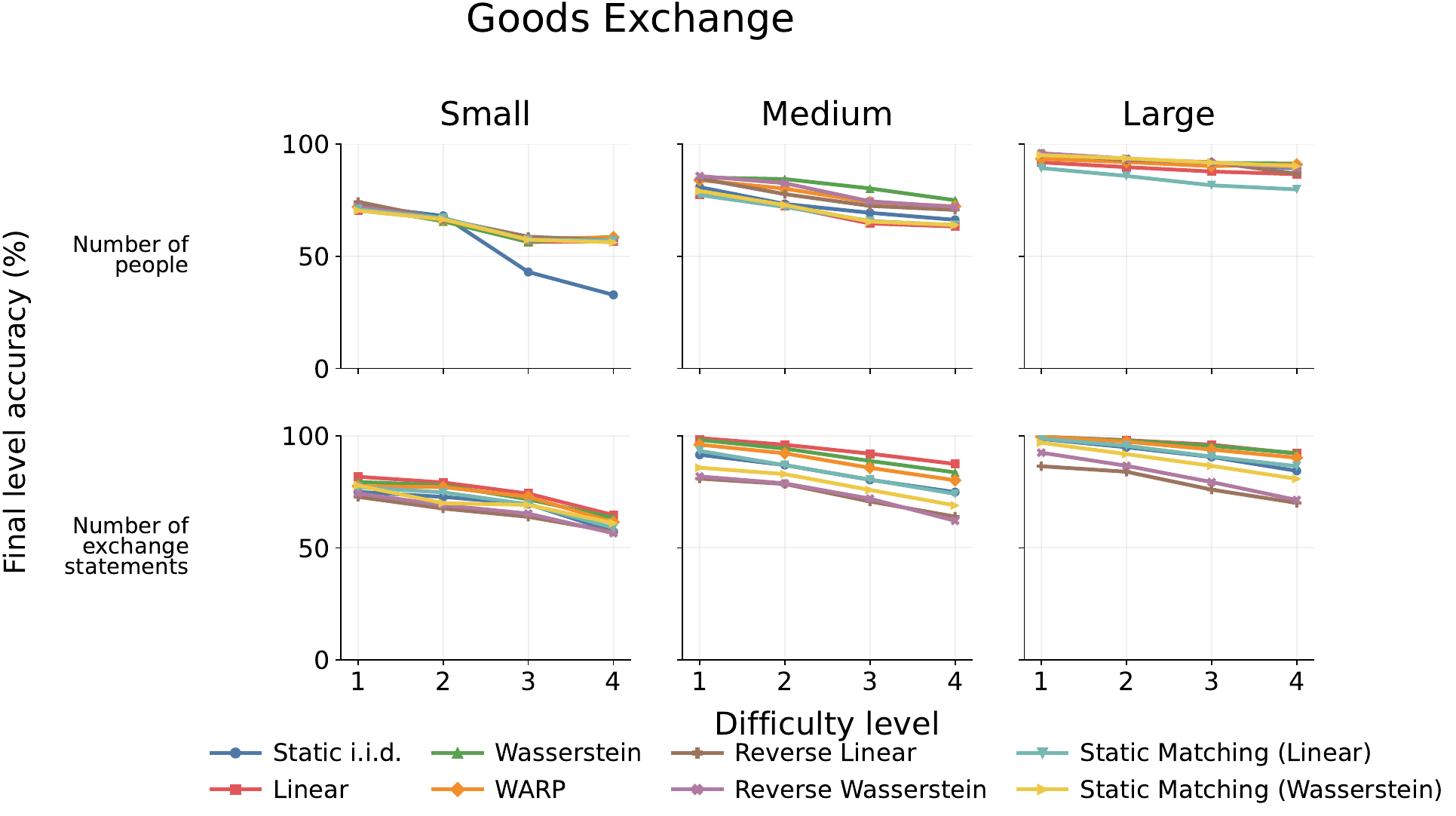}{Goods Exchange}{fig:appendix_b3_goods_exchange}
\tasklevelprofilefigure{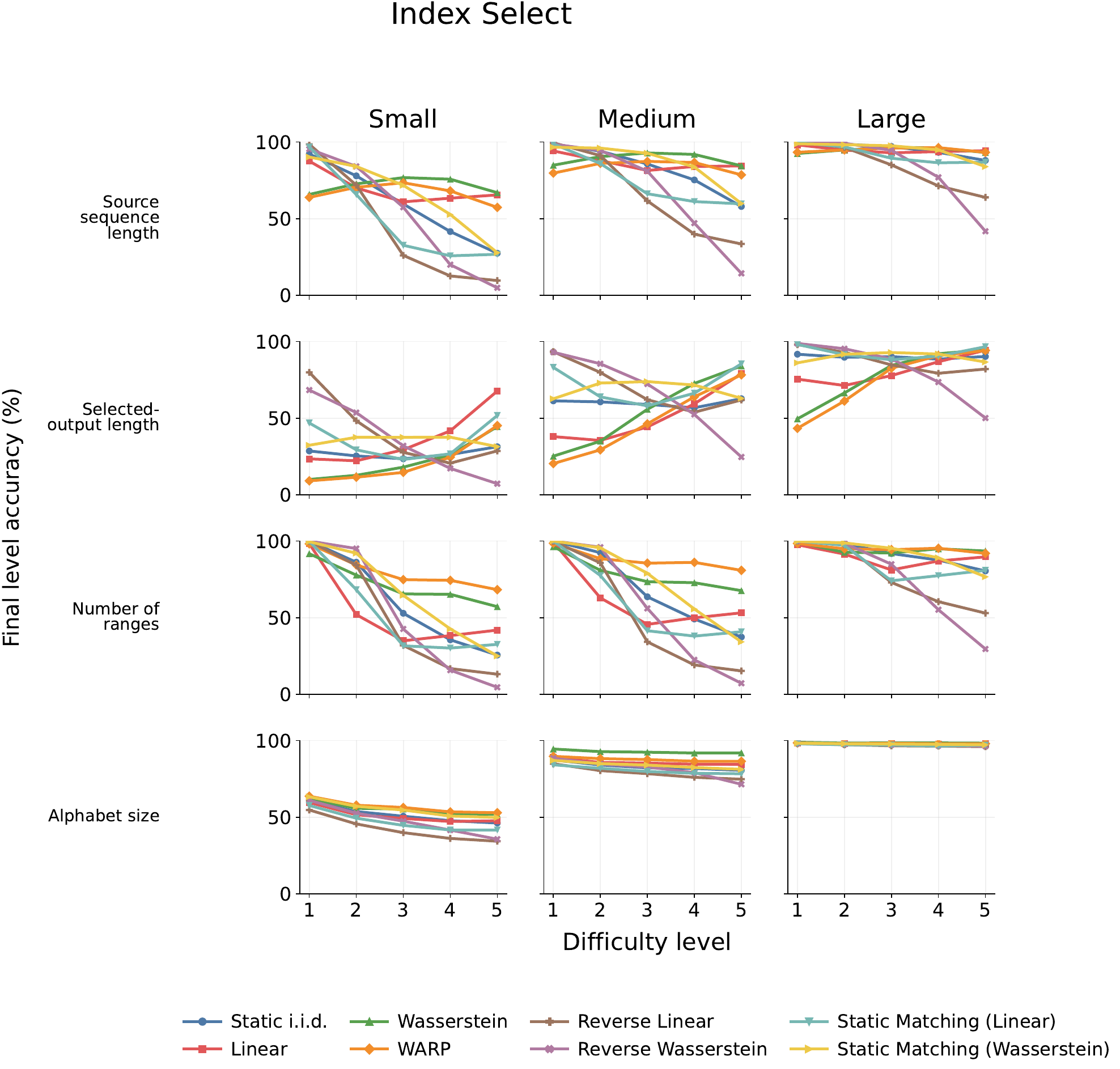}{Index Select}{fig:appendix_b3_index_select}
\tasklevelprofilefigure{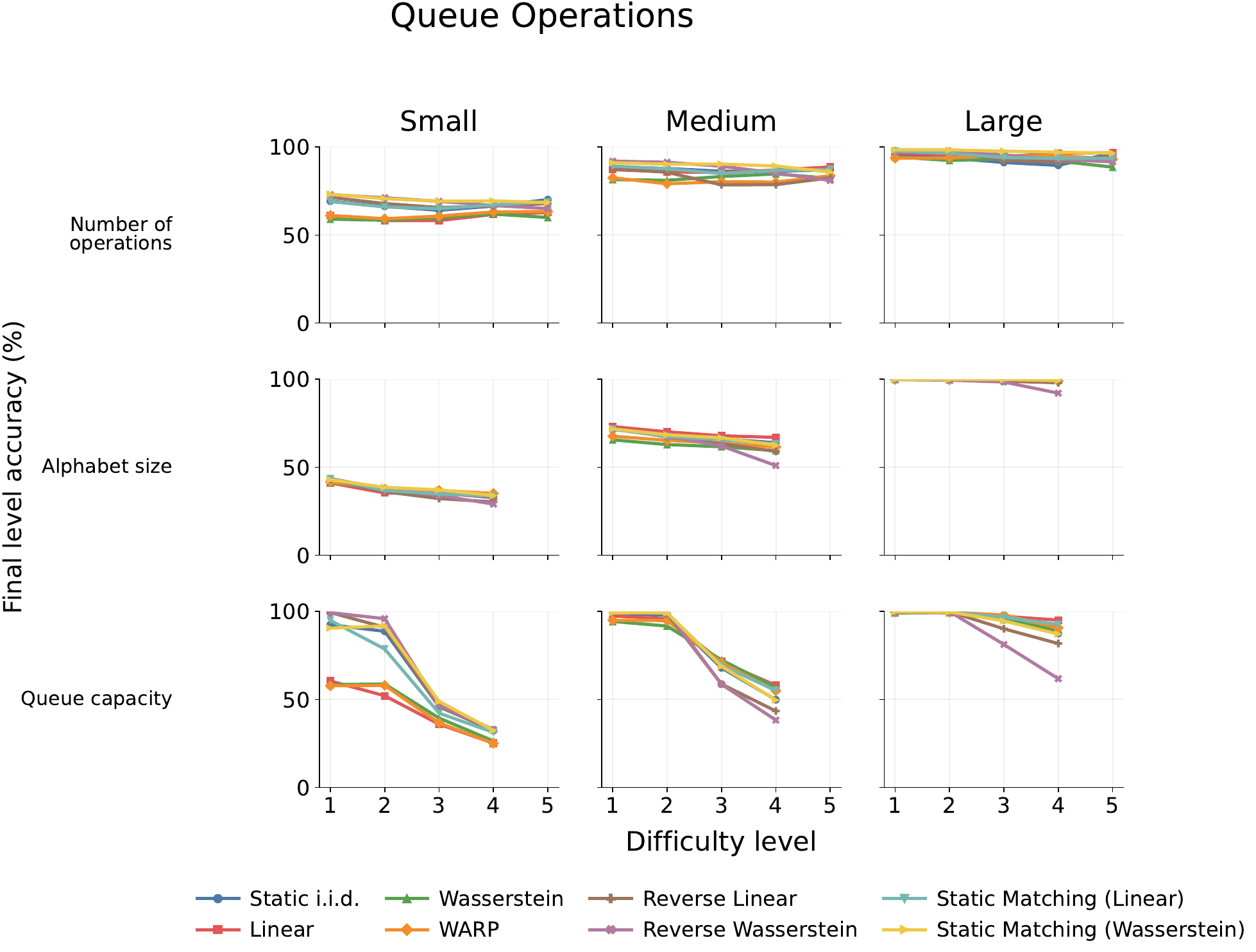}{Queue Operations}{fig:appendix_b3_queue_operations}
\tasklevelprofilefigure{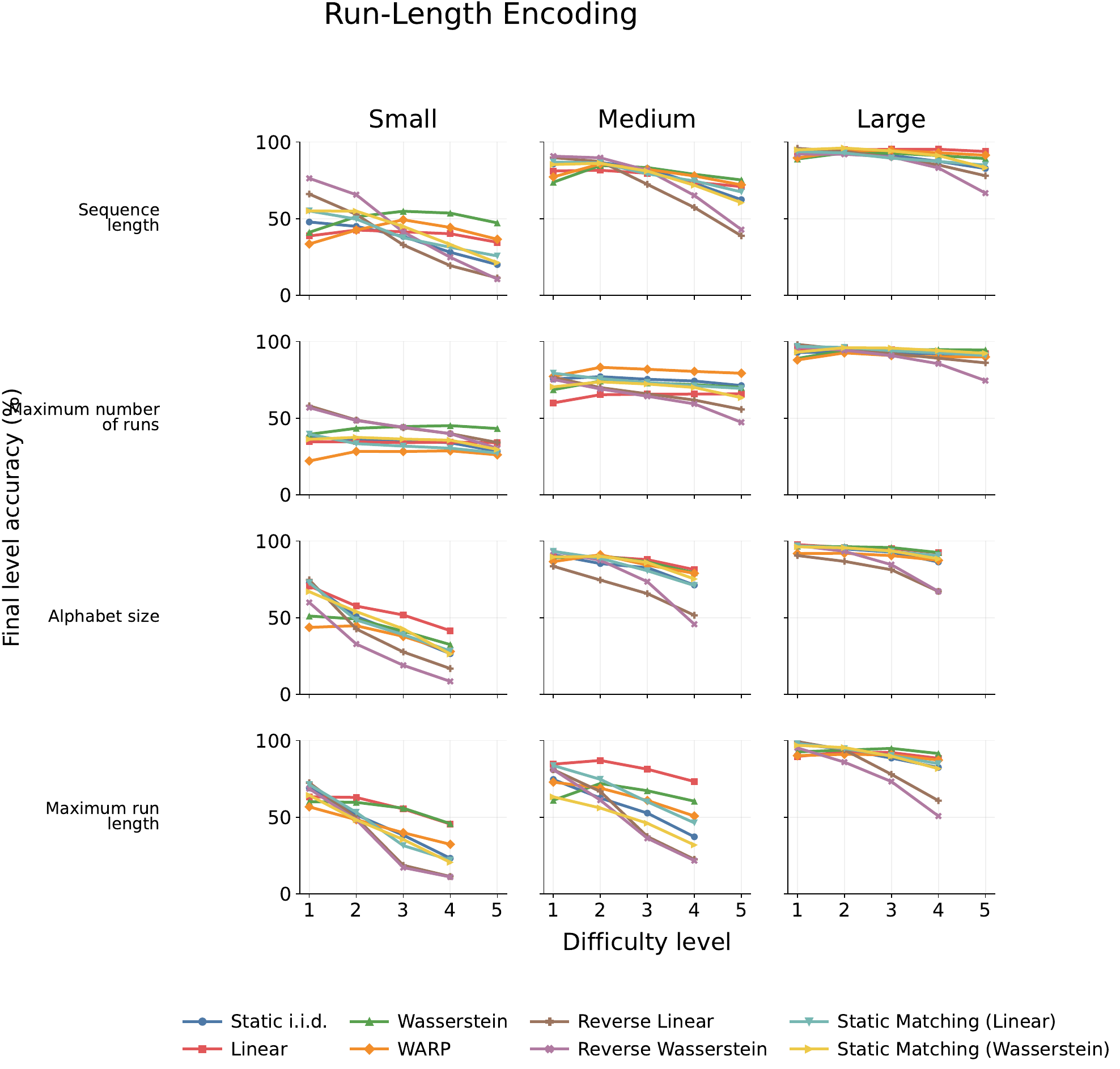}{Run-Length Encoding}{fig:appendix_b3_run_length_encode}
\tasklevelprofilefigure{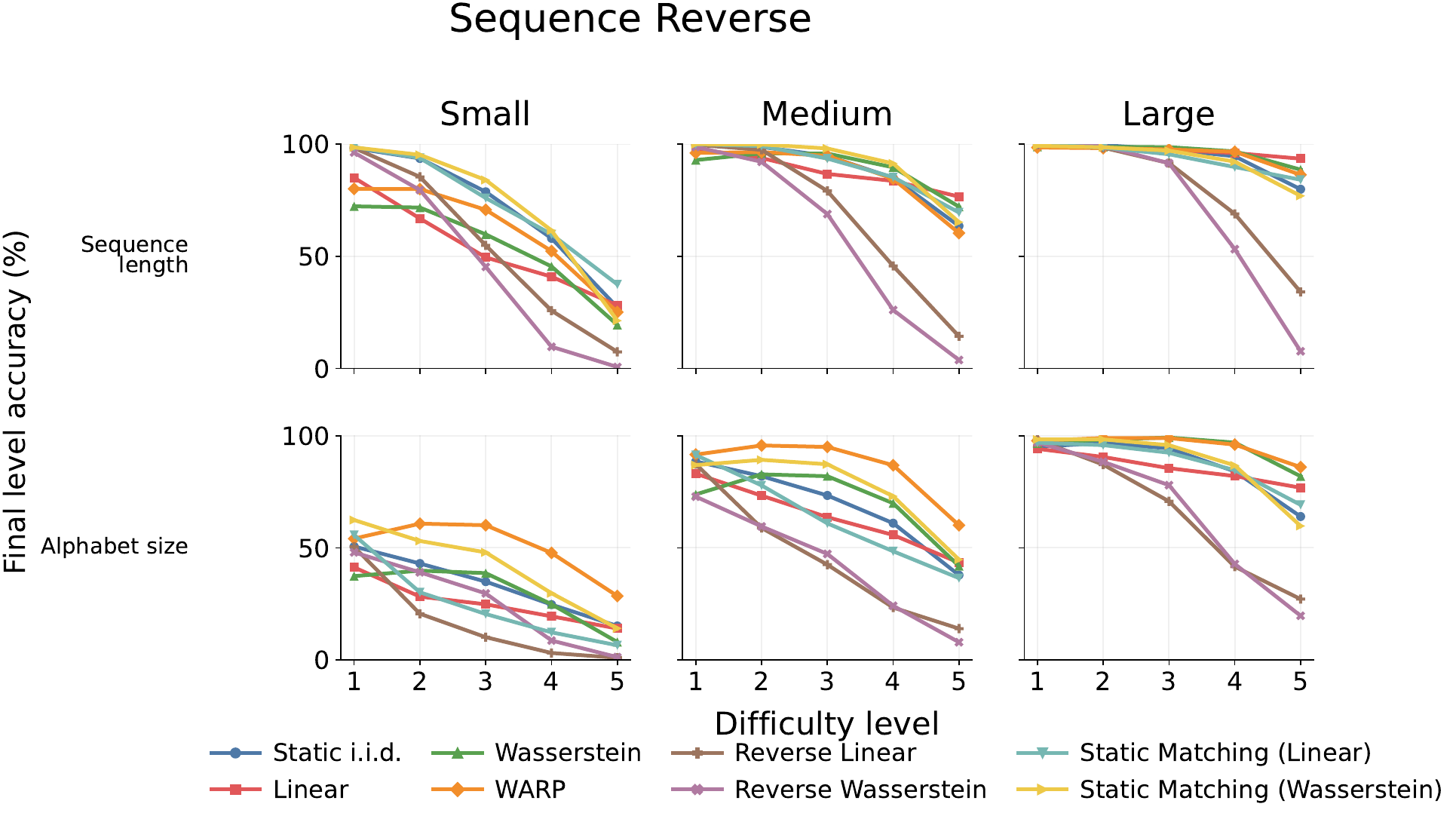}{Sequence Reverse}{fig:appendix_b3_sequence_reverse}
\tasklevelprofilefigure{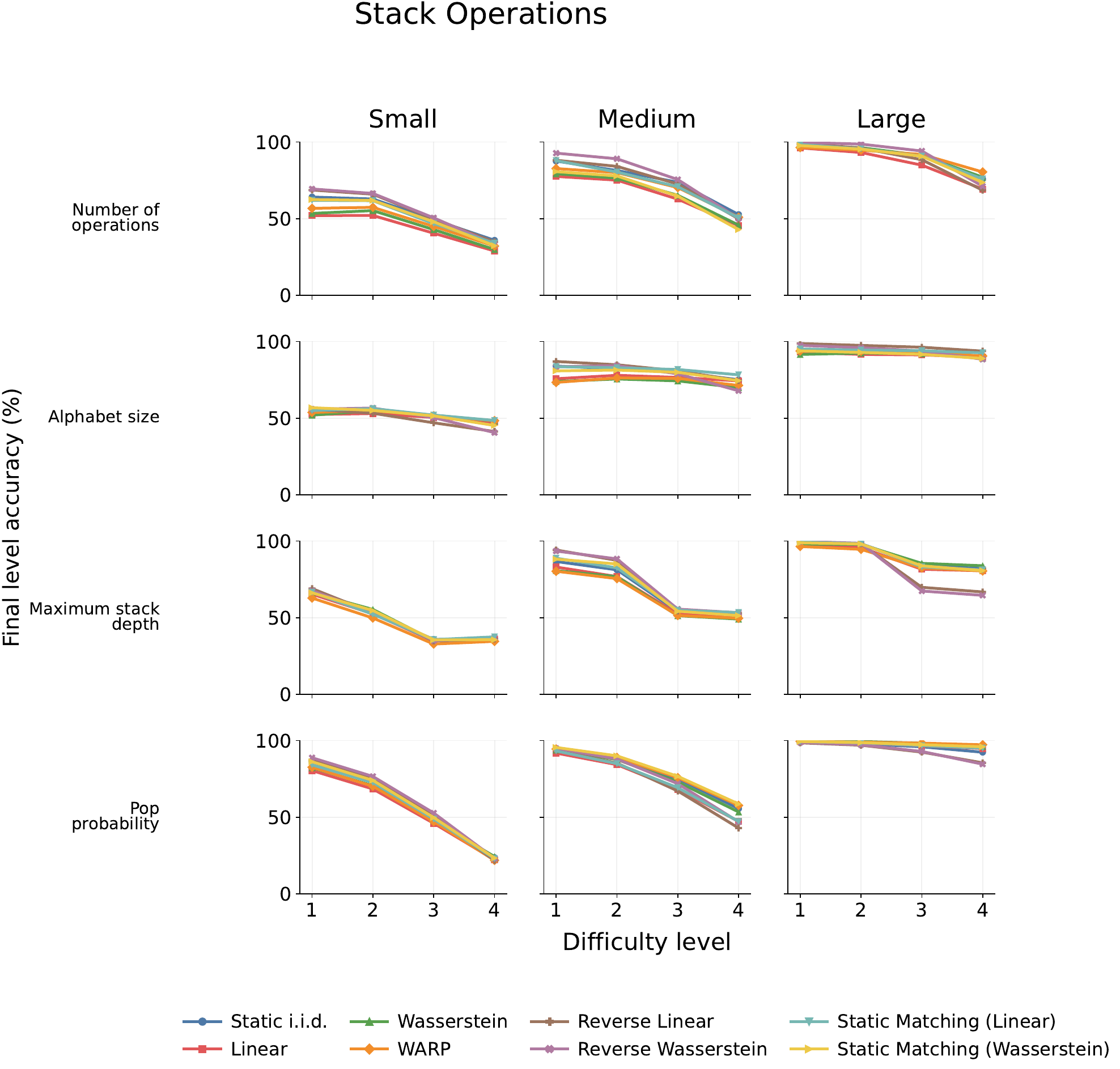}{Stack Operations}{fig:appendix_b3_stack_operations}

\subsection{Exploratory Analysis of When Easy-to-Hard Progression Helps}\label{appsubsec:natural_progression_axes}

Many curriculum methods depend strongly on the chosen difficulty signal or decomposition \citep{platanios2019competence,hacohen2019power,swayamdipta2020dataset,jia2025makes}. We group the 99 task-by-axis-and-budget conditions by task family, difficulty-axis category, and their intersection, and count where \textit{Wasserstein} or \SYSNAME{} attains the best mean among the four forward curricula.

\paragraph{I. Tasks.}
We first ask the same question at the task level by grouping the 12 benchmarks into four coarse task families listed below. Unlike the difficulty-axis taxonomy below, these task families do not overlap.

\begin{itemize}[leftmargin=1.5em,itemsep=1pt,topsep=2pt,parsep=0pt,partopsep=0pt]
  \item \textbf{Sequence Transformation:} \textsc{Index Select}, \textsc{Run-Length Encoding}, \textsc{Sequence Reverse}
  \item \textbf{Formal Structure:} \textsc{Dyck Language}, \textsc{Dyck Language Errors}, \textsc{K-Parity}
  \item \textbf{Data-Structure Execution:} \textsc{Queue Operations}, \textsc{Stack Operations}
  \item \textbf{Constraint and Search:} \textsc{Calendar Scheduling}, \textsc{Game of 24}, \textsc{Goods Exchange}, \textsc{Survo}
\end{itemize}

\begin{center}
\begin{minipage}{\textwidth}
\centering
\setlength{\tabcolsep}{8pt}
\begin{tabular}{lcc}
\toprule
Task family & Overall wins & Hardest-level wins \\
\midrule
\textbf{Sequence Transformation} & \(18/30\) (\(10\) W, \(8\) A) & \(21/30\) (\(14\) W, \(7\) A) \\
\textbf{Formal Structure} & \(15/21\) (\(4\) W, \(11\) A) & \(13/21\) (\(4\) W, \(9\) A) \\
\textbf{Data-Structure Execution} & \(7/21\) (\(2\) W, \(5\) A) & \(8/21\) (\(2\) W, \(6\) A) \\
\textbf{Constraint and Search} & \(15/27\) (\(7\) W, \(8\) A) & \(15/27\) (\(8\) W, \(7\) A) \\
\bottomrule
\end{tabular}
\captionof{table}{Task-family summary of when natural easy-to-hard progression helps. We group the 12 benchmarks into four coarse task families and count the number of task-by-axis-and-budget conditions for which \textit{Wasserstein} or \SYSNAME{} attains the best mean among the four forward curricula (\textit{static i.i.d.}, linear, Wasserstein, \SYSNAME{}). The left numeric pair reports overall wins; the right reports wins on the hardest available level. `W' denotes Wasserstein and `A' denotes \SYSNAME{}.}
\label{tab:appendix_c3_natural_progression_tasks}
\end{minipage}
\end{center}

\paragraph{Task findings.}
The task taxonomy is somewhat cleaner than the axis taxonomy, but it still does not yield a single simple rule. Natural easy-to-hard progression is strongest on \emph{Sequence Transformation} tasks, where it wins in a majority of conditions overall and even more often on the hardest level. \emph{Formal Structure} and the broader \emph{Constraint and Search} family also show substantial support, especially at the hard end. By contrast, \emph{Data-Structure Execution} remains more mixed.

\paragraph{II. Difficulty Axes.}
We next simplify the raw axis list by grouping the axes into four broad categories: \emph{Context Length} (how much input must be processed), \emph{Answer Length} (how much output must be produced), \emph{Entity Complexity} (the number or variety of entities/attributes), and \emph{Procedural Complexity} (the depth or expressivity of the reasoning chain). We again examine winners both in final overall accuracy and on the hardest available level.

\begin{itemize}[leftmargin=1.5em,itemsep=1pt,topsep=2pt,parsep=0pt,partopsep=0pt]
  \item \textbf{Context Length:} representative pairs include \textsc{Calendar Scheduling} / \emph{Time horizon}, \textsc{Dyck Language} / \emph{Total sequence length}, \textsc{Index Select} / \emph{Source sequence length}, and \textsc{Survo} / \emph{Number of blanks}
  \item \textbf{Answer Length:} representative pairs include \textsc{Dyck Language} / \emph{Hidden suffix length}, \textsc{Index Select} / \emph{Selected-output length}, \textsc{Run-Length Encoding} / \emph{Maximum run length}, and \textsc{Survo} / \emph{Number of blanks}
  \item \textbf{Entity Complexity:} representative pairs include \textsc{Dyck Language} / \emph{Number of bracket types}, \textsc{Game of 24} / \emph{Input-number range}, \textsc{Index Select} / \emph{Alphabet size}, and \textsc{Survo} / \emph{Maximum allowed cell value}
  \item \textbf{Procedural Complexity:} representative pairs include \textsc{Calendar Scheduling} / \emph{Number of precedence constraints}, \textsc{Goods Exchange} / \emph{Number of exchange statements}, \textsc{Index Select} / \emph{Number of ranges}, and \textsc{Stack Operations} / \emph{Maximum stack depth}
\end{itemize}

\begin{center}
\begin{minipage}{\textwidth}
\centering
\setlength{\tabcolsep}{8pt}
\begin{tabular}{lcc}
\toprule
Axis category & Overall wins & Hardest-level wins \\
\midrule
\textbf{Context Length} & \(25/39\) (\(12\) W, \(13\) A) & \(23/39\) (\(13\) W, \(10\) A) \\
\textbf{Answer Length} & \(10/18\) (\(6\) W, \(4\) A) & \(15/18\) (\(11\) W, \(4\) A) \\
\textbf{Entity Complexity} & \(26/39\) (\(10\) W, \(16\) A) & \(23/39\) (\(8\) W, \(15\) A) \\
\textbf{Procedural Complexity} & \(11/30\) (\(3\) W, \(8\) A) & \(16/30\) (\(8\) W, \(8\) A) \\
\bottomrule
\end{tabular}
\captionof{table}{Difficulty-axis-category summary of when natural easy-to-hard progression helps. We group difficulty axes into four broad categories and count the number of task-by-axis-and-budget conditions for which \textit{Wasserstein} or \SYSNAME{} attains the best mean among the four forward curricula (\textit{static i.i.d.}, linear, Wasserstein, \SYSNAME{}). The left numeric pair reports overall wins; the right reports wins on the hardest available level. `W' denotes Wasserstein and `A' denotes \SYSNAME{}. Because some axes belong to more than one category, the denominators overlap across rows.}
\label{tab:appendix_c3_natural_progression_axes}
\end{minipage}
\end{center}

\paragraph{Axis findings.}
The axis taxonomy helps simplify the picture, but it still does not collapse the results to a single winning axis type. \emph{Context Length} and \emph{Entity Complexity} contain many overall wins, suggesting that natural easy-to-hard progression often helps when an axis enlarges the amount of input/state to process or the variety of entities/attributes. \emph{Answer Length} is the strongest category on the hardest level (\(15/18\)), driven by axes such as \emph{Hidden suffix length}, \emph{Maximum run length}, and \emph{Selected-output length}. \emph{Procedural Complexity} is more mixed: it is the weakest category overall, but its hardest-level rate improves substantially, suggesting that easy-to-hard progression can still help on the difficult end without consistently improving aggregate performance.

\begin{figure}[t]
  \centering
  \includegraphics[width=0.98\textwidth]{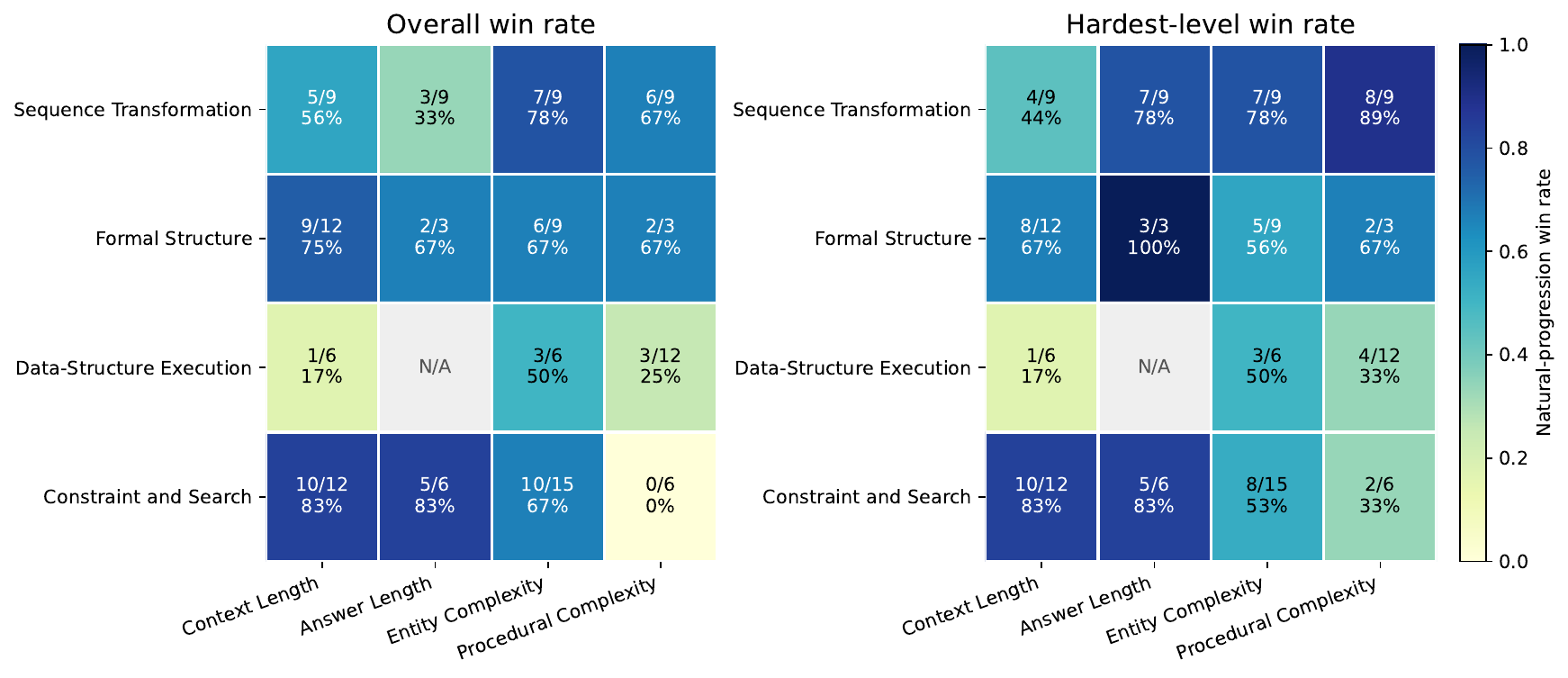}
  \caption{Joint task-family $\times$ difficulty-axis-category view of when natural easy-to-hard progression helps. Rows denote task families and columns denote difficulty-axis categories. The left panel shows overall win rates and the right panel shows hardest-level win rates for \textit{Wasserstein} or \SYSNAME{} against the other forward curricula. Each cell reports both the raw count and the percentage. Gray cells indicate that no task in that family has an axis of the corresponding category.}
  \label{fig:appendix_c3_task_axis_joint_heatmap}
\end{figure}

\paragraph{III. Joint View.}
Figure~\ref{fig:appendix_c3_task_axis_joint_heatmap} combines the two taxonomies by showing, for each task family and difficulty-axis category pair, the fraction of conditions in which \textit{Wasserstein} or \SYSNAME{} attains the best mean among the four forward curricula. The left panel reports overall wins and the right reports hardest-level wins; each cell is annotated with both the raw count and the corresponding win rate.

\paragraph{Joint findings.}
The interaction view clarifies where the previous summaries line up and where they do not. \emph{Sequence Transformation} is strongest when paired with \emph{Entity Complexity} or \emph{Procedural Complexity} axes, especially on the hardest level. \emph{Constraint and Search} is strongest when paired with \emph{Context Length} or \emph{Answer Length} axes, but shows little support on \emph{Procedural Complexity}. \emph{Formal Structure} remains broadly positive across categories, though with fewer total conditions, while \emph{Data-Structure Execution} stays weak across most cells. These three views suggest that easy-to-hard progression is easier to characterize at the level of coarse task family than at the level of semantic axis type, but even the joint taxonomy remains only partially predictive.

The taxonomy views above describe where natural easy-to-hard progression can help. We next ask what mechanisms may explain those gains by examining isolated level-to-level transfer and controlled bridge effects.

\subsection{Level Transfer Experiments}\label{appsubsec:level_transfer}

We study isolated transfer between difficulty levels to ask a simple mechanism question: when source-level training helps a target level, does it mainly give the target stage a \emph{head start}, or does it make target-stage learning itself \emph{faster}?

\paragraph{Setup.}
For each task, difficulty axis, seed, and ordered pair of distinct levels, we train on a source level and then continue only on a destination level. Let \(B\) denote the task's per-level small budget used in the main synthetic experiments. We write \(s \in \{0,0.2,0.4,0.6,0.8,1.0\}\) for the source-stage share, so the source stage uses \(sB\) steps while the destination stage always uses \(B\) steps. We compare each transferred run against the matched \(s=0\) no-transfer control. We start with lower-to-higher transfer and then return to higher-to-lower transfer when analyzing reverse ordering.

\begin{figure}[t]
\centering
\includegraphics[width=0.98\textwidth]{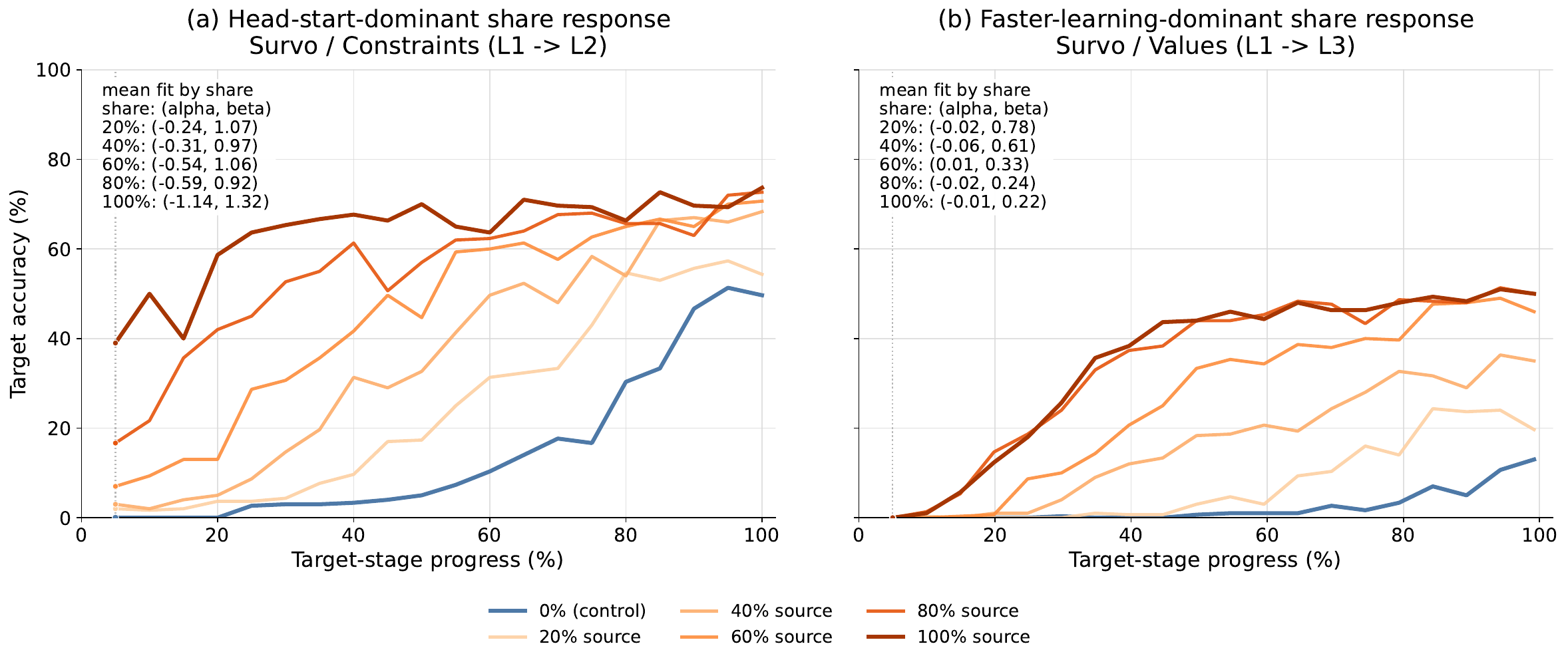}
\caption{Two easy-to-hard examples under the first-stage fit \(t_{\mathrm{tr},i}(a;s) \approx \alpha_{i,s} + \beta_{i,s}\, t_{\mathrm{ctrl},i}(a)\). We define \(t(a)\) as the first time a target-accuracy curve reaches \(a\). Blue: matched \(0\%\) source-share control. Orange shades: transferred runs with source shares \(s \in \{20,40,60,80,100\}\%\). Curves start at the first target-stage evaluation, at \(5\%\) progress. Panel~(a) is head-start dominant. Panel~(b) is faster-learning dominant. All axes are in percent.}
\label{fig:appendix_d1_level_transfer_decomposition}
\end{figure}

\begin{figure}[t]
\centering
\includegraphics[width=0.93\textwidth]{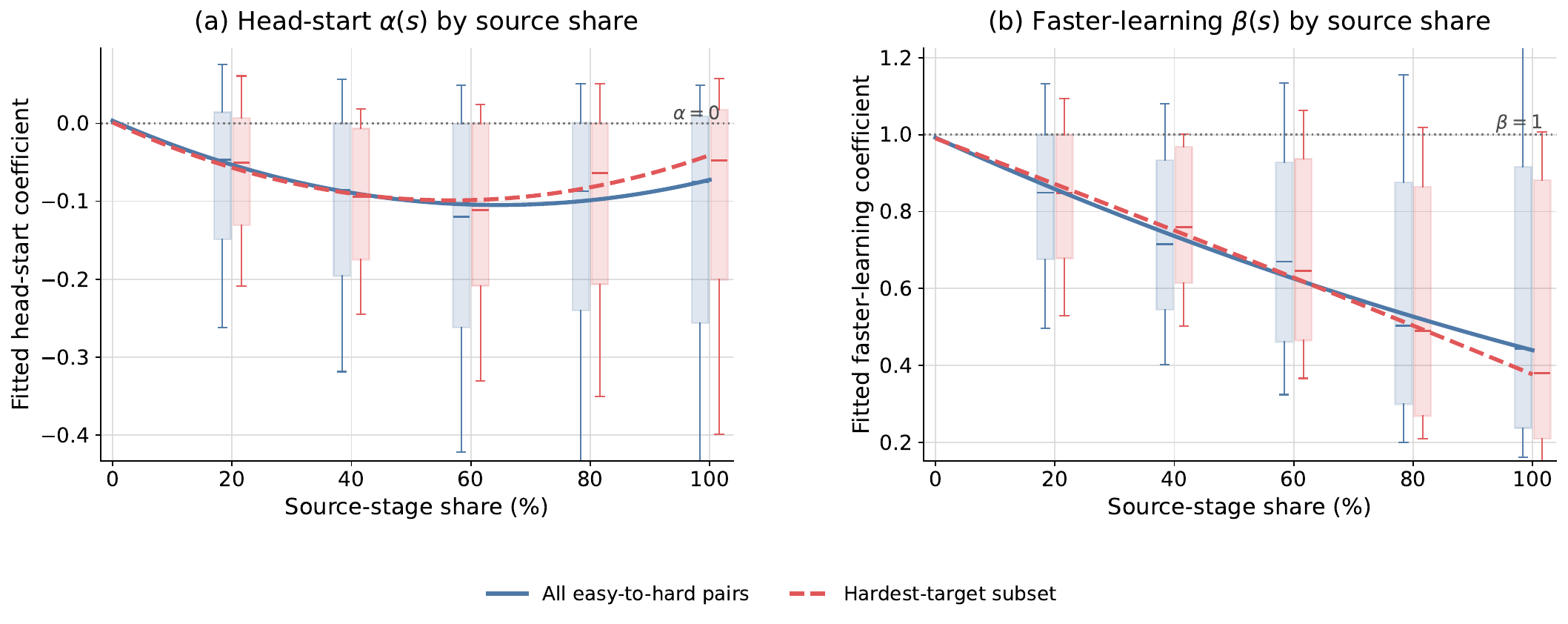}
\caption{Transfer coefficients by source share. We estimate \((\alpha_{i,s},\beta_{i,s})\) at each observed source share from earliest hitting times on target-accuracy curves, then fit quadratic trends to the median coefficients across shares (anchored at the no-transfer point). Curves show the median-by-share trend and boxplots show the empirical distribution. Solid blue uses all easy-to-hard ordered pairs. Dashed red uses the hardest-target subset.}
\label{fig:appendix_d1_fit_parameter_distributions}
\end{figure}

\paragraph{Fit by source share.}
For each easy-to-hard ordered pair \(i\) and each fixed \(s>0\), let \(A_{\mathrm{tr},i}(t;s)\) denote target accuracy during the target stage, where \(t \in [0,1]\) is normalized target-stage progress. Let \(A_{\mathrm{ctrl},i}(t)=A_{\mathrm{tr},i}(t;0)\) denote the matched no-transfer control. Over the accuracy range reached by both runs, we define \(t_{\mathrm{tr},i}(a;s)\) and \(t_{\mathrm{ctrl},i}(a)\) as the earliest target-stage progress at which each run first reaches accuracy \(a\), using linear interpolation between evaluations. We first fit
\[
\qquad
t_{\mathrm{tr},i}(a;s) \approx \alpha_{i,s} + \beta_{i,s}\, t_{\mathrm{ctrl},i}(a).
\]
This gives one fitted pair \((\alpha_{i,s},\beta_{i,s})\) for each ordered pair and each source share. We then model how these coefficients vary with source share:
\[
\alpha_{i,s}=u_{\alpha,i}+f_\alpha(s), \qquad \log \beta_{i,s}=u_{\beta,i}+f_\beta(s),
\]
where \(u_{\alpha,i}\) and \(u_{\beta,i}\) are pair-specific effects and \(f_\alpha,f_\beta\) are quadratic functions of \(s\). We anchor the fit at the no-transfer point \((s=0,\alpha=0,\beta=1)\). Negative \(\alpha_{i,s}\) indicates a head start. Values \(\beta_{i,s}<1\) indicate faster target-stage learning. Figure~\ref{fig:appendix_d1_level_transfer_decomposition} shows two representative first-stage examples. Figure~\ref{fig:appendix_d1_fit_parameter_distributions} shows how the fitted coefficients vary with source share, together with the empirical distributions.

\paragraph{Results.}
Figure~\ref{fig:appendix_d1_level_transfer_decomposition} shows two recurring patterns: larger source share either shifts the target-stage curve earlier or steepens the remaining climb. Figure~\ref{fig:appendix_d1_fit_parameter_distributions} shows the same pattern in aggregate. Across all easy-to-hard pairs, the fitted coefficients move from $(-0.08, 0.83)$ at $20\%$ source share to $(-0.43, 0.45)$ at $100\%$, so both the head-start effect and the faster-learning effect grow with source share. On hardest-target pairs, the pattern leans further toward faster learning: the fitted coefficients move from $(-0.07, 0.85)$ at $20\%$ to $(-0.12, 0.36)$ at $100\%$. The high end also flattens: on this subset, mean $\beta$ changes from $0.63$ to $0.55$ between $80\%$ and $100\%$, while mean $\alpha$ changes from $-0.15$ to $-0.11$.

\finding[find:level_transfer_warm_start]{Easy-to-hard transfer helps through both head start and faster target-stage learning. Across source shares, faster target-stage learning is especially prominent on the hardest targets.}

\paragraph{Directional asymmetry.}
Using isolated transfer at \(100\%\) source share, we compare each lower/higher level pair in both directions. Figure~\ref{fig:appendix_d1_reverse_directionality}(a) shows mean final target-accuracy gain over the matched \(0\%\) control; hard-to-easy transfer is often larger, especially toward easy targets. Panels~(b) and~(c) show why reverse still loses: it spends early budget on hard levels, leaving less remaining margin when it reaches easier targets, so the Wasserstein-over-reverse gap is small (often negative) on the easy bucket and large on the hard bucket. Reverse’s extra easy-bucket gains therefore do not offset Wasserstein’s larger hard-bucket gains.

\begingroup
\captionsetup{skip=4pt}
\begin{figure}[H]
\centering
\includegraphics[width=0.95\textwidth]{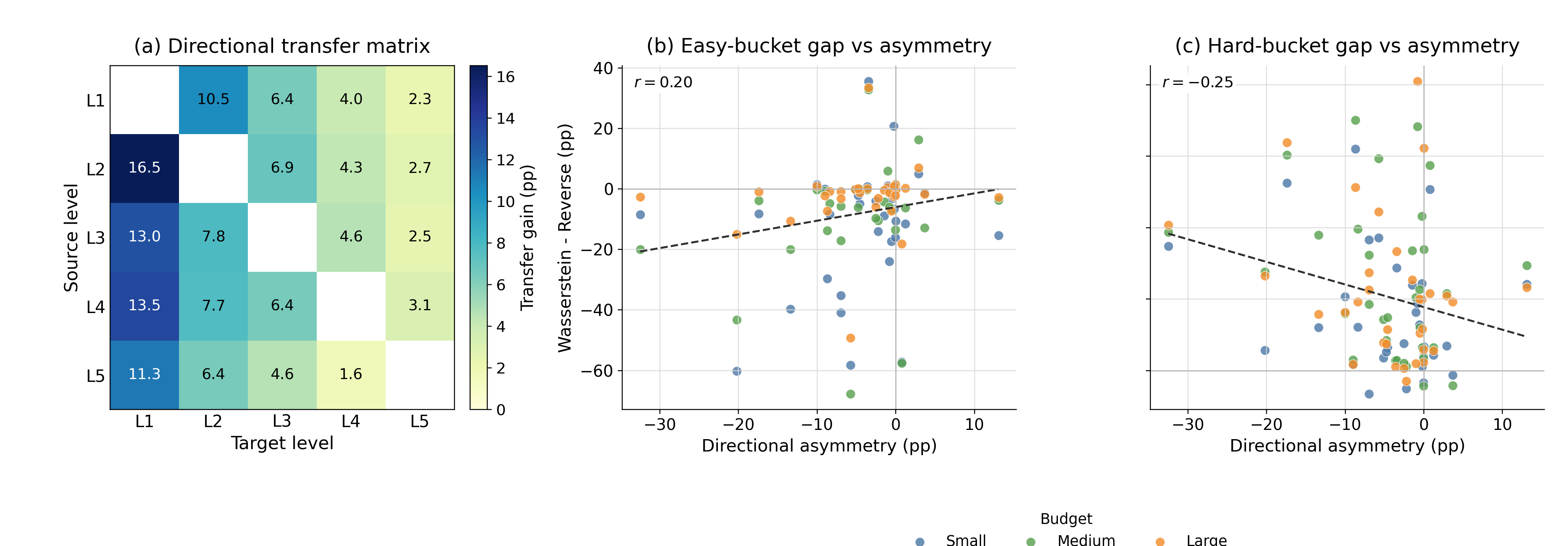}
\caption{Directional asymmetry in isolated transfer at \(100\%\) source share. Panel~(a) shows mean final target-accuracy gain over matched \(0\%\) controls for each ordered pair; the upper triangle is easy-to-hard and the lower triangle is hard-to-easy. Panels~(b) and~(c) show Wasserstein minus reverse gaps on the easy and hard buckets as a function of final-gain asymmetry.}
\label{fig:appendix_d1_reverse_directionality}
\end{figure}
\endgroup
\vspace{-0.8em}

\finding[find:level_transfer_reverse]{Hard-to-easy transfer is often larger than easy-to-hard transfer, but reverse Wasserstein still loses: it trains on hard levels first and reaches easier targets later with less remaining margin, so easy-bucket gains do not offset Wasserstein’s larger hard-bucket gains.}

\FloatBarrier
\subsection{Bridge Effect Experiments}\label{appsubsec:bridge_effect}

\paragraph{Setup.}
We test whether an intermediate level helps a hard target by running three-stage experiments on contiguous triples \((\ell,\ell+1,\ell+2)\). Stage 1 trains on the source level, stage 2 uses a fixed refine budget that is split between staying on the source and moving to the bridge level, and stage 3 trains on the target level. We write \(b \in \{0,0.25,0.5,0.75,1.0\}\) for the fraction of the stage-2 budget spent on the bridge level, where \(b=0\) is the skip-bridge control and \(b=1\) is the full-bridge condition. For triple \(i\), let \(A_i(b)\) denote final target accuracy. To use the full bridge-share sweep, we fit
\[
A_i(b) \approx u_i + \lambda_i b,
\]
where \(u_i\) is a triple-specific intercept and \(\lambda_i\) is the bridge-share slope. Larger \(\lambda_i\) means that moving more of stage 2 onto the bridge level helps more.

\begin{figure}[t]
\centering
\includegraphics[width=0.90\textwidth]{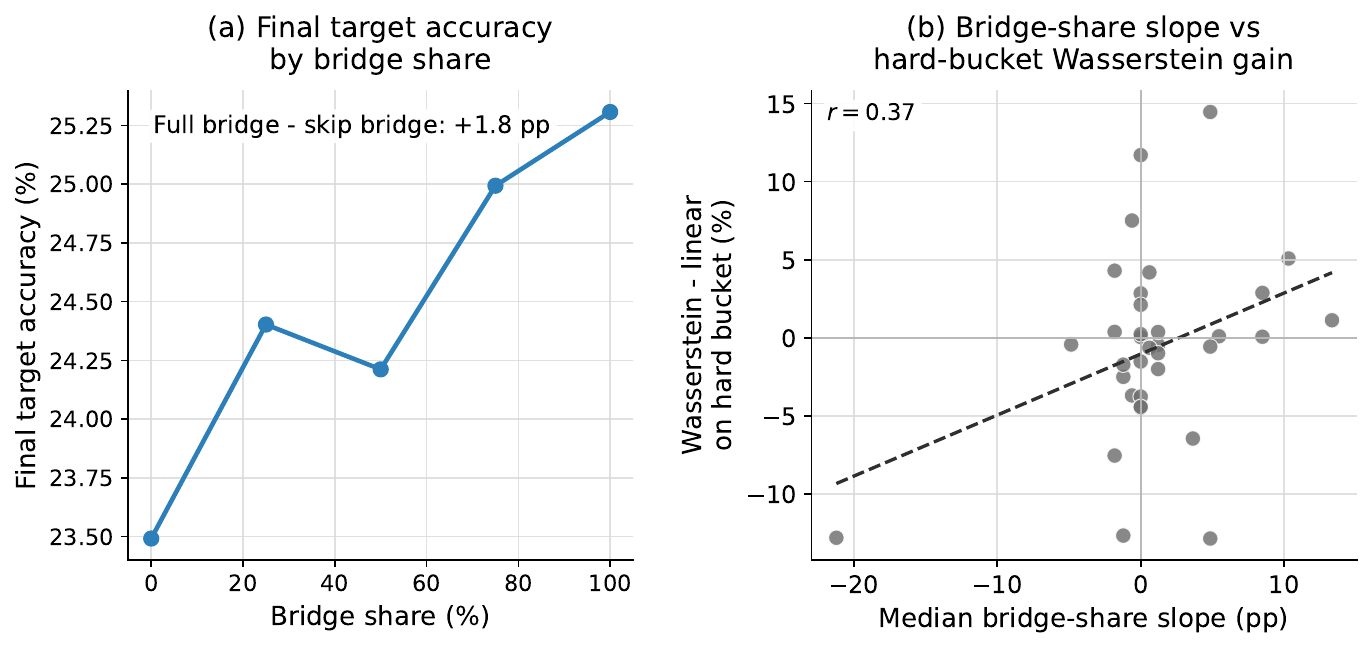}
\caption{Bridge-share response experiments. Panel~(a) shows mean final target accuracy versus bridge share, averaged over bridge triples and seeds. Panel~(b) plots, for each task-by-axis setting, the median bridge-share slope over hardest-target triples against the medium-budget hard-bucket Wasserstein--linear gap.}
\label{fig:appendix_d2_bridge_effects}
\end{figure}

\paragraph{Results.}
Figure~\ref{fig:appendix_d2_bridge_effects}(a) shows an overall upward response: mean final target accuracy rises from \(23.49\%\) at \(0\%\) bridge share to \(25.31\%\) at \(100\%\), though the sweep is not perfectly monotone. Figure~\ref{fig:appendix_d2_bridge_effects}(b) then plots, for each task-by-axis setting, the median bridge-share slope over hardest-target triples against the medium-budget hard-bucket Wasserstein--linear gap from Section~\ref{subsec:when_do_curricula_help}. The association is positive (\(r=0.37\)). Settings whose hard targets improve more as bridge share increases also tend to be the settings where Wasserstein gains more over linear on the hard bucket. This is consistent with the same local-transfer picture suggested by Figure~\ref{fig:appendix_d1_reverse_directionality}(a): an intermediate level can split one hard jump into two easier ones.

\finding[find:bridge_effect_hard_efficiency]{Larger bridge effects tend to align with larger Wasserstein-over-linear gains.}

\section{Real-World SFT Analyses}\label{app:real_world_sft_analyses}

This appendix studies the pretrained SFT setting. We first evaluate real-world SFT across three base models, then use a synthetic-pretraining analogue to test whether pretraining itself weakens curriculum effects.

\subsection{Real-World SFT Benchmarks}\label{appsubsec:sft_benchmarks}

\paragraph{Setup.}
We run LoRA \citep{hu2022lora} SFT on ARC \citep{clark2018arc}, MMLU-STEM (the STEM subset of MMLU; \citealt{hendrycks2021mmlu}), and StrategyQA \citep{geva2021strategyqa}, following the benchmark construction and difficulty labels from \citet{hase2024unreasonable}. We compare \texttt{SmolLM3-3B}\footnote{\url{https://huggingface.co/HuggingFaceTB/SmolLM3-3B-Base}}, \texttt{OLMo2-1B}\footnote{\url{https://huggingface.co/allenai/OLMo-2-0425-1B}}, and \texttt{Qwen3-1.7B}\footnote{\url{https://huggingface.co/Qwen/Qwen3-1.7B-Base}} under the same curriculum sweep, with three seeds, learning rate \(5\times 10^{-6}\), and batch size 128.

\paragraph{Benchmarks and difficulty axes.}
We study 12 benchmark-by-axis contexts: six for ARC, three for MMLU-STEM, and three for StrategyQA. ARC uses \emph{human grade} (6 levels), \emph{human difficulty} (3), \emph{human Bloom level} (5), \emph{human depth of knowledge} (3), \emph{question length} (5 quantile bins), and \emph{answer length} (5 quantile bins). MMLU-STEM uses \emph{human hardness} (2 levels), \emph{question length} (5 quantile bins), and \emph{answer length} (5 quantile bins) over the five STEM subject groups from \citet{hase2024unreasonable}. StrategyQA uses \emph{decomposition length} (5 levels), \emph{question length} (5 quantile bins), and \emph{reasoning length} (5 quantile bins). Training targets contain a short reasoning chain plus the final answer, while evaluation uses only the final answer.

\paragraph{Budgets.}
We calibrate small, medium, and large budgets separately for each dataset: ARC uses 32, 64, and 128 update steps; MMLU-STEM uses 16, 32, and 64; and StrategyQA uses 64, 96, and 128. Unless otherwise stated, we average over the same three random seeds.

\paragraph{Static i.i.d. Remains the Strongest Summary Baseline.}
Tables~\ref{tab:appendix_c0_sft_basic_curriculum_summary_smolm3}, \ref{tab:appendix_c0_sft_basic_curriculum_summary_olmo}, and \ref{tab:appendix_c0_sft_basic_curriculum_summary_qwen} summarize the three basic curricula over these 12 contexts for each model. Across all three models, \textit{static i.i.d.} has the highest mean overall accuracy and the highest mean hardest-level accuracy in every model-budget block. Linear and Wasserstein still win isolated contexts, but the gaps are small and much less systematic than in the synthetic suite.

\begin{center}
\begin{minipage}{0.98\textwidth}
\centering
\footnotesize
\setlength{\tabcolsep}{4pt}
\begin{tabular}{lcccc}
\toprule
Curriculum & Mean overall (\%) & Mean hardest (\%) & Overall wins & Hardest wins \\
\midrule
\multicolumn{5}{l}{\textit{Small budget}} \\
Static i.i.d. & \(\mathbf{57.5}\) & \(\mathbf{53.7}\) & \textbf{8} & \textbf{8} \\
Linear & \(57.0\) & \(52.5\) & 3 & 3 \\
Wasserstein & \(57.1\) & \(52.7\) & 1 & 1 \\
\midrule
\multicolumn{5}{l}{\textit{Medium budget}} \\
Static i.i.d. & \(\mathbf{63.4}\) & \(\mathbf{58.3}\) & \textbf{6} & \textbf{6} \\
Linear & \(62.7\) & \(57.3\) & 3 & 1 \\
Wasserstein & \(62.7\) & \(58.1\) & 3 & 5 \\
\midrule
\multicolumn{5}{l}{\textit{Large budget}} \\
Static i.i.d. & \(\mathbf{67.6}\) & \(\mathbf{63.4}\) & \textbf{9} & \textbf{6} \\
Linear & \(66.0\) & \(62.1\) & 1 & 4 \\
Wasserstein & \(66.6\) & \(62.9\) & 2 & 2 \\
\bottomrule
\end{tabular}
\captionof{table}{SFT summary for the three basic curricula on ARC, MMLU-STEM, and StrategyQA using \texttt{SmolLM3-3B}. Each budget block contains 12 benchmark-by-axis contexts: six ARC axes, three MMLU-STEM axes, and three StrategyQA axes. ``Hardest'' denotes the last available level.}
\label{tab:appendix_c0_sft_basic_curriculum_summary_smolm3}
\end{minipage}
\vspace{0.9em}
\begin{minipage}{0.98\textwidth}
\centering
\footnotesize
\setlength{\tabcolsep}{4pt}
\begin{tabular}{lcccc}
\toprule
Curriculum & Mean overall (\%) & Mean hardest (\%) & Overall wins & Hardest wins \\
\midrule
\multicolumn{5}{l}{\textit{Small budget}} \\
Static i.i.d. & \(\mathbf{37.6}\) & \(\mathbf{35.5}\) & 4 & \textbf{6} \\
Linear & \(37.4\) & \(35.4\) & 3 & 3 \\
Wasserstein & \(37.5\) & \(35.5\) & \textbf{5} & 3 \\
\midrule
\multicolumn{5}{l}{\textit{Medium budget}} \\
Static i.i.d. & \(\mathbf{39.0}\) & \(\mathbf{36.8}\) & \textbf{6} & \textbf{5} \\
Linear & \(38.2\) & \(36.1\) & 1 & 3 \\
Wasserstein & \(38.5\) & \(36.4\) & 5 & 4 \\
\midrule
\multicolumn{5}{l}{\textit{Large budget}} \\
Static i.i.d. & \(\mathbf{45.6}\) & \(\mathbf{42.8}\) & \textbf{8} & \textbf{6} \\
Linear & \(44.4\) & \(41.6\) & 3 & 5 \\
Wasserstein & \(44.4\) & \(41.3\) & 1 & 1 \\
\bottomrule
\end{tabular}
\captionof{table}{SFT summary for the three basic curricula on ARC, MMLU-STEM, and StrategyQA using \texttt{OLMo2-1B}. Each budget block contains 12 benchmark-by-axis contexts: six ARC axes, three MMLU-STEM axes, and three StrategyQA axes. ``Hardest'' denotes the last available level.}
\label{tab:appendix_c0_sft_basic_curriculum_summary_olmo}
\end{minipage}
\vspace{0.9em}
\begin{minipage}{0.98\textwidth}
\centering
\footnotesize
\setlength{\tabcolsep}{4pt}
\begin{tabular}{lcccc}
\toprule
Curriculum & Mean overall (\%) & Mean hardest (\%) & Overall wins & Hardest wins \\
\midrule
\multicolumn{5}{l}{\textit{Small budget}} \\
Static i.i.d. & \(\mathbf{62.6}\) & \(\mathbf{60.4}\) & \textbf{6} & \textbf{5} \\
Linear & \(62.1\) & \(60.2\) & 3 & 3 \\
Wasserstein & \(61.8\) & \(60.4\) & 3 & 4 \\
\midrule
\multicolumn{5}{l}{\textit{Medium budget}} \\
Static i.i.d. & \(\mathbf{68.6}\) & \(\mathbf{66.4}\) & \textbf{5} & \textbf{5} \\
Linear & \(68.1\) & \(66.0\) & 4 & \textbf{5} \\
Wasserstein & \(68.1\) & \(65.7\) & 3 & 2 \\
\midrule
\multicolumn{5}{l}{\textit{Large budget}} \\
Static i.i.d. & \(\mathbf{71.4}\) & \(\mathbf{67.8}\) & \textbf{7} & \textbf{6} \\
Linear & \(70.5\) & \(67.4\) & 4 & 4 \\
Wasserstein & \(70.6\) & \(67.4\) & 1 & 2 \\
\bottomrule
\end{tabular}
\captionof{table}{SFT summary for the three basic curricula on ARC, MMLU-STEM, and StrategyQA using \texttt{Qwen3-1.7B}. Each budget block contains 12 benchmark-by-axis contexts: six ARC axes, three MMLU-STEM axes, and three StrategyQA axes. ``Hardest'' denotes the last available level.}
\label{tab:appendix_c0_sft_basic_curriculum_summary_qwen}
\end{minipage}
\end{center}

\subsection{Final Level Distributions, Bucket Profiles, and Remaining Efficiency Differences}\label{appsubsec:sft_profiles}
Appendix~\ref{appsubsec:sft_benchmarks} showed that \textit{static i.i.d.} is the strongest summary baseline in pretrained SFT. We next ask whether the forward curricula still differ in exposure allocation and exposure-adjusted accuracy.

\paragraph{Setup.}
We study the medium budget and the four forward curricula: \textit{static i.i.d.}, linear, Wasserstein, and \SYSNAME{}. To keep the level view direct, we restrict to the 10 contexts whose native difficulty axes already have 3 or 5 levels, excluding only ARC \emph{human grade} (6 levels) and MMLU-STEM \emph{human hardness} (2 levels). Thus 3-level axes map directly to \emph{easy}/\emph{middle}/\emph{hard}, and 5-level axes to \emph{easy}/\emph{lower-mid}/\emph{middle}/\emph{upper-mid}/\emph{hard}. We report both five-position exposure/accuracy summaries and bucketed easy/middle/hard summaries with exposure-adjusted accuracy.

Figure~\ref{fig:appendix_c1b_sft_full_level_profiles} shows the five-position view. Exposure shifts much more than accuracy: linear pushes exposure toward the ends, while Wasserstein and \SYSNAME{} push it toward the middle, but the final accuracy curves remain close. \SYSNAME{} also stays close to fixed Wasserstein.

\begin{center}
  \begin{minipage}{\textwidth}
  \centering
  \includegraphics[width=0.98\textwidth]{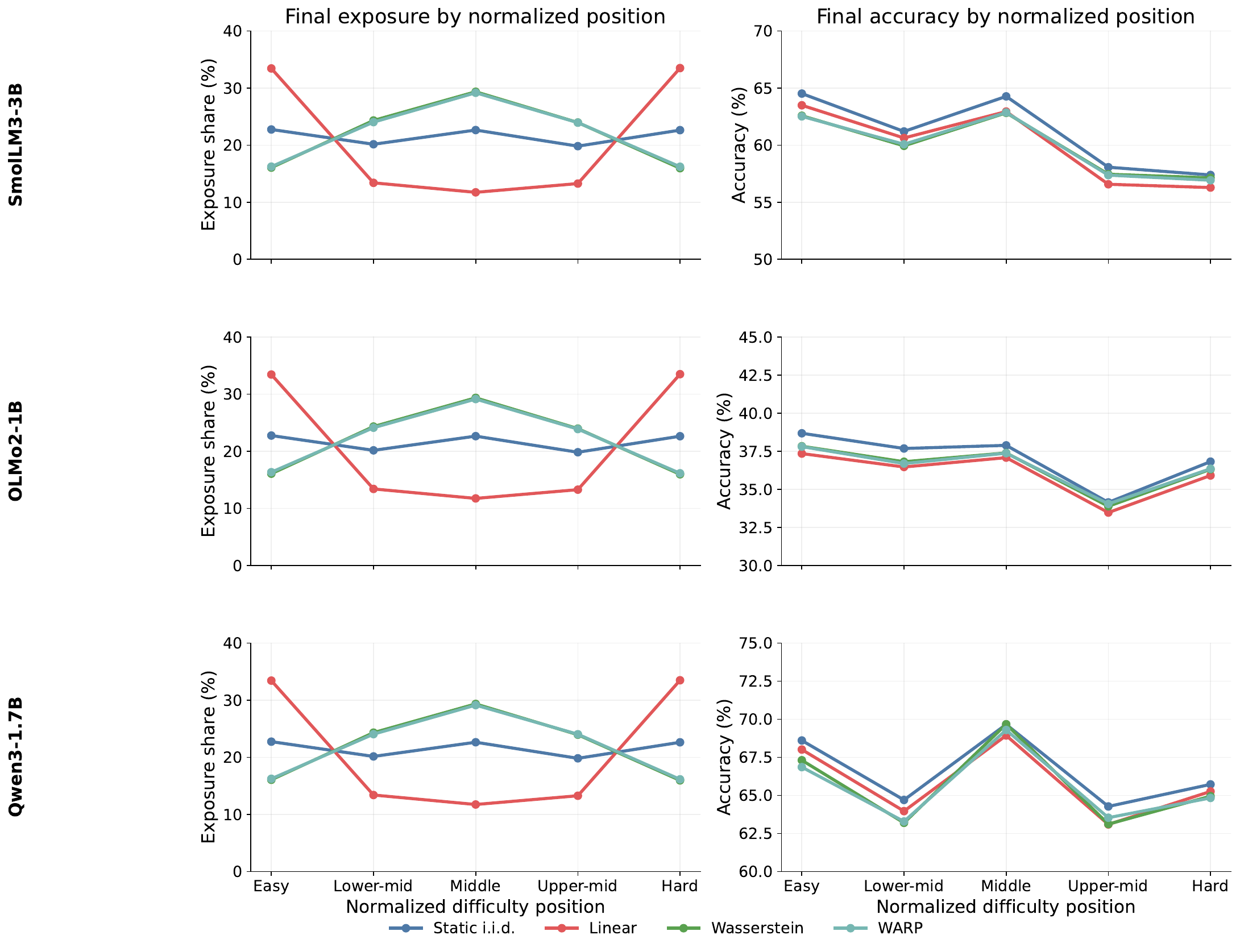}
  \captionof{figure}{Medium-budget SFT exposure and final accuracy on the 10 retained 3- or 5-level contexts. Rows show the three models; curves show the four forward curricula.}
  \label{fig:appendix_c1b_sft_full_level_profiles}
  \end{minipage}
\end{center}

Figure~\ref{fig:appendix_c1_sft_curriculum_profiles} shows the bucketed view. Raw bucket accuracies remain compressed despite large exposure differences, so the exposure-adjusted view is more informative. Linear is strongest in the middle bucket, while Wasserstein and \SYSNAME{} are strongest on the hard bucket, but the gaps are modest and model-dependent.

\begin{center}
  \begin{minipage}{\textwidth}
  \centering
  \includegraphics[width=0.98\textwidth]{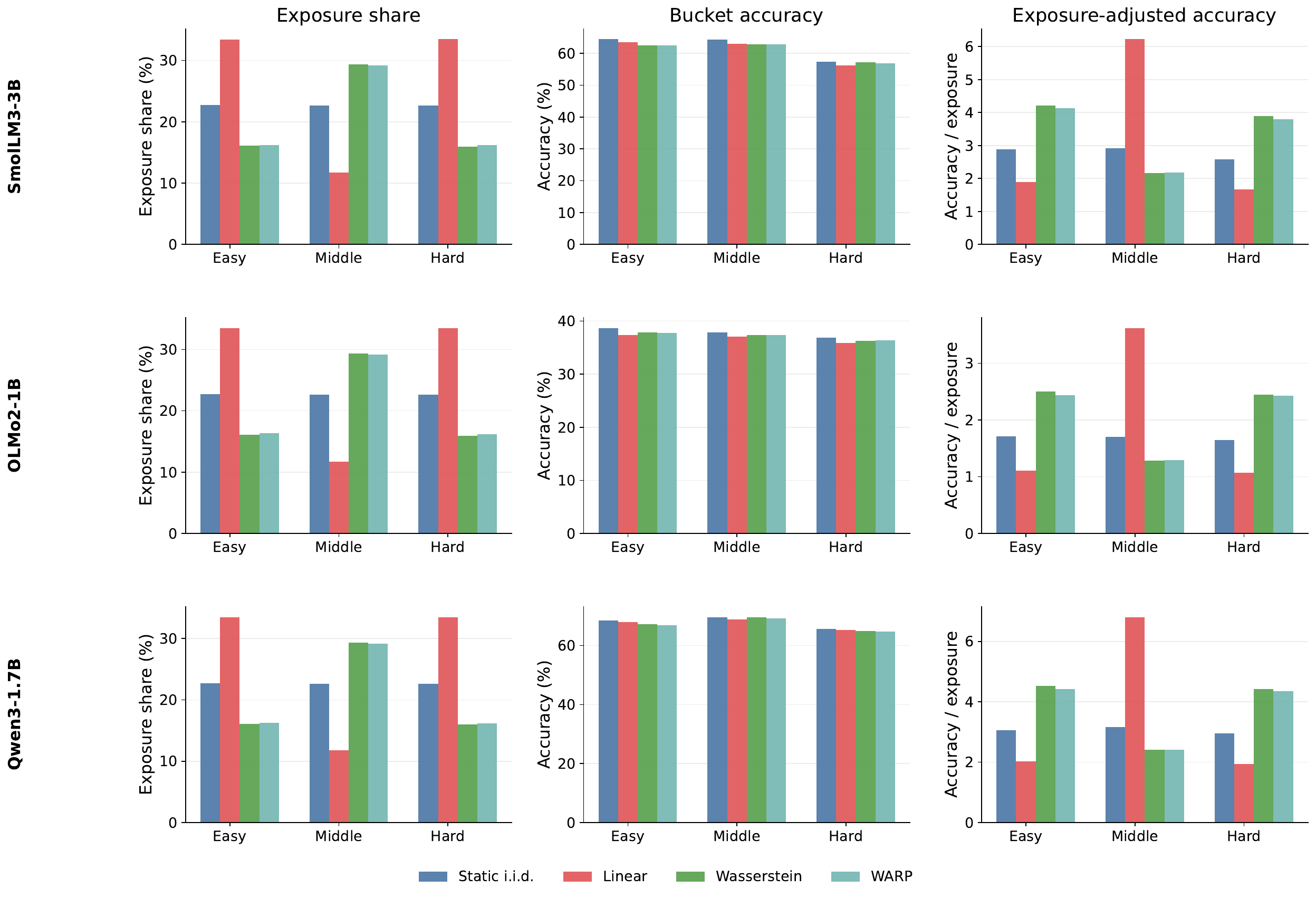}
  \captionof{figure}{Bucket-wise exposure, final accuracy, and exposure-adjusted accuracy for medium-budget SFT on the same 10 retained contexts. Rows show the three models; columns show exposure, accuracy, and exposure-adjusted accuracy under the four forward curricula.}
  \label{fig:appendix_c1_sft_curriculum_profiles}
  \end{minipage}
\end{center}

\subsection{Easy-to-Hard Ordering Effect Is Weak in Pretrained SFT}

Appendix~\ref{appsubsec:sft_profiles} showed that pretrained SFT largely flattens the forward-curriculum profiles. We next ask whether any ordering signal survives once exposure geometry is held fixed.

\paragraph{Setup.}
Figure~\ref{fig:appendix_c2_sft_ordering_profiles} compares three Wasserstein-family controls across the small, medium, and large budgets: forward Wasserstein, reverse Wasserstein, and the exposure-matched static baseline. We keep all 12 benchmark-by-axis contexts and report easy, middle, and hard bucket means for each model.

\begin{center}
  \begin{minipage}{\textwidth}
  \centering
  \includegraphics[width=0.98\textwidth]{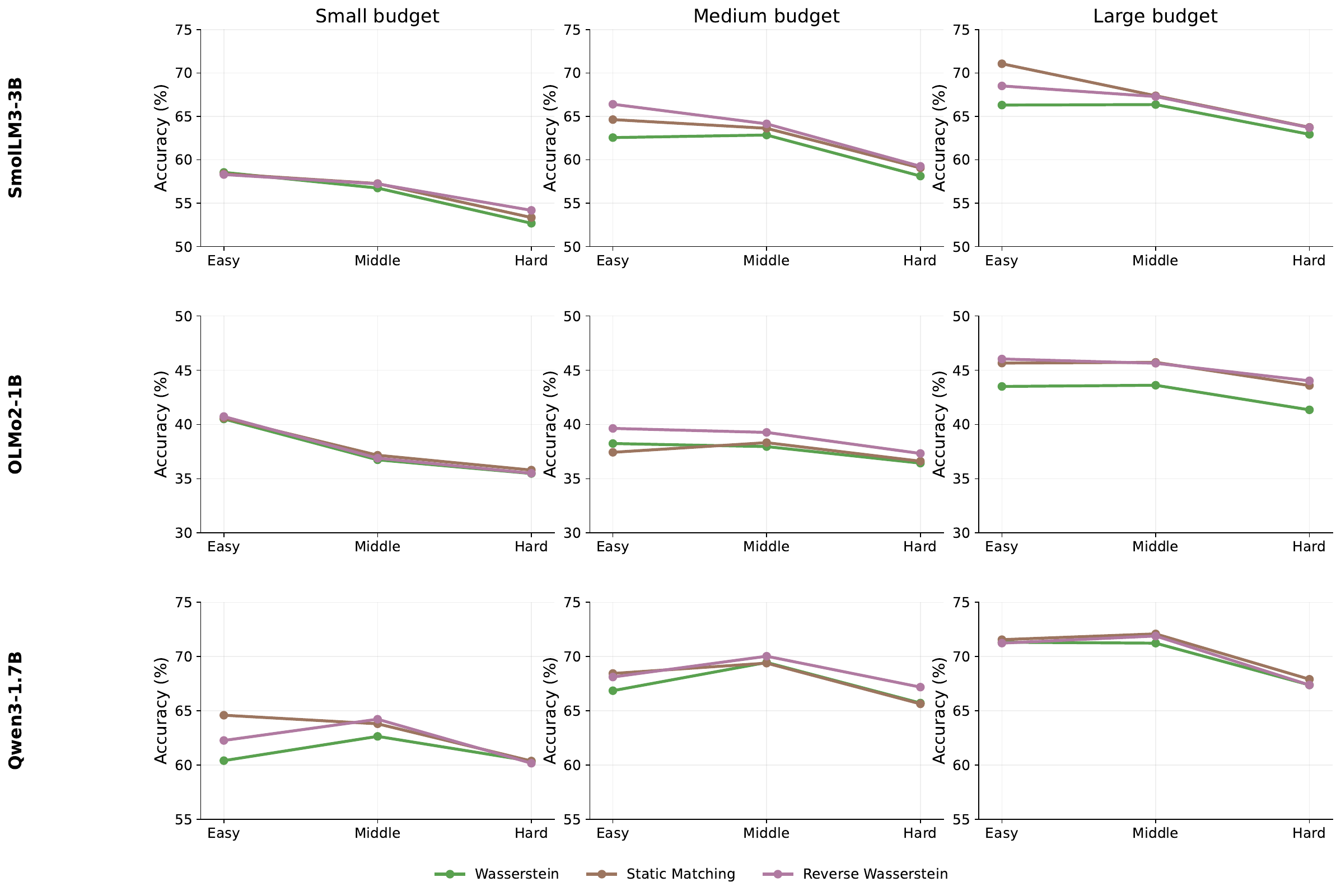}
  \captionof{figure}{Ordering comparison within the Wasserstein family for SFT on ARC, MMLU-STEM, and StrategyQA. Rows show the three models; columns show the small, medium, and large budgets. ``Static Matching'' denotes the exposure-matched static baseline.}
  \label{fig:appendix_c2_sft_ordering_profiles}
  \end{minipage}
\end{center}

\paragraph{Forward ordering is no longer distinctly stronger.}
Across all nine model-by-budget blocks, forward Wasserstein never has the highest hard-bucket mean; the exposure-matched static baseline or the reverse-order control is always as good or better. Under pretrained SFT, changing the ordering within a fixed geometry no longer yields a robust easy-to-hard advantage.

\finding[find:sft_pretrained_summary]{Under pretrained SFT, most curriculum effects seen in the synthetic study weaken: \textit{static i.i.d.} becomes the strongest summary baseline, bucket accuracies flatten, and easy-to-hard ordering no longer separates reliably. The clearest remaining difference is in exposure-adjusted data efficiency.}

\subsection{Exploratory Analysis of When Easy-to-Hard Progression Helps}
We next ask whether the remaining easy-to-hard wins cluster by benchmark or difficulty-axis type.

\paragraph{Setup.}
We analyze conditions in which \textit{Wasserstein} or \SYSNAME{} attains the best mean among the four forward curricula, counting ties for all tied curricula under each metric. We then summarize these wins by benchmark and by difficulty-axis type.

\begin{center}
  \begin{minipage}{\textwidth}
  \centering
  \includegraphics[width=0.98\textwidth]{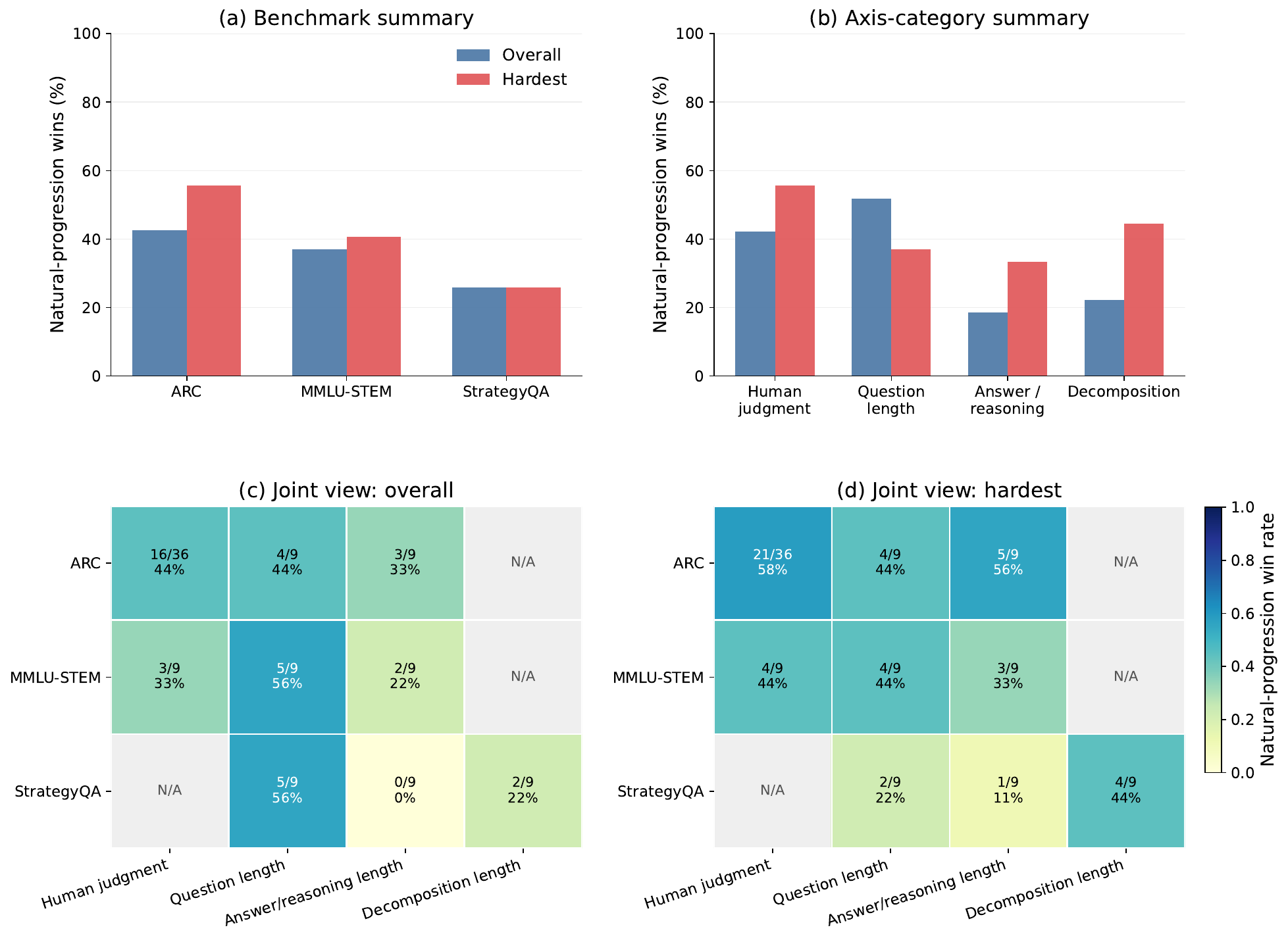}
  \captionof{figure}{Where natural easy-to-hard progression helps in SFT. Panel~(a) summarizes wins by benchmark, panel~(b) by axis category, and panels~(c)--(d) show the joint benchmark $\times$ axis-category view for overall and hardest-level wins.}
  \label{fig:appendix_d4_sft_taxonomy_summary}
  \end{minipage}
\end{center}

\paragraph{The remaining easy-to-hard wins are concentrated on ARC and human-annotated difficulty axes.} Figure~\ref{fig:appendix_d4_sft_taxonomy_summary} shows only a weak concentration of wins. ARC has the most wins, especially on hardest-level accuracy, and human-annotated difficulty axes have the highest hardest-level win rate among the axis categories. However, neither pattern is consistent across both overall and hardest-level accuracy. Unlike Appendix~\ref{appsubsec:natural_progression_axes}, the SFT setting does not yield a stable benchmark- or axis-level pattern.

\subsection{Controlled Synthetic Pretraining Analogue}\label{appsubsec:synthetic_pretraining_analogue}
To isolate the effect of pretraining, we vary the pretraining budget on the synthetic suite while holding the downstream tasks and evaluation protocol fixed.

\paragraph{Setup.}
In the pretraining stage, we train a shared 6-layer decoder on nine synthetic tasks from Appendix~\ref{app:task_details}. We use a uniform multitask mixture that splits each task's small-budget quota evenly across its reported difficulty axes and levels, then globally shuffles the resulting examples. We study three pretraining regimes, \emph{small}, \emph{medium}, and \emph{large}, corresponding to \(20\%\), \(50\%\), and \(100\%\) of the sum of the task-wise small budgets. Downstream, we keep the earlier synthetic setup, restrict the sweep to the small budget, and remove pretraining-seen examples from the downstream train split.

\paragraph{Aggregate gaps narrow as pretraining grows.}
Table~\ref{tab:appendix_c3_synthetic_pretraining_fraction_summary} shows the aggregate picture. As pretraining increases, the three basic curricula become harder to distinguish. The static baseline strengthens substantially: its mean hardest-level accuracy rises from \(84.7\%\) under small pretraining to \(91.5\%\) under medium pretraining and \(93.7\%\) under large pretraining. Even so, unlike the SFT setting, static does not become the uniformly strongest hardest-level baseline. Linear and Wasserstein continue to match or exceed it on many contexts.

\begin{center}
\setlength{\tabcolsep}{6pt}
\begin{tabular}{lcccc}
\toprule
Curriculum & Mean overall (\%) & Mean hardest (\%) & Overall wins & Hardest wins \\
\midrule
\multicolumn{5}{l}{\textit{Small pretraining}} \\
Static i.i.d. & \(\mathbf{91.3}\) & \(84.7\) & 10 & 6 \\
Linear & \(91.2\) & \(\mathbf{88.1}\) & \textbf{13} & \textbf{17} \\
Wasserstein & \(91.2\) & \(87.1\) & 9 & 11 \\
\midrule
\multicolumn{5}{l}{\textit{Medium pretraining}} \\
Static i.i.d. & \(95.0\) & \(91.5\) & 13 & 13 \\
Linear & \(95.0\) & \(\mathbf{92.9}\) & \textbf{17} & \textbf{18} \\
Wasserstein & \(95.0\) & \(92.3\) & 10 & 11 \\
\midrule
\multicolumn{5}{l}{\textit{Large pretraining}} \\
Static i.i.d. & \(96.0\) & \(93.7\) & \textbf{19} & \textbf{18} \\
Linear & \(96.4\) & \(\mathbf{94.6}\) & 16 & \textbf{18} \\
Wasserstein & \(\mathbf{96.5}\) & \(\mathbf{94.6}\) & 14 & 14 \\
\bottomrule
\end{tabular}
\captionof{table}{Synthetic-pretraining summary for the three basic curricula under a fixed small downstream budget. Each pretraining block contains the same 26 synthetic task-by-axis contexts. Here, ``large pretraining'' denotes the full multitask pretraining budget. The win columns count the number of contexts in which a curriculum attains the best mean under the corresponding metric; ties are counted for all tied curricula. ``Hardest'' denotes the last available level.}
\label{tab:appendix_c3_synthetic_pretraining_fraction_summary}
\end{center}

\paragraph{Level-wise profiles flatten as pretraining grows.}
Figure~\ref{fig:appendix_d4_synthetic_pretraining_curriculum_profiles} shows the same narrowing in level-wise profiles. Under \textit{static i.i.d.}, the mean easy--hard gap falls from \(10.9\) points under small pretraining to \(5.9\) under medium pretraining and \(4.8\) under large pretraining; under Wasserstein, it shrinks from \(5.5\) to \(4.5\) to \(3.3\). This mirrors the SFT trend, though part of the narrowing here comes from easy-level saturation near \(100\%\). Even so, the exposure-adjusted pattern survives: linear remains strongest on the middle bucket, while Wasserstein remains strongest on the hard bucket.

\begin{center}
  \begin{minipage}{\textwidth}
  \centering
  \includegraphics[width=0.98\textwidth]{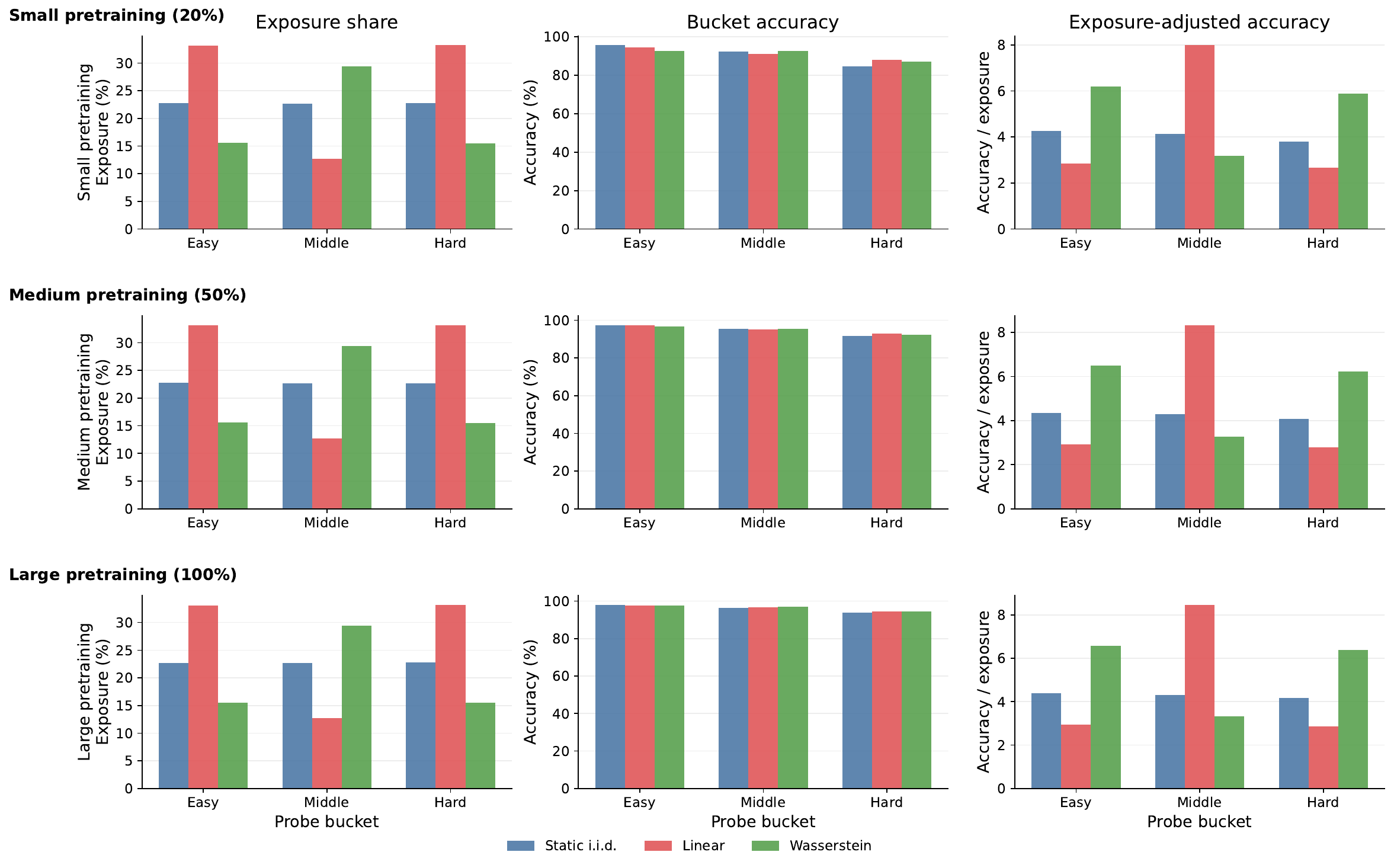}
  \captionof{figure}{Synthetic-pretraining analogue under a fixed small downstream budget. Rows vary pretraining amount; columns show level-wise exposure, final accuracy, and exposure-adjusted accuracy for \textit{static i.i.d.}, linear, and Wasserstein.}
  \label{fig:appendix_d4_synthetic_pretraining_curriculum_profiles}
  \end{minipage}
\end{center}

\paragraph{Ordering weakens, but does not disappear.} Figure~\ref{fig:appendix_d4_synthetic_pretraining_ordering_profiles} shows that the easy-to-hard advantage within the Wasserstein family shrinks steadily as pretraining increases. The Wasserstein--reverse gap falls from \(13.5\) points under small pretraining to \(2.3\) under large pretraining, while the Wasserstein--exposure-matched-static gap falls from \(3.9\) to \(0.2\). The corresponding linear--reverse-linear gap similarly decreases from \(8.5\) to \(1.6\). Thus, increasing pretraining weakens ordering effects on this suite without eliminating them.

\begin{center}
  \begin{minipage}{\textwidth}
  \centering
  \includegraphics[width=0.98\textwidth]{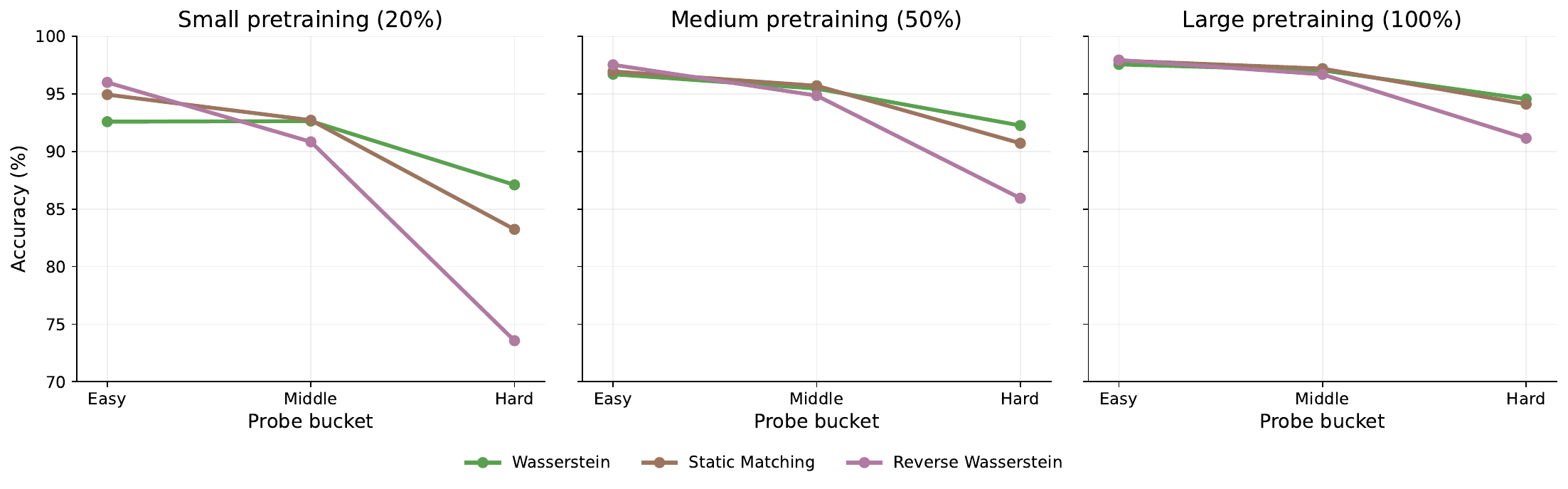}
  \captionof{figure}{Ordering comparison within the Wasserstein family under a fixed small downstream budget. Panels vary pretraining amount. ``Static Matching'' denotes the exposure-matched static baseline.}
  \label{fig:appendix_d4_synthetic_pretraining_ordering_profiles}
  \end{minipage}
\end{center}

\paragraph{Relation to pretrained SFT.} The controlled synthetic analogue isolates pretraining as one factor behind the weaker curriculum effects observed in SFT. As pretraining increases, the differences among curricula shrink, static i.i.d.\ becomes stronger, and the easy-to-hard ordering advantage weakens. In the synthetic setting, these changes coincide with less remaining room for improvement, especially as easier levels approach saturation, suggesting one mechanism through which pretraining can reduce curriculum gains. Because saturation is less evident in the SFT experiments, other properties of natural post-training data may also matter.

\finding[find:synth_pretraining_summary]{
Pretraining weakens downstream curriculum effects, helping explain why they are weaker in pretrained SFT; less remaining room for improvement may be one mechanism.
}

\end{document}